# Gathering Strength, Gathering Storms:

## The One Hundred Year Study on Artificial Intelligence (AI100) 2021 Study Panel Report

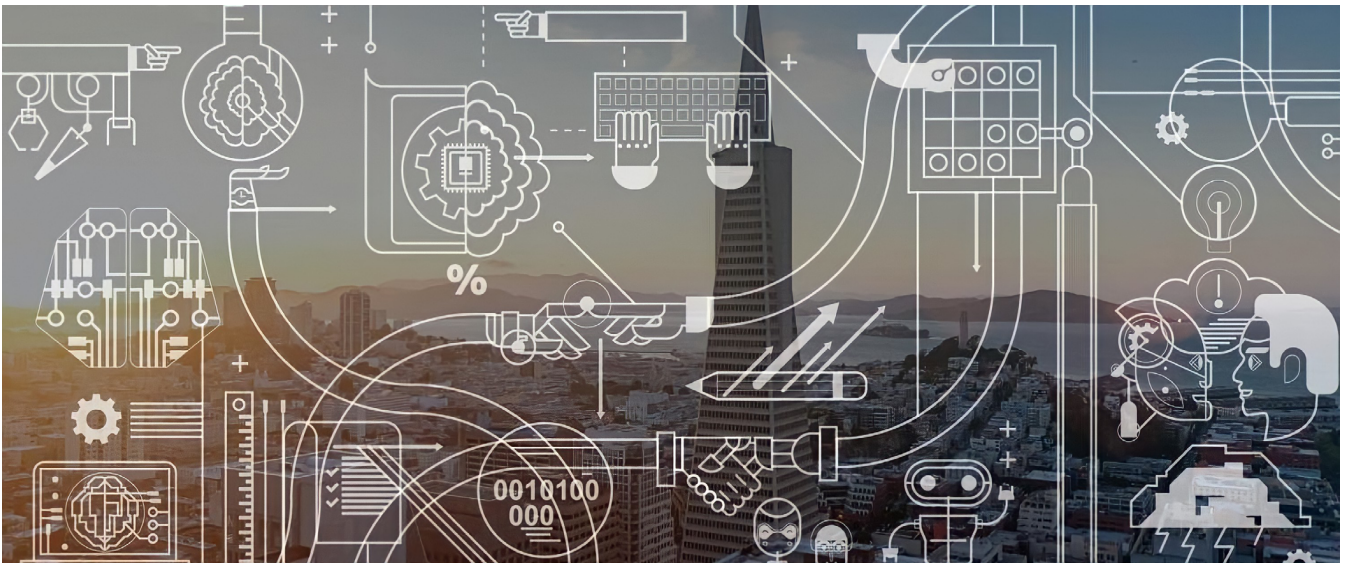

## PREFACE

In the five years since we released the first AI100 report, much has been written about the state of artificial intelligence and its influences on society. Nonetheless, AI100 remains unique in its combination of two key features. First, it is written by a Study Panel of core multi-disciplinary researchers in the field—experts who create artificial intelligence algorithms or study their influence on society as their main professional activity, and who have been doing so for many years. The authors are firmly rooted within the field of AI and provide an "insider's" perspective. Second, it is a longitudinal study, with reports by such Study Panels planned once every five years, for at least one hundred years.

This report, the second in that planned series of studies, is being released five years after the first report. Published on September 1, 2016, the first report was covered widely in the popular press and is known to have influenced discussions on governmental advisory boards and workshops in multiple countries. It has also been used in a variety of artificial intelligence curricula.

In preparation for the second Study Panel, the Standing Committee commissioned two study-workshops held in 2019. These workshops were a response to feedback on the first AI100 report. Through them, the Standing Committee aimed to engage a broader, multidisciplinary community of scholars and stakeholders in its next study. The goal of the workshops was to draw on the expertise of computer scientists and engineers, scholars in the social sciences and humanities (including anthropologists, economists, historians, media scholars, philosophers, psychologists, and sociologists), law and public policy experts, and representatives from business management as well as the private and public sectors.

An expanded Standing Committee, with more expertise in ethics and the social sciences, formulated a call and actively encouraged proposals from the international community of AI researchers and practitioners with a broad representation of fields relevant to AI's impact in the world. By convening scholars from the full range of disciplines that rigorously explore ethical and societal impacts of technologies, the study-workshops were aimed at expanding and deepening discussions of the ways in which AI shapes the hopes, concerns, and realities of people's lives, and, relatedly, the ethical and societal-impact challenges that AI raises.

After circulating a call for proposals and reviewing more than 100 submissions from around the world, two workshops were selected for funding. One, on "Prediction in Practice," studied the use of AI-driven predictions of human behavior, such as how likely a borrower is to eventually repay a loan, in settings where data and cognitive modeling fail to account for the social dimensions that shape people's decision-making. The other, on "Coding Caring," studied the challenges and opportunities of incorporating AI technologies into the process of humans caring for one another and the role that gender and labor relationships play in addressing the pressing need for innovation in healthcare.

Drawing on the findings from these study-workshops, as well as the annual AI Index report, a project spun off from AI100, the Standing Committee defined a charge for the Study Panel in the summer of 2019[1] and recruited Professor Michael Littman, Professor of Computer Science at Brown University, to chair the panel. The 17-member Study Panel, composed of a diverse set of experts in AI, from academia and industry research laboratories, representing computer science, engineering, law, political science, policy, sociology, and economics, was launched in mid-fall 2020. In addition to representing a range of scholarly specialties, the panel had diverse representation in terms of home geographic regions, genders, and career stages. As readers may note in the report, convening this diverse, interdisciplinary set of scholarly experts, allowed varying perspectives, rarely brought together, to be reconciled and juxtaposed within the report. The accomplishment of the Study Panel is that much more impressive considering the inability to meet face-to-face during the ongoing COVID-19 global pandemic.

Whereas the first study report focused explicitly on the impact of AI in North American cities, we sought for the 2021 study to explore in greater depth the impact that AI is having on people and societies worldwide. AI is being deployed in applications that touch people's lives in a critical and personal way (for example, through loan approvals, criminal sentencing, healthcare, emotional care, and influential recommendations in multiple realms, for example). Since these society-facing applications will influence people's relationship with AI technologies, as well as have far-reaching socioeconomic implications, we entitled the charge, "Permeating Influences of AI in Everyday Life: Hopes, Concerns, and Directions."

In addition to including topics directly related to these society-facing applications that resulted from the aforementioned workshops (as represented by WQ1 and WQ2 of this report), the Standing Committee carefully considered how to launch the Study Panel for the second report in such a way that it would set a precedent for all subsequent Study Panels, emphasizing the unique longitudinal aspect of the AI100 study. Motivated by the notion that it takes at least two points to define a line, as noted by AI100 founder Eric Horvitz, the Study Panel charge suggested a set of "standing questions" for the Study Panel to consider that could potentially then be answered by future Study Panels as well (as represented by SQ1-SQ12 of this report) and included a call to reflect on the first report, indicating what has changed and what remains the same (as represented here).

While the scope of this charge was broader than the inaugural panel's focus on typical North American cities, it still does not—and cannot—cover all aspects of AI's influences on society, leaving some topics to be introduced or explored further in subsequent reports. In particular, military applications were outside the scope of the first report; and while military AI is used as a key case study in one section of this report (SQ7), vigorous discussions of the subject are still continuing worldwide and opinions are evolving.

Like the first report, the second report aspires to address four audiences. For the general public, it aims to provide an accessible, scientifically and technologically accurate portrayal of the current state of AI and its potential. For industry, the report describes relevant technologies and legal and ethical challenges, and may help guide resource allocation. The report is also directed to local, national, and international governments to help them better plan for AI in governance. Finally, the report can help AI researchers, as well as their institutions and funders, to set priorities and consider the economic, ethical, and legal issues raised by AI research and its applications.

The Standing Committee is grateful to the members of the Study Panel for investing their expertise, perspectives, and significant time into the creation of this report. We are also appreciative of the contributions of the leaders and participants of the workshops mentioned above, as well as past members of the Standing Committee, whose contributions were invaluable in setting the stage for this report: Yoav Shoham and Deirdre Mulligan (2015-2017); Tom Mitchell and Alan Mackworth (2015-2018); Milind Tambe (2018); and Eric Horvitz (2015-2019).We especially thank Professor Michael Littman for agreeing to serve as chair of the study and for his wise, skillful, and dedicated leadership of the panel, its discussions, and creation of the report.

## Standing Committee of the One Hundred Year Study of Artificial Intelligence

Peter Stone, The University of Texas at Austin and Sony AI, *Chair*

Russ Altman, Stanford University

Erik Brynjolfsson, Stanford University

Vincent Conitzer, Duke University and University of Oxford

Mary L. Gray, Microsoft Research

Barbara Grosz, Harvard University

Ayanna Howard, The Ohio State University

Percy Liang, Stanford University

Patrick Lin, California Polytechnic State University

James Manyika, McKinsey & Company

Sheila McIlraith, University of Toronto

Liz Sonenberg, The University of Melbourne

Judy Wajcman, London School of Economics and The Alan Turing Institute

## Organizers of the Preparatory Workshops

Thomas Arnold, Tufts University

Solon Barocas, Microsoft Research

Miranda Bogen, Upturn

Morgan Currie, The University of Edinburgh

Andrew Elder, The University of Edinburgh

Jessica Feldman, American University of Paris

Johannes Himmelreich, Syracuse University

Jon Kleinberg, Cornell University

Karen Levy, Cornell University

Fay Niker, Cornell Tech

Helen Nissenbaum, Cornell Tech

David G. Robinson, Upturn

# ABOUT AI100

*The following history of AI100 first appeared in the preface of the 2016 report.*

"The One Hundred Year Study on Artificial Intelligence (AI100), launched in the fall of 2014, is a long-term investigation of the field of Artificial Intelligence (AI) and its influences on people, their communities, and society. It considers the science, engineering, and deployment of AI-enabled computing systems. As its core activity, the Standing Committee that oversees the One Hundred Year Study forms a Study Panel every five years to assess the current state of AI. The Study Panel reviews AI's progress in the years following the immediately prior report, envisions the potential advances that lie ahead, and describes the technical and societal challenges and opportunities these advances raise, including in such arenas as ethics, economics, and the design of systems compatible with human cognition. The overarching purpose of the One Hundred Year Study's periodic expert review is to provide a collected and connected set of reflections about AI and its influences as the field advances. The studies are expected to develop syntheses and assessments that provide expert-informed guidance for directions in AI research, development, and systems design, as well as programs and policies to help ensure that these systems broadly benefit individuals and society.

"The One Hundred Year Study is modeled on an earlier effort informally known as the "AAAI Asilomar Study." During 2008-2009, the then president of the Association for the Advancement of Artificial Intelligence (AAAI), Eric Horvitz, assembled a group of AI experts from multiple institutions and areas of the field, along with scholars of cognitive science, philosophy, and law. Working in distributed subgroups, the participants addressed near-term AI developments, long-term possibilities, and legal and ethical concerns, and then came together in a three-day meeting at Asilomar to share and discuss their findings. A short written report on the intensive meeting discussions, amplified by the participants' subsequent discussions with other colleagues, generated widespread interest and debate in the field and beyond.

"The impact of the Asilomar meeting, and important advances in AI that included AI algorithms and technologies starting to enter daily life around the globe, spurred the idea of a long-term recurring study of AI and its influence on people and society. The One Hundred Year Study was subsequently endowed at a university to enable extended deep thought and cross-disciplinary scholarly investigations that could inspire innovation and provide intelligent advice to government agencies and industry."

# INTRODUCTION

This report is structured as a collection of responses by the 2021 Study Panel to a collection of 12 standing questions (SQs) and two workshop questions (WQs) posed by the AI100 Standing Committee. The report begins with a list of the 14 questions and short summaries of the panel's responses, which serves as an overview of the report's findings. It then dives into the responses themselves and a brief conclusion section. An appendix includes a collection of annotations to the prior report in the AI100 series, published in 2016.

# HOW TO CITE THIS REPORT

## Study Panel

Michael L. Littman, Brown University, *Chair*

Ifeoma Ajunwa, University of North Carolina

Guy Berger, LinkedIn

Craig Boutilier, Google

Morgan Currie, The University of Edinburgh

Finale Doshi-Velez, Harvard University

Gillian Hadfield, University of Toronto

Michael C. Horowitz, University of Pennsylvania

Charles Isbell, Georgia Institute of Technology

Hiroaki Kitano, Okinawa Institute of Science and Technology Graduate University and Sony AI

Karen Levy, Cornell University

Terah Lyons

Melanie Mitchell, Santa Fe Institute and Portland State University

Julie Shah, Massachusetts Institute of Technology

Steven Sloman, Brown University

Shannon Vallor, The University of Edinburgh

Toby Walsh, University of New South Wales

## Acknowledgments

The panel would like to thank the members of the Standing Committee, listed in the preface. In addition to setting the direction and vision for the report, they provided detailed and truly insightful comments on everything from tone to detailed word choices that made the report clearer and, we hope!, more valuable in the long run. Standing Committee chair Peter Stone, in particular, deserves particular credit for his remarkable ability to find ways to negotiate clever solutions to the not-uncommon differences of opinion that inevitably arise by design of having a diverse set of contributors. We are grateful to Hillary Rosner, who, with the help of Philip Higgs and Stephen Miller, provided exceptionally valuable writing and editorial support. Jacqueline Tran and Russ Altman were deeply and adeptly involved in coordinating the efforts of both the Standing Committee and the Study Panel. We also thank colleagues who have provided pointers or feedback or other insights that helped inform our treatment on technical issues such as the use of AI in healthcare. They include: Nigam Shah, Jenna Wiens, Mark Sendak, Michael Sjoding, Jim Fackler, Mert Sabuncu, Leo Celi, Susan Murphy, Dan Lizotte, Jacqueline Kueper, Ravninder Bahniwal, Leora Horwitz, Russ Greiner, Philip Resnik, Manal Siddiqui, Jennifer Rizk, Martin Wattenberg, Na Li, Weiwei Pan, Carlos Carpi, Yiling Chen, Sarah Rathnam.

# TABLE OF CONTENTS

# STANDING QUESTIONS AND SECTION SUMMARIES

## SQ1. What are some examples of pictures that reflect important progress in AI and its influences?

One picture appears in each of the sections that follow.

## SQ2. What are the most important advances in AI?

People are using AI more today to dictate to their phone, get recommendations, enhance their backgrounds on conference calls, and much more. Machine-learning technologies have moved from the academic realm into the real world in a multitude of ways. Neural network language models learn about how words are used by identifying patterns in naturally occurring text, supporting applications such as machine translation, text classification, speech recognition, writing aids, and chatbots. Image-processing technology is now widespread, but applications such as creating photo-realistic pictures of people and recognizing faces are seeing a backlash worldwide. During 2020, robotics development was driven in part by the need to support social distancing during the COVID-19 pandemic. Predicted rapid progress in fully autonomous driving failed to materialize, but autonomous vehicles have begun operating in selected locales. AI tools now exist for identifying a variety of eye and skin disorders, detecting cancers, and supporting measurements needed for clinical diagnosis. For financial institutions, uses of AI are going beyond detecting fraud and enhancing cybersecurity to automating legal and compliance documentation and detecting money laundering. Recommender systems now have a dramatic influence on people's consumption of products, services, and content, but they raise significant ethical concerns.

## SQ3. What are the most inspiring open grand challenge problems?

Recent years have seen remarkable progress on some of the challenge problems that help drive AI research, such as answering questions based on reading a textbook, helping people drive so as to avoid accidents, and translating speech in real time. Others, like making independent mathematical discoveries, have remained open. A lesson learned from social science- and humanities-inspired research over the past five years is that AI research that is overly tuned to concrete benchmarks can take us further away from the goal of cooperative and well-aligned AI that serves humans' needs, goals, and values. A number of broader challenges should be kept in mind: exhibiting greater generalizability, detecting and using causality, and noticing and exhibiting normativity are three particularly important ones. An overarching and inspiring challenge that brings many of these ideas together is to build machines that can cooperate and collaborate seamlessly with humans and can make decisions that are aligned with fluid and complex human values and preferences.

## SQ4. How much have we progressed in understanding the key mysteries of human intelligence?

A view of human intelligence that has gained prominence over the last five years holds that it is collective—that individuals are just one cog in a larger intellectual machine. AI is developing in ways that improve its ability to collaborate with and support people, rather than in ways that mimic human intelligence. The study of intelligence has become the study of how people are able to adapt and succeed, not just how an impressive information-processing system works. Over the past half decade, major shifts in the understanding of human intelligence have favored three topics: collective intelligence, the view that intelligence is a property not only of individuals, but also of collectives; cognitive neuroscience, studying how the brain's hardware is involved in implementing psychological and social processes; and computational modeling, which is now full of machine-learning-inspired models of visual recognition, language processing, and other cognitive activities. The nature of consciousness and how people integrate information from multiple modalities, multiple senses, and multiple sources remain largely mysterious. Insights in these areas seem essential in our quest for building machines that we would truly judge as "intelligent."

## SQ5. What are the prospects for more general artificial intelligence?

The field is still far from producing fully general AI systems. However, in the last few years, important progress has been made in the form of three key capabilities. First is the ability for a system to learn in a self-supervised or self-motivated way. A self-supervised model called transformers has become the go-to approach for natural language processing, and has been used in diverse applications, including machine translation and Google web search. Second is the ability for a single AI system to learn in a continual way to solve problems from many different domains without requiring extensive retraining for each. One influential approach is to train a deep neural network on a variety of tasks, where the objective is for the network to learn general-purpose, transferable representations, as opposed to representations tailored specifically to any particular task. Third is the ability for an AI system to generalize between tasks— that is, to adapt the knowledge and skills the system has acquired for one task to new situations. A promising direction is the use of intrinsic motivation, in which an agent is rewarded for exploring new areas of the problem space. AI systems will likely remain very far from human abilities, however, without being more tightly coupled to the physical world.

## SQ6. How has public sentiment towards AI evolved, and how should we inform/educate the public?

Over the last few years, AI and related topics have gained traction in the zeitgeist. In the 2017–18 session of the US Congress, for instance, mentions of AI-related words were more than ten times higher than in previous sessions. Media coverage of AI often distorts and exaggerates AI's potential at both the positive and negative extremes, but it has helped to raise public awareness of legitimate concerns about AI bias, lack of transparency and accountability, and the potential of AI-driven automation to contribute to rising inequality. Governments, universities, and nonprofits are attempting to broaden the reach of AI education, including investing in new AI-related curricula. Nuanced views of AI as a human responsibility are growing, including an increasing effort to engage with ethical considerations. Broad international movements in Europe, the US, China, and the UK have been pushing back against the indiscriminate use of facial-recognition systems on the general public. More public outreach from AI scientists would be beneficial as society grapples with the impacts of these technologies. It is important that the AI research community move beyond the goal of educating or talking to the public and toward more participatory engagement and conversation with the public.

## SQ7. How should governments act to ensure AI is developed and used responsibly?

Since the publication of the last AI100 report just five years ago, over 60 countries have engaged in national AI initiatives, and several significant new multilateral efforts are aimed at spurring effective international cooperation on related topics. To date, few countries have moved definitively to regulate AI specifically, outside of rules directly related to the use of data. As of 2020, 24 countries had opted for permissive laws to allow autonomous vehicles to operate in limited settings. As yet, only Belgium has enacted laws on the use of autonomous lethal weapons. The oversight of social media platforms has become a hotly debated issue worldwide. Cooperative efforts among countries have also emerged in the last several years. Appropriately addressing the risks of AI applications will inevitably involve adapting regulatory and policy systems to be more responsive to the rapidly advancing pace of technology development. Researchers, professional organizations, and governments have begun development of AI or algorithm impact assessments (akin to the use of environmental impact assessments before beginning new engineering projects).

## SQ8. What should the roles of academia and industry be, respectively, in the development and deployment of AI technologies and the study of the impacts of AI?

As AI takes on added importance across most of society, there is potential for conflict between the private and public sectors regarding the development, deployment, and oversight of AI technologies. The commercial sector continues to lead in AI investment, and many researchers are opting out of academia for full-time roles in industry. The presence and influence of industry-led research at AI conferences has increased dramatically, raising concerns that published research is becoming more applied and that topics that might run counter to commercial interests will be underexplored. As student interest in computer science and AI continues to grow, more universities are developing standalone AI/machine-learning educational programs. Company-led courses are becoming increasingly common and can fill curricular gaps. Studying and assessing the societal impacts of AI, such as concerns about the potential for AI and machine-learning algorithms to shape polarization by influencing content consumption and user interactions, is easiest when academic-industry collaborations facilitate access to data and platforms.

## SQ9. What are the most promising opportunities for AI?

AI approaches that augment human capabilities can be very valuable in situations where humans and AI have complementary strengths. An AI system might be better at synthesizing available data and making decisions in well-characterized parts of a problem, while a human may be better at understanding the implications of the data. It is becoming increasingly clear that all stakeholders need to be involved in the design of AI assistants to produce a human-AI team that outperforms either alone. AI software can also function autonomously, which is helpful when large amounts of data needs to be examined and acted upon. Summarization and interactive chat technologies have great potential. As AI becomes more applicable in lower-data regimes, predictions can increase the economic efficiency of everyday users by helping people and businesses find relevant opportunities, goods, and services, matching producers and consumers. We expect many mundane and potentially dangerous tasks to be taken over by AI systems in the near future. In most cases, the main factors holding back these applications are not in the algorithms themselves, but in the collection and organization of appropriate data and the effective integration of these algorithms into their broader sociotechnical systems.

## SQ10. What are the most pressing dangers of AI?

As AI systems prove to be increasingly beneficial in real-world applications, they have broadened their reach, causing risks of misuse, overuse, and explicit abuse to proliferate. One of the most pressing dangers of AI is techno-solutionism, the view that AI can be seen as a panacea when it is merely a tool. There is an aura of neutrality and impartiality associated with AI decision-making in some corners of the public consciousness, resulting in systems being accepted as objective even though they may be the result of biased historical decisions or even blatant discrimination. Without transparency concerning either the data or the AI algorithms that interpret it, the public may be left in the dark as to how decisions that materially impact their lives are being made. AI systems are being used in service of disinformation on the internet, giving them the potential to become a threat to democracy and a tool for fascism. Insufficient thought given to the human factors of AI integration has led to oscillation between mistrust of the system and over-reliance on the system. AI algorithms are playing a role in decisions concerning distributing organs, vaccines, and other elements of healthcare, meaning these approaches have literal life-and-death stakes.

## SQ11. How has AI impacted socioeconomic relationships?

Though characterized by some as a key to increasing material prosperity for human society, AI's potential to replicate human labor at a lower cost has also raised concerns about its impact on the welfare of workers. To date, AI has not been responsible for large aggregate economic effects. But that may be because its impact is still relatively localized to narrow parts of the economy. In the grand scheme of rising inequality, AI has thus far played a very small role. The first reason, most importantly, is that the bulk of the increase in economic inequality across many countries predates significant commercial use of AI. Since these technologies might be adopted by firms simply to redistribute surplus/gains to their owners, AI could have a big impact on inequality in the labor market and economy without registering any impact on productivity growth. No evidence of such a trend is yet apparent, but it may become so in the future and is worth watching closely. To date, the economic significance of AI has been comparatively small—particularly relative to expectations, among both optimists and pessimists. Other forces—globalization, the business cycle, and a pandemic—have had a much, much bigger and more intense impact in recent decades. But if policymakers underreact to coming changes, innovations may simply result in a pie that is sliced ever more unequally.

## SQ12. Does it appear "building in how we think" works as an engineering strategy in the long run?

AI has its own fundamental nature-versus-nurture-like question. Should we attack new challenges by applying general-purpose problem-solving methods, or is it better to write specialized algorithms, designed by experts, for each particular problem? Roughly, are specific AI solutions better engineered in advance by people (nature) or learned by the machine from data (nurture)? The pendulum has swung back and forth multiple times in the history of the field. In the 2010s, the addition of big data and faster processors allowed general-purpose methods like deep learning to outperform specialized hand-tuned methods. But now, in the 2020s, these general methods are running into limits—available computation, model size, sustainability, availability of data, brittleness, and a lack of semantics—that are starting to drive researchers back into designing specialized components of their systems to try to work around them. Indeed, even machine-learning systems benefit from designers using the right architecture for the right job. The recent dominance of deep learning may be coming to an end. To continue making progress, AI researchers will likely need to embrace both general- and special-purpose hand-coded methods, as well as ever faster processors and bigger data.

# WORKSHOP QUESTIONS AND SECTION SUMMARIES

## WQ1. How are AI-driven predictions made in high-stakes public contexts, and what social, organizational, and practical considerations must policymakers consider in their implementation and governance?: Lessons from "Prediction in Practice" workshop

Researchers are developing predictive systems to respond to contentious and complex public problems across all types of domains, including criminal justice, healthcare, education, and social services—high-stakes contexts that can impact quality of life in material ways. Success is greatly influenced by how a system is or is not integrated into existing decision-making processes, policies, and institutions. The ways we define and formalize prediction problems shape how an algorithmic system looks and functions. Even subtle differences in problem definition can significantly change resulting policies. The most successful predictive systems are not dropped into place but are thoughtfully integrated into existing social and organizational environments and practices. Matters are further complicated by questions about jurisdiction and the imposition of algorithmic objectives at a state or regional level that are inconsistent with the goals held by local decision-makers. Successfully integrating AI into high-stakes public decision-making contexts requires difficult work, deep and multidisciplinary understanding of the problem and context, cultivation of meaningful relationships with practitioners and affected communities, and a nuanced understanding of the limitations of technical approaches.

## WQ2. What are the most pressing challenges and significant opportunities in the use of artificial intelligence to provide physical and emotional care to people in need?: Lessons from "Coding Caring" workshop

Smart home devices can give Alzheimer's patients medication reminders, pet avatars and humanoid robots can offer companionship, and chatbots can help veterans living with PTSD treat their mental health. These intimate forms of AI caregiving challenge how we think of core human values, like privacy, compassion, trust, and the very idea of care itself. AI offers extraordinary tools to support caregiving and increase the autonomy and well-being of those in need. Some patients may even express a preference for robotic care in contexts where privacy is an acute concern, as with intimate bodily functions or other activities where a non-judgmental helper may preserve privacy or dignity. However, in elder care, particularly for dementia patients, companion robots will not replace the human decision-makers who increase a patient's comfort through intimate knowledge of their conditions and needs. The use of AI technologies in caregiving should aim to supplement or augment existing caring relationships, not replace them, and should be integrated in ways that respect and sustain those relationships. Good care demands respect and dignity, things that we simply do not know how to code into procedural algorithms. Innovation and convenience through automation should not come at the expense of authentic care.

# Study Panel Responses to Standing Questions and Workshop Questions

## SQ1. WHAT ARE SOME EXAMPLES OF PICTURES THAT REFLECT IMPORTANT PROGRESS IN AI AND ITS INFLUENCES?

One picture appears in each of the sections that follow.

## SQ2. WHAT ARE THE MOST IMPORTANT ADVANCES IN AI?

In the last five years, the field of AI has made major progress in almost all its standard sub-areas, including vision, speech recognition and generation, natural language processing (understanding and generation), image and video generation, multi-agent systems, planning, decision-making, and integration of vision and motor control for robotics. In addition, breakthrough applications emerged in a variety of domains including games, medical diagnosis, logistics systems, autonomous driving, language translation, and interactive personal assistance. The sections that follow provide examples of many salient developments.

## Underlying Technologies

People are using AI more today to dictate to their phone, get recommendations for shopping, news, or entertainment, enhance their backgrounds on conference calls, and so much more. The core technology behind most of the most visible advances is *machine learning*, especially deep learning (including generative adversarial networks or GANs) and reinforcement learning powered by large-scale data and computing resources. GANs are a major breakthrough, endowing deep networks with the ability to produce artificial content such as fake images that pass for the real thing. GANs consist of two interlocked components—a generator, responsible for creating realistic content, and a discriminator, tasked with distinguishing the output of the generator from naturally occurring content. The two learn from each other, becoming better and better at their respective tasks over time. One of the practical applications can be seen in GAN-based medical-image augmentation, in which artificial images are produced automatically to expand the data set used to train networks for producing diagnoses.[1] Recognition of the remarkable power of deep learning has been steadily growing over the last decade. Recent studies have begun to uncover why and under what conditions deep learning works well.[2] In the past ten years, machine-learning technologies have moved from the academic realm into the real world in a multitude of ways that are both promising and concerning.

---

1 Antreas Antoniou, Amos Storkey, and Harrison Edwards, "Data Augmentation Generative Adversarial Networks," March 2018 https://arxiv.org/abs/1711.04340v3
2 Zeyuan Allen-Zhu, Yuanzhi Li, and Zhao Song, "A Convergence Theory for Deep Learning via Over-Parameterization," June 2019 https://arxiv.org/abs/1811.03962v5; Chiyuan Zhang, Samy Bengio, Moritz Hardt, Benjamin Recht, and Oriol Vinyals, "Understanding deep learning requires rethinking generalization," February 2017 https://arxiv.org/abs/1611.03530v2

## Language Processing

Language processing technology made a major leap in the last five years, leading to the development of network architectures with enhanced capability to learn from complex and context-sensitive data. These advances have been supported by ever-increasing data resources and computing power.

Of particular note are neural network language models, including ELMo, GPT, mT5, and BERT.[3] These models learn about how words are used in context—including elements of grammar, meaning, and basic facts about the world—from sifting through the patterns in naturally occurring text. They consist of billions of tunable parameters and are engineered to be able to process unprecedented quantities of data (over one trillion words for GPT-3, for example). By stringing together likely sequences of words, several of these models can generate passages of text that are often indistinguishable from human-generated text, including news stories, poems, fiction and even computer code. Performance on question-answering benchmarks (large quizzes with questions like "Where was Beyoncé born?") have reached superhuman levels,[4] although the models that achieve this level of proficiency exploit spurious correlations in the benchmarks and exhibit a level of competence on naturally occurring questions that is still well below that of human beings.

These models' facility with language is already supporting applications such as machine translation, text classification, speech recognition, writing aids, and chatbots. Future applications could include improving human-AI interactions across diverse languages and situations. Current challenges include how to obtain quality data for languages used by smaller populations, and how to detect and remove biases in their behavior. In addition, it is worth noting that the models themselves do not exhibit deep understanding of the texts that they process, fundamentally limiting their utility in many sensitive applications. Part of the art of using these models, to date, is in finding scenarios where their incomplete mastery still provides value.

Related to language processing is the tremendous growth in conversational interfaces over the past five years. The near ubiquity of voice-control systems like Google Assistant, Siri, and Alexa is a consequence of both improvements on the voice-recognition side, powered by the AI advances discussed above, and also improvements in how information is organized and integrated for voice-based delivery. Google Duplex, a conversational interface that can call businesses to make restaurant reservations and appointments, was rolled out in 2018 and received mixed initial reviews due to its impressive engineering but off-putting system design.[5]

## Computer Vision and Image Processing

Image-processing technology is now widespread, finding uses ranging from video-conference backgrounds to the photo-realistic images known as deepfakes. Many image-processing approaches use deep learning for recognition, classification, conversion, and other tasks. Training time for image processing has been substantially reduced. Programs running on ImageNet, a massive standardized collection of over 14 million photographs used to train and test visual identification programs, complete their work 100 times faster than just three years ago.[6]

Real-time object-detection systems such as YOLO

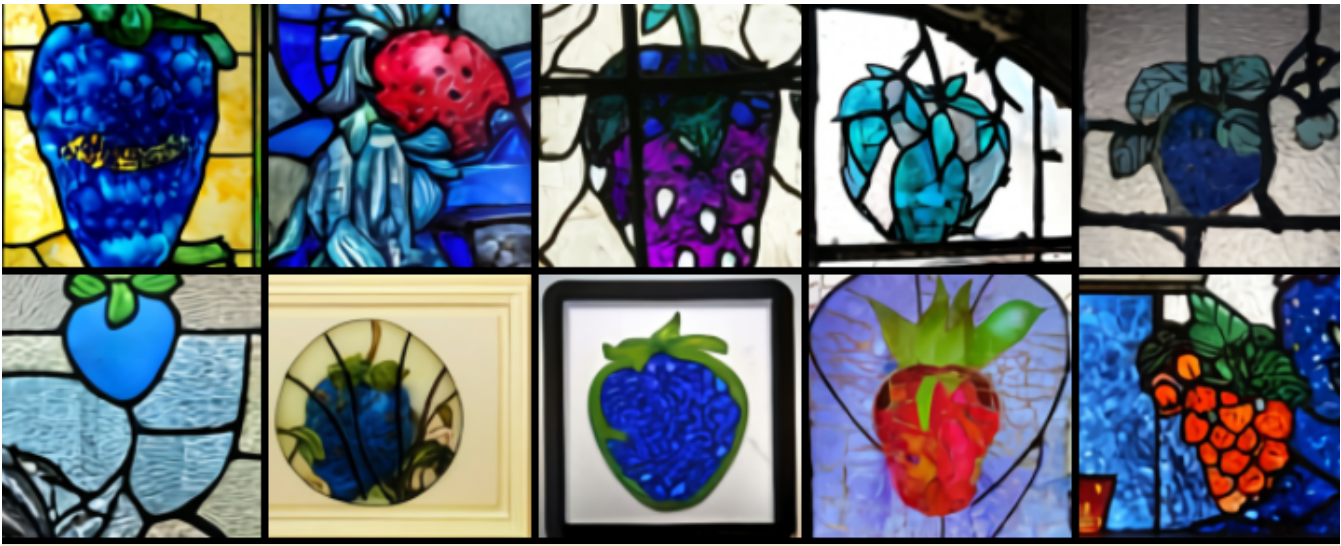

*The GAN technology for generating images and the transformer technology for producing text can be integrated in various ways. These images were produced by OpenAI's "DALL-E" given the prompt: "a stained glass window with an image of a blue strawberry." A similar query to a web-based image search produces blue strawberries, blue stained-glass windows, or stained-glass windows with red strawberries, suggesting that the system is not merely retrieving relevant images but producing novel combinations of visual features. From:* [https://openai.com/blog/dall-e/](https://openai.com/blog/dall-e/)

(You Only Look Once) that notice important objects when they appear in an image are widely used for video surveillance of crowds and are important for mobile robots including self-driving cars. Face-recognition technology has also improved significantly over the last five years, and now some smartphones and even office buildings rely on it to control access. In China, facial-recognition technology ➤ SEE SQ6.C is used widely in society, from security to payment, although there are very recent moves to pull back on the broad deployment of this technology.[7] Of course, while facial-recognition technology can be a powerful tool to improve efficiency and safety, it raises issues around bias and privacy. Some companies have suspended providing face-recognition services. And, in fact, the creator of YOLO has said that he no longer works on the technology because "the military applications and privacy concerns became impossible to ignore."[8]

It is now possible to generate photorealistic images and even videos using GANs. Sophisticated image-processing systems enhanced by deep learning let users seamlessly replace existing images with new ones, such as inserting someone into a video of an event they did not attend. While such modifications could be carried out by skilled artists decades ago, AI automation has substantially lowered the barriers. These so-called *deepfakes* are being used in illicit activity such as "revenge porn," in which an attacker creates artificial sexual content featuring a specific victim, and identity theft, in which a profile of a non-existent person is generated and used to gain access to services, and have spurred research into improving automatic detection of deepfake images.

## Games
➤ SQ2.C

Developing algorithms for games and simulations in adversarial situations has long been a fertile training ground and a showcase for the advancement of AI techniques. DeepMind's application of deep networks to Atari video games and the game of Go around 2015 helped bring deep learning to wide public prominence, and the last five years have seen significant additional progress. AI agents have now out-maneuvered their human counterparts in combat and multiplayer situations

---

including the games StarCraft II[9], Quake III[10], and Alpha Dogfight[11]—a US Defense Department-sponsored jet-fighter simulation—as well as classical games like poker.[12]

The DeepMind team that developed AlphaGo went on to create AlphaGoZero,[13] which discarded the use of direct human guidance in the form of a large collection of data from past Go matches. Instead, it developed moves and tactics on its own, starting from scratch. This idea was further augmented with AlphaZero,[14] a single network architecture that could learn to play expert-level Go, Shogi, or Chess.

## Robotics

The last five years have seen consistent progress in intelligent robotics driven by machine learning, powerful computing and communication capabilities, and increased availability of sophisticated sensor systems. Although these systems are not fully able to take advantage of all the advances in AI, primarily due to the physical constraints of the environments, highly agile and dynamic robotics systems are now available for home and industrial use. In industrial robotics, with the implementation of deep-learning-based vision systems, manipulator-type robots—those that grab things, as opposed to those that roll across the floor—can pick up randomly placed overlapping objects at speeds that are practical for real-world applications.

Bipedal and four-legged robots continue to advance in agility. Atlas, a state-of-the-art humanoid robot built by Boston Dynamics, demonstrated the ability to jump, run, backflip, and maneuver uneven terrain—feats that were impossible for robots just a few years ago. Spot, a quadruped robot also from Boston Dynamics,[15] also maneuvers through difficult environments and is being used on construction sites for delivery and monitoring of lightweight materials and tools. It is worth noting, however, that these systems are built using a combination of learning techniques honed in the last several years, classical control theory akin to that used in autopilots, and painstaking engineering and design. Cassie, a biped robot developed by Agility Robotics and Oregon State University, uses deep reinforcement learning for its walking and running behaviors.[16] Whereas deployment of AI in user-facing vision and language technologies is now commonplace, the majority of types of robotics systems remain lab-bound.

During 2020, robotics development was driven in part by the need to support social distancing during the COVID-19 pandemic. A group of restaurants opened in China staffed by a team of 20 robots to help cook and serve food.[17] Some early delivery robots were deployed on controlled campuses[18] to carry books and food. A diverse collection of companies worldwide are actively pursuing business opportunities in autonomous delivery systems for the last mile. While these types of robots are being increasingly used in the real world, they are by no means mainstream yet and are still prone to mistakes, especially when deployed in unmapped or novel environments. In Japan, a new legal framework is being discussed to ensure that autonomous robotics systems are able to be safely deployed on public roads at limited speeds.[19]

The combination of deep learning with agile robotics is opening up new opportunities in industrial robotics as well. Leveraging improvements in vision, robotic grippers are beginning to be able to select and pick randomly placed objects and use them to construct stacks. Being able to pick up and put down diverse objects is a key competence in a variety of potential applications, from tidying up homes to preparing packages for shipping.

## Mobility

➤ SQ2.E

Autonomous vehicles or self-driving cars have been one of the hottest areas in deployed robotics, as they impact the entire automobile industry as well as city planning. The design of self-driving cars requires integration of a range of technologies including sensor fusion, AI planning and decision-making, vehicle dynamics prediction, on-the-fly rerouting, inter-vehicle communication, and more. Driver assist systems are increasingly widespread in production vehicles.[20] These systems use sensors and AI-based analysis to carry out tasks such as adaptive cruise control to safely adjust speed, and lane-keeping assistance to keep vehicles centered on the road.

The optimistic predictions from five years ago of rapid progress in fully autonomous driving have failed to materialize. The reasons may be complicated,[21] but the need for exceptional levels of safety in complex physical environments makes the problem more challenging, and more expensive, to solve than had been anticipated. Nevertheless, autonomous vehicles are now operating in certain locales such as Phoenix, Arizona, where driving and weather conditions are particularly benign, and outside Beijing, where 5G connectivity allows remote drivers to take over if needed.[22]

## Health

➤ SQ2.F

AI is increasingly being used in biomedical applications, particularly in diagnosis, drug discovery, and basic life science research. ➤ SEE SQ9.D

Recent years have seen AI-based imaging technologies move from an academic pursuit to commercial projects.[23] Tools now exist for identifying a variety of eye and skin disorders,[24] detecting cancers,[25] and supporting measurements needed for clinical diagnosis.[26] Some of these systems rival the diagnostic abilities of expert pathologists and radiologists, and can help alleviate tedious tasks (for example, counting the number of cells dividing in cancer tissue). In other domains, however, the use of automated systems raises significant ethical concerns.[27]

AI-based risk scoring in healthcare is also becoming more common. Predictors of health deterioration are now integrated into major health record platforms (for example, EPIC Deterioration Index), and individual health centers are increasingly integrating AI-based risk predictions into their operations.[28] Although some amount of bias ➤ SEE SQ10.E is evident in these systems,[29] they appear exceptionally promising for overall improvements in healthcare.

Beyond treatment support, AI now augments a number of other health operations and measurements, such as helping predict durations of surgeries to optimize scheduling, and identifying patients at risk of needing transfer to intensive care.[30] There are technologies for digital medical transcription,[31] for reading ECG systems, for producing super-resolution images to reduce the amount of time patients are in MRI machines, and for identifying questions for clinicians to ask pediatric patients.[32] While current penetration is relatively low, we can expect to see uses of AI expand in this domain in the future; in many cases, these are applications of already-mature technologies in other areas of operations making their way into healthcare.

> Apart from a general trend toward more online activity and commerce, the AI technologies powering recommender systems have changed considerably in the past five years.

## Finance

AI has been increasingly adopted into finance. Deep learning models now partially automate lending decisions for several lenders[33] and have transformed payments with credit scoring, for example WeChat Pay.[34] These new systems often take advantage of consumer data that are not traditionally used in credit scoring. In some cases, this approach can open up credit to new groups of people; in others, it can be used to force people to adopt specific social behaviors.[35]

High-frequency trading relies on a combination of models as well as the ability to make fast decisions. In the space of personal finance, so-called robo-advising—automated financial advice—is quickly becoming mainstream for investment and overall financial planning.[36] For financial institutions, uses of AI are going beyond detecting fraud and enhancing cybersecurity to automating legal and compliance documentation as well as detecting money laundering.[37] Government Pension Investment Fund (GPIF) of Japan, the world's largest pension fund, introduced a deep-learning-based system to monitor investment styles of contracting fund managers and identify risk from unexpected change in market situations known as *regime switch*[38]. Such applications enable financial institutions to recognize otherwise invisible risks, contributing to more robust and stable asset-management practices.

## Recommender Systems

With the explosion of information available to us, recommender systems that automatically prioritize what we see when we are online have become absolutely essential. Such systems have always drawn heavily on AI, and now they have a dramatic influence on people's consumption of products, services, and content—from news, to music, to videos, and more. Apart from a general trend toward more online activity and commerce, the AI technologies powering recommender systems have changed considerably in the past five years. One shift is the near-universal incorporation of deep neural networks to better predict user responses to recommendations.[39] There has also been increased usage of sophisticated machine-learning techniques for analyzing the content of recommended items, rather than using only meta-data and user click or consumption behavior. That is, AI systems are making more of an effort to understand why a specific item might be a good recommendation for a particular person or query. Examples include Spotify's use of audio analysis of music[40] or the application of large language models ➤ SEE SQ2.A such as BERT to improve recommendations of news or social media posts.[41] Another trend is modeling and prediction of multiple distinct user behaviors, instead of making recommendations for only one activity at a time; functionality facilitated by the use of so-called multi-task models.[42] Of course, applying recommendation to multiple tasks simultaneously raises the challenging question of how best to make tradeoffs among these different objectives.

---

30 For example, at Vector and St. Michael's Hospital and also using other forms of risk (for example, AlgoAnalyzer, TruScore, OptimaAI).
31 For example, Nuance Dragon, 3M M*Modal, Kara, NoteSwift.
32 For example, Child Health Improvement.
33 https://developer.squareup.com/blog/a-peek-into-machine-learning-at-square/
34 https://fintechnews.hk/12261/fintechchina/how-wechat-pay-determines-if-you-are-trustworthy-with-their-credit-score
35 https://ash.harvard.edu/files/ash/files/disciplining_of_a_society_social_disciplining_and_civilizing_processes_in_contemporary_china.pdf
36 https://www.nerdwallet.com/best/investing/robo-advisors
37 https://www.ibm.com/downloads/cas/DLJ28XP7
38 https://www.gpif.go.jp/en/investment/research_2017_1_en.pdf; https://www.gpif.go.jp/en/investment/ai_report_summary_en.pdf

The use of ever-more-sophisticated machine-learned models for recommending products, services, and (especially) content has raised significant concerns about the issues of fairness ➤ SEE SQ10.E, diversity, polarization, and the emergence of filter bubbles, where the recommender system suggests, for example, news stories that other people like you are reading instead of what is truly most important. While these problems require more than just technical solutions, increasing attention is paid to technologies that can at least partly address such issues. Promising directions include research on the tradeoffs between popularity and diversity of content consumption,[43] and fairness of recommendations among different users and other stakeholders (such as the content providers or creators).[44]

# SQ3. WHAT ARE THE MOST INSPIRING OPEN GRAND CHALLENGE PROBLEMS?

The concept of a "grand challenge" has played a significant role in AI research at least since 1988, when Turing Award winner Raj Reddy, an AI pioneer, gave a speech titled "Foundations and Grand Challenges of Artificial Intelligence."[45] In the address, Reddy outlined the major achievements of the field and posed a set of challenges as a way of articulating the motivations behind research in the field. Some of Reddy's challenges have been fully or significantly solved: a "self-organizing system" that can read a textbook and answer questions;[46] a world-champion chess machine;[47] an accident-avoiding car;[48] a translating telephone.[49] Others have remained open: mathematical discovery,[50] a self-replicating system that enables a small set of machine tools to produce other tools using locally available raw materials.

---

Some of today's grand challenges in AI are carefully defined goals with a clear marker of success, similar to Reddy's chess challenge. The 2016 AI100 report was released just after one such grand challenge was achieved, with DeepMind's AlphaGo beating a world champion in Go ❯ SEE SQ2.C. There are also a number of open grand challenges with less specific criteria for completion, but which inspire AI researchers to achieve needed breakthroughs—such as AlphaFold's ❯ SEE SQ9.D 2020 success at predicting protein structures.

Reddy's grand challenges were framed in terms of concrete tasks to be completed—drive a car, win a game of chess. Similar challenges—such as improving accuracy rates on established datasets like ImageNet[51]—continue to drive creativity and progress in AI research. One of the leading machine-learning conferences, Neural Information Processing Systems, began a "competition track" for setting and solving such challenges in 2017.[52]

❯ SQ3.A   But, as the field of AI has matured, so has the idea of a grand challenge. Perhaps the most inspiring challenge is to build machines that can cooperate ❯ SEE SQ4.A and collaborate seamlessly with humans and can make decisions that are aligned with fluid and complex human values and preferences. This challenge cannot be solved without collaboration between computer scientists with social scientists and humanists. But these are domains in which research challenges are not as crisply defined as a measurable task or benchmark. And indeed, a lesson learned from social science and humanities-inspired research over the past five years is that AI research that is overly tuned to concrete benchmarks and tasks—such as accuracy rates on established datasets—can take us further away from the goal of cooperative and well-

> A lesson learned from social science and humanities-inspired research over the past five years is that AI research that is overly tuned to concrete benchmarks and tasks—such as accuracy rates on established datasets—can take us further away from the goal of cooperative and well-aligned AI that serves humans' needs, goals, and values aligned AI that serves humans' needs, goals, and values.[53] The problems of racial, gender, and other biases ❯ SEE SQ10.C in machine-learning models,[54] for example, can be at least partly attributed to a blind spot created by research that aimed only to improve accuracy on available datasets and did not investigate the representativeness or quality of the data, or the ways in which different errors have different human values and consequences. Mislabeling a car as an airplane is one thing; mislabeling a person as a gorilla is another.[55] So we include in our concept of a grand challenge the open research questions that, like the earlier grand challenges, should inspire a new generation of interdisciplinary AI researchers.

## Turing Test

Alan Turing formulated his original challenge in 1950 in terms of the ability of an interrogator to distinguish between a woman and a machine attempting to mimic a woman, through written question-and-answer exchange.[56] A machine passes the Turing test if it is able to do as good a job as a man at imitating a woman. Today, the challenge is understood to be more demanding (and less sexist): engaging in fluent human text-based conversation requiring a depth of syntactic, cultural, and contextual knowledge so the machine would be mistaken as a human being. Attempts have been made over the years to improve on the basic design.[57] Barbara Grosz, an AI pioneer on the topic of natural communication between people and machines, proposed a modern version of the Turing Test in 2012:

> A computer (agent) team member [that can] behave, over the long term and in uncertain, dynamic environments, in such a way that people on the team will not notice that it is not human.[58]

Has the Turing challenge been solved? Although language models such as GPT-3 ➤ SEE SQ2.A are capable of producing significant amounts of text that are difficult to distinguish from human-generated text,[59] these models are still unreliable and often err by producing language that defies human rules, conventions, and common sense, especially in long passages or an interactive setting. Indeed, major risks still surround these errors. Language-generating chatbots easily violate human norms about acceptable language, producing hateful, racist, or sexist statements in contexts where a socially competent human clearly would not.[60]

Today's version of the Turing challenge should also take into consideration the very real harms that come from building machines that trick humans into believing they are interacting with other humans. The initial roll out of Google Duplex ➤ SEE SQ2.B generated significant public outcry because the system uses an extremely natural voice and injects umms and ahs when booking an appointment; it looked as though it was trying to fool people. (The current version discloses its computer identity.)[61] With the enormous advances in the capacity for machine learning to produce images, video, audio, and text that are indistinguishable from human-generated versions have come significant challenges to the quality and stability of human relationships and systems. AI-mediated content on social media platforms, for example, has contributed in the last few years to political unrest and violence.[62]

A contemporary version of the Turing challenge might therefore be the creation of a machine that can engage in fluent communication with a human *without* being mistaken for a human, especially because people adapt so readily to human-like conversational interaction.[63] Grosz's version of the test recognizes the importance of this concern: It "does not ask that the computer system act like a person or be mistaken for one. Instead it asks that the computer's nonhumanness not hit one in the face, that it is not noticeable, and that the computer act intelligently enough that it does not baffle its teammates." This approach would be consistent with principles that are emerging in regulation, such as the European Union's 2021 proposal for legislation requiring that providers of AI systems design and develop mechanisms to inform people that they are interacting with AI technology.[64]

---

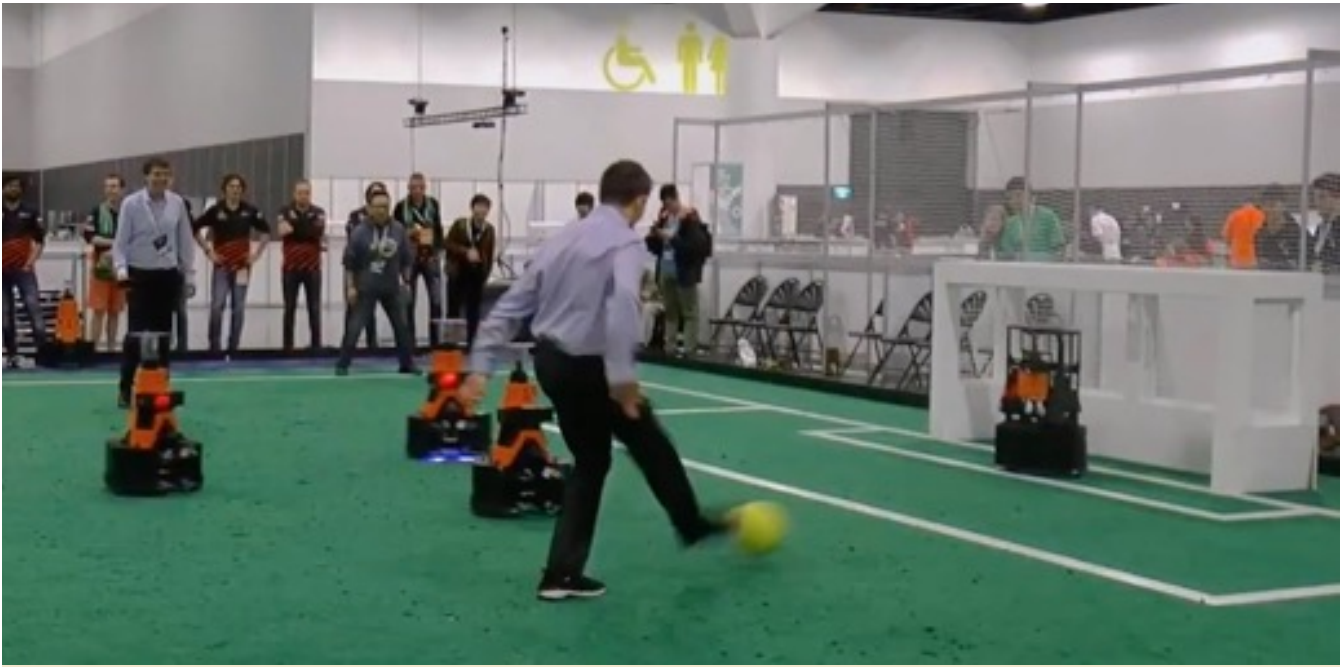

*Ball control, passing strategy, and shooting accuracy have continued to improve over the quarter century the RoboCup competition has been held. While still dominated by human players, even in their researcher clothes, the best robot teams can occasionally score in the yearly human-robot match. Peter Stone, the AI-100 Standing Committee chair, is shown here taking a shot in the RoboCup 2019 match in Sydney, managed by ICMS Australasia. From: https://spectrum.ieee.org/automaton/robotics/robotics-hardware/watch-world-champion-soccer-robots-take-on-humans-at-robocup*

## RoboCup

RoboCup is an established grand challenge in AI and robotics with the goal of developing a fully autonomous robot team capable of beating the FIFA World Cup champion soccer (football) team by 2050. Researchers from over 35 countries are involved in this initiative, with a series of international and regional competitions, symposia, summer schools, and other activities. While RoboCup's main goal is to develop a super-human team of robots, an alternative goal is to form a human-robot hybrid championship team. This alternative goal stresses human-robot collaboration, fostering symbiotic human-robot relationships.

Since 2007, RoboCup has moved toward trials of robots playing soccer on an outdoor field, and has matched a winning robot team against human players indoors. While the level of play remains far from real-world soccer, these steps constitute major progress toward more realistic play. RoboCup has also introduced and fostered novel competitions for intelligent robotics including home-based, industrial, and disaster-response robotics.

## International Math Olympiad

The International Math Olympiad (IMO) is an international mathematics competition for high-school students. Related to Reddy's challenge in mathematical discovery is to build an AI system that can win a gold medal in the IMO. The committee sponsoring this challenge has set precise parameters for success: The AI must be capable of producing, with the same time limit as a human contestant, solutions to the problems in the annual IMO that can be checked by an automated theorem prover in 10 minutes (the time it usually takes a human judge to evaluate a human's solution) and achieving a score that would have earned a gold medal in a given year.[65]

---

## The AI Scientist

The AI Scientist challenge[66] envisions the development, by 2050, of AI systems that can engage in autonomous scientific research. They would be "capable of making major discoveries some of which are at the level worthy of the Nobel Prize or other relevant recognition" and "make strategic choices about research goals, design protocols and experiments to collect data, notice and characterize a significant discovery, communicate in the form of publications and other scientific means to explain the innovation and methods behind the discovery and articulate the significance of the discovery [and] its impact." A workshop organized by the Alan Turing Institute in February 2020 proposed the creation of a global initiative to develop such an AI Scientist.[67]

## Broader Challenges

We now turn to open grand challenges that do not have the structure of a formal competition or crisp benchmark. These research challenges are among the most inspiring.

### GENERALIZABILITY

Modern machine-learning models are trained on increasingly massive datasets (over one trillion words for GPT-3, ➤ SEE SQ2.A for example) and optimized to accomplish specific tasks or maximize specified reward functions. While these methods enable surprisingly powerful systems, and performance appears to follow a power law—showing increases that continue to grow with increasing data sets or model size[68]—many believe that major advances in AI will require developing the capacity for generalizing or transferring learning from a training task to a novel one. Although modern machine-learning techniques are making headway on this problem, a robust capacity for generalizability and transfer learning ➤ SEE SQ5.B will likely require the integration of symbolic

> Increasing generality is likely to require that machines learn from small samples and by analogy. Generalizability is a key component of robustness, allowing an AI system to respond and adapt to shifts in the frequency with which it sees different examples, something that continues to interfere with modern machine-learning-based systems.

and probabilistic reasoning—combining a primary focus on logic ➤ SEE SQ12.B with a more statistical point of view. Some characterize the skill of extrapolating from few examples as a form of common sense, meaning that it requires broad knowledge about the world and the ability to adapt that knowledge in novel circumstances. Increasing generality is likely to require that machines learn, as humans do, from small samples and by analogy. Generalizability is also a key component of robustness, allowing an AI system to respond and adapt to shifts in the frequency with which it sees different examples, something that continues to interfere with modern machine-learning-based systems.

 CAUSALITY

An important source of generality in natural intelligence is knowledge of cause and effect. Current machine-learning techniques are capable of discovering hidden patterns in data, and these discoveries allow the systems to solve ever-increasing varieties of problems. Neural network language models, **> SEE SQ2.A** for example, built on the capacity to predict words in sequence, display tremendous capacity to correct grammar, answer natural language questions, write computer code, translate languages, and summarize complex or extended specialized texts. Today's machine-learning models, however, have only limited capacity to discover causal knowledge of the world, as Turing award winner Judea Pearl has emphasized.[69] They have very limited ability to predict how novel interventions might change the world they are interacting with, or how an environment might have evolved differently under different conditions. They do not know what is possible in the world. To create systems significantly more powerful than those in use today, we will need to teach them to understand causal relationships. It remains an open question whether we'll be able to build systems with good causal models of sufficiently complex systems from text alone, in the absence of interaction.

NORMATIVITY

Nils J. Nilsson,[70] one of AI's pioneers and an author of an early textbook in the field, defined intelligence as the capacity to function appropriately and with foresight in an environment. When that environment includes humans, appropriate behavior is determined by complex and dynamic normative schemes. Norms govern almost everything we do; whenever we make a decision, we are aware of whether others would consider it "acceptable" or "not acceptable."[71] And humans have complex processes for choosing norms with their own dynamics and characteristics.[72] Normatively competent AI systems will need to understand and adapt to dynamic and complex regimes of normativity.

Aligning with human normative systems is a massive challenge in part because what is "good" and what is "bad" varies tremendously across human cultures, settings, and time. Even apparently universal norms such as "do not kill" are highly variable and nuanced: Some modern societies say it is okay for the state to kill someone who has killed another or revealed state secrets; historically, many societies approved of killing a woman who has had pre-marital or extra-marital sex or whose family has not paid dowry, and some groups continue to sanction such killing today. And most killing does not occur in deliberate, intentional contexts. Highways and automobiles are designed to trade off speed and traffic flow with a known risk that a non-zero number of people will be killed by the design. AI researchers can choose not to participate in the building of systems that violate the researcher's own values, by refusing to work on AI that supports state surveillance or military applications, say. But a lesson from the social sciences and humanities is that it is naive to think that there is a definable and core set of universal values that can directly be built into AI systems. Moreover, a core value that *is* widely shared is the concept of group self-determination and national sovereignty. AI systems built for Western values, with Western tradeoffs, violate other values.

Even within a given shared normative framework, the **> SQ3.E** challenges are daunting. As an example, there has been an explosion of interest in the last five years in the problem of developing algorithms that are unbiased and fair.[73] Given the marked cultural differences in what is even considered "fair," doing this will require going beyond the imposition of statistical constraints on outputs of AI systems. Like a competent human, advanced AI systems will need to be able to both read and interact with cultural and social norms, and sometimes highly local practices, rules, and laws, and to adapt as these features of the normative environment change. At the same time, AI systems will need to have features that allow them to be integrated into the institutions through which humans implement normative systems. For an AI system to be accountable, for example, it will require that accounts of how and why it acted as it did are reviewable by independent third parties tasked with ensuring that the account is consistent with applicable rules.

Humans who are alleged to have engaged in unlawful conduct are held accountable by independent adjudicators applying consistent rules and procedures. For AI to be ethical, fair and value-aligned, it needs to have good normative models and to be capable of integrating its behavior into human normative institutions and processes. Although significant progress is being made on making AI more explainable[74]—and avoiding opaque models in high-stakes settings when possible[75]—systems of accountability require more than causal accounts of how a decision was reached; they require normative accounts of how and why the decision is consistent with human values. Explanation is an interaction between a machine and human; *justification* is an interaction between a machine and an entire normative community and its institutions.

# SQ4. HOW MUCH HAVE WE PROGRESSED IN UNDERSTANDING THE KEY MYSTERIES OF HUMAN INTELLIGENCE?

AI, the study of how to build an intelligent machine, and cognitive science, the study of human intelligence, have been evolving in complementary ways. A view of human intelligence that has gained prominence over the last five years holds that it is *collective*—that individuals are just one cog in a larger intellectual machine. Their roles in that collective are likely to be different than the roles of machines, because their strengths are different. Humans are able to share intentionality with other humans—to pursue common goals as a team—and doing so may require having features that are uniquely human: a certain kind of biological hardware with its associated needs and a certain kind of cultural experience. In contrast, machines have vast storehouses of data, along with the ability to process it and communicate with other machines at remarkable speeds. In that sense, AI is developing in ways that improve its ability to collaborate with and support people, rather than in ways that mimic human intelligence. Still, there are remarkable parallels between the operation of individual human minds and that of deep learning machines. AI seems to be evolving in a way that adopts some core human features—specifically, those that relate to perception and memory.

In the early days of AI, a traditional view of human intelligence understood it as a distinct capacity of human cognition, related to a person's ability to process information. Intelligence was a property that could be measured in any sufficiently complex cognitive system, and individuals differed in their mental horsepower. Today, this view is rare among cognitive scientists and others who study human intelligence. The key mysteries of human intelligence that have come to concern researchers for more than a decade include questions not only about how people are able to interpret complex inputs, solve difficult problems, and make reasonable judgments and decisions quickly, but also how we are able to negotiate emotionally nuanced relationships, use attitudes and emotions and other bodily signals to guide our decision-making, and understand other people's intentions.[76] The study of intelligence has become the study of how people are able to adapt and succeed, not just how an impressive information-processing system works.

The modern study of human intelligence draws on a variety of forms of evidence. *Cognitive psychology* uses experimental studies of cognitive performance to look at the nature of human cognition and its capabilities. *Collective intelligence* is the study of how intelligence is designed for and emerges from group (rather than individual) activity. *Psychometrics* is the study of how people vary in what they can do, how their capabilities are determined, and how abilities relate to demographic variables. *Cognitive neuroscience* looks at how the brain's hardware is involved in implementing psychological and social processes. In the context of cognitive science, *artificial intelligence* is concerned with how advances in automating skills associated with humans provide proofs-of-concept about how humans might go about doing the same things.

Developments in human-intelligence research in the last five years have been inspired more by collective intelligence,[77] cognitive neuroscience,[78] and artificial intelligence[79] than by cognitive psychology or psychometrics. The study of working memory, attention, and executive processing in cognitive psychology, once understood as the mental components supporting

> Intelligence is a property not only of individuals, but also of collectives. Deliberating groups can also exhibit greater intelligence than individuals, especially when they follow norms that encourage challenge and constructive criticism.

intelligence, have become central topics in the study of cognitive neuroscience.[80] Psychometric work on intelligence itself has splintered, due to the recognition that a single "intelligence" dimension like IQ does not adequately characterize human problem-solving potential.[81] Abilities like empathy, impulse control, and storytelling turn out to be just as important. Over the past half decade, major shifts in the understanding of human intelligence have favored the topics discussed below.

## Collective Intelligence

Research from a variety of fields reinforces the view that intelligence is a property not only of individuals, but also of collectives.[82] As we know from work on the wisdom of crowds,[83] collectives can be surprisingly insightful, especially when many individuals with relevant knowledge make independent contributions, unaffected by pressures to conform to group norms. Deliberating groups can also exhibit greater intelligence than individuals, especially when they follow norms that encourage challenge and constructive criticism.

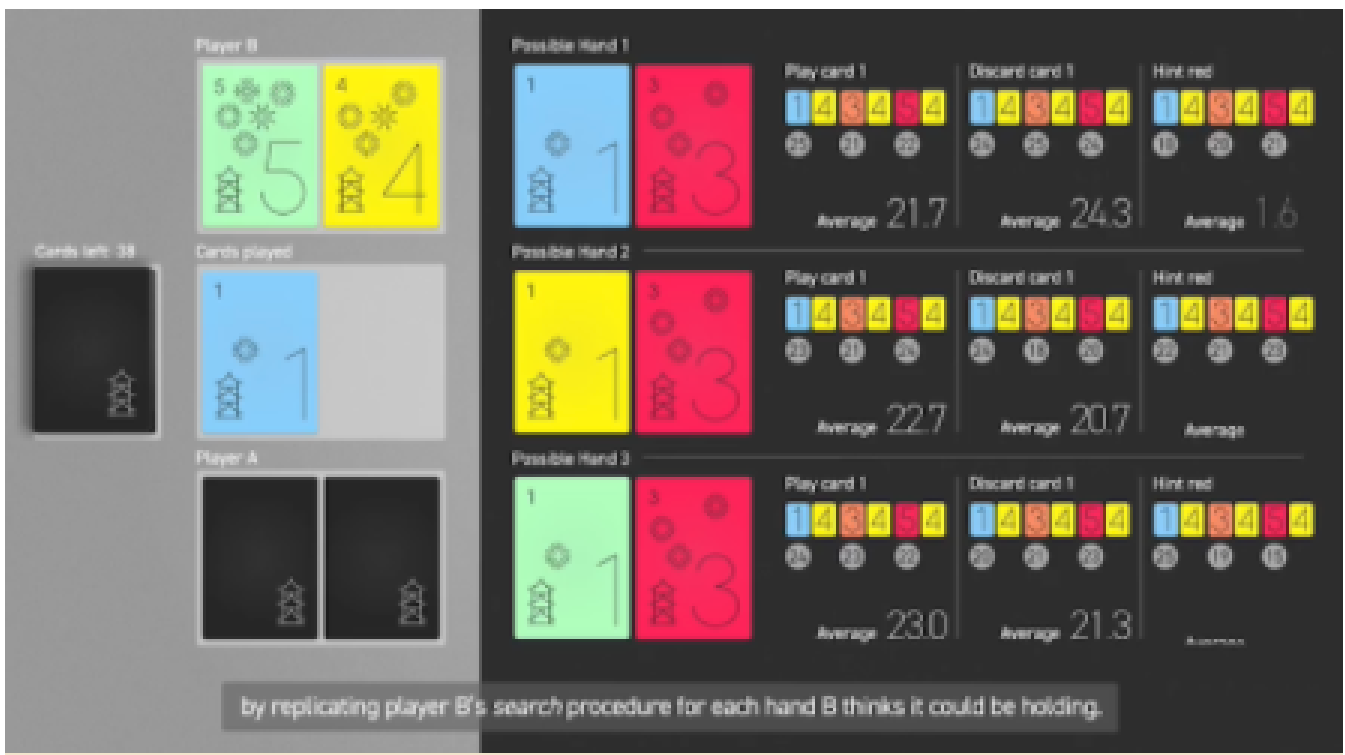

Intelligence is a group property in the sense that the quality of a group's performance does not depend on the IQs of the individual members of the group.[84] It is easier to predict group performance if you know how good the group is at turn-taking or how empathetic the members are than if you know the IQs of group members. Research on children shows they are sensitive to what others know when deciding whose advice to take.[85]

Studies of social networks have shown the role of collective intelligence in determining individual beliefs. Some of those studies help explain the distribution of beliefs across society, showing that patterns of message transmission in social networks can account for both broad acceptance of beliefs endorsed by science and simultaneous minority acceptance of conspiracy theories.[86] Such studies also offer a window into political polarization by showing that even a collection of rational decision-makers can end up splitting into incompatible subgroups under the influence of information bubbles.[87]

In the most general sense, the research community is starting to see the mind as a collective entity spread across members of a group. People obviously have skills that they engage in as individuals, but the majority of knowledge that allows them to operate day by day sits in the heads of other members of their community.[88] Our sense of understanding is affected by what others know, and we rely on others for the arguments that constitute our explanations, often without knowing that we are doing so. For instance, we might believe we understand the motivation for a health policy (wear a mask in public!) but actually we rely on experts or the internet to spell it out. We suppose we understand how everyday objects like toilets work—and discover our ignorance when we try to explain their mechanism,[89] or when they break. At a broader level, our communities determine our political and social beliefs and attitudes. Political partisanship influences many beliefs and actions,[90] some that have nothing to do with politics,[91] even some related to life and death.[92]

## Cognitive Neuroscience

Work in cognitive neuroscience has started to productively examine a variety of higher-level skills associated with the more traditional view of intelligence. Three partially competing ideas have reached some consensus over the last few years.

First, a pillar of cognitive neuroscience is that properties of individuals such as working memory and executive control are central to domain-independent intelligence, that which governs performance on all cognitive tasks regardless of their mode or topic. A common view is that this sort of intelligence is governed by neural speed.[93] But there is increasing recognition that what matters is not global neural speed per se, but the efficiency of higher-order processing. Efficiency is influenced not just by speed, but by how processing is organized.[94]

A second idea gaining support is that higher-ability individuals are characterized by more efficient patterns of brain connectivity. Both of these ideas are consistent with the dominant view that intelligence is associated with higher-level brain areas in the parieto-frontal cortex.

The third idea is more radical. It suggests that the neural correlates of intelligence are distributed throughout the brain.[95] In this view, the paramount feature of human intelligence is flexibility, the ability to continually update prior knowledge and to generate predictions. Intelligence derives from the brain's ability to dynamically generate inferences that anticipate sensory inputs. This flexibility is realized as brain plasticity—the ability to change—housed in neural connections that exhibit what network scientists call a "small-world" pattern, where the brain balances relatively focal, densely interconnected, functional centers with long-range connections that allow for more global integration of information.

Cognitive neuroscience has taken a step in the direction of collective cognition via a sub-discipline called *social neuroscience*. Its motivation is the recognition that one of the brain's unique and most important capacities

> Neural net models have been in the spotlight—due in small part to the success of computational neuroscience and in large part to the success of deep learning in AI.

is its ability to grasp what others are thinking and feeling. The field has thus focused on issues that are old stalwarts of social psychology—fairness, punishment, and people's tendency to cooperate versus compete—and on identifying hormones and brain networks that are involved in these activities. Unlike other branches of cognitive neuroscience, social neuroscience recognizes that human cognitive, emotional, and behavioral processes have been shaped by our social environments.

A corollary of developments in cognitive neuroscience is the growth of the related field of computational neuroscience, which brings a computational perspective to the investigation of brains and minds. This field has been aided tremendously by the machine-learning paradigm known as reinforcement learning, which is concerned with learning from evaluative feedback—reward and punishment.[96] It has proven to be a goldmine of ideas for understanding learning in the brain, since each element of the computational theory can be linked to processes at the cellular level. For instance, there is now broad consensus about the central role of the dopamine system in learning, decision-making, motivation, prediction, motor control, habit, and addiction.[97]

## Computational Modeling

For decades now, trends in computational modeling of cognition have followed a recurring pattern, cycling between a primary focus on logic (symbolic reasoning) and on pattern recognition (neural networks) ➤ SEE SQ12.B. In the past five to 10 years, neural net models have been in the spotlight—due in small part to the success of computational neuroscience and in large part to the success of deep learning in AI. The computational modeling field is now full of deep-learning-inspired models of visual recognition, language processing, and other cognitive activities. There remains a fair amount of excitement about Bayesian modeling—a type of logic infused with probabilities. But the clash with deep learning techniques has stirred a heated debate. Is it better to make highly accurate predictions without understanding exactly why, or better to make less accurate predictions but with a clear logic behind them?[98] We expect this debate will be further explored in future AI100 reports.

Beyond efforts to build computational models, deep learning models have become central methodological weapons in the cognitive science arsenal. They are the state-of-the-art tools for classification, helping experimentalists to quickly construct large stimulus sets for experiments and analysis. Moreover, huge networks trained on enormous quantities of data, such as GPT-3 and Grover, ➤ SEE SQ2.A have opened new territory for the study of language and discourse at multiple levels.

## The State of the Art

The nature of consciousness remains an open question. Some see progress;[99] others believe we are no further along in understanding how to build a conscious agent than we were 46 years ago, when the philosopher Thomas Nagel famously posed the question, "What is it like to be a bat?"[100] It is not even clear that understanding consciousness is necessary for understanding human intelligence. The question has become less pressing for this purpose as we have begun to recognize the limits of conscious information processing in human cognition,[101] and as our models become increasingly based on emergent processes instead of central design.

Cognitive models motivate an analysis of how people integrate information from multiple modalities, multiple senses, and multiple sources: our brains, our physical bodies, physical objects (pen, paper, computers), and social entities (other people, Wikipedia). Although there is now a lot of evidence that it is the ability to do this integration that supports humanity's more remarkable achievements, how we do so remains largely mysterious. Relatedly, there is increased recognition of the importance of processes that support intentional action, shared intentionality, free will, and agency. But there has been little fundamental progress on building rigorous models of these processes.

The cognitive sciences continue to search for a paradigm for studying human intelligence that will endure. Still, the search is uncovering critical perspectives—like collective cognition—and methodologies that will shape future progress, like cognitive neuroscience and the latest trends in computational modeling. These insights seem essential in our quest for building machines that we would truly judge as "intelligent."

## SQ5: WHAT ARE THE PROSPECTS FOR MORE GENERAL ARTIFICIAL INTELLIGENCE?

Since the dawn of the field, AI research has had two different, though interconnected, goals: *narrow AI*, to develop systems that excel on specific tasks, and *general AI*, to create systems that achieve the flexibility and adaptability of human intelligence. While all of today's state-of-the-art AI applications are examples of narrow AI, many researchers are pursuing more general AI systems, an effort that some in the field have labeled AGI, for artificial general intelligence.

Most successful AI systems developed in the last several years have relied, at least in part, on supervised learning, in which the system is trained on examples that have been labeled by humans. Supervised learning has proven to be very powerful, especially when the learning systems are deep neural networks and the set of training examples is very large.

Reinforcement learning is another framework that has produced impressive AI successes in the last decade. In contrast with supervised learning, reinforcement learning relies not on labeled examples but on "reward signals" received by an agent taking actions in an (often simulated) environment. Deep reinforcement learning, which combines deep neural networks with reinforcement learning, has generated considerable excitement in the AI community following its role in creating AlphaGo, the program that was able to beat the world's best human Go players. ➤ SEE SQ2.C (We will return to AlphaGo in a moment.)

The chair of the committee working for the AI-based organization creates examples.
The chair of the committee working for the AI-based organization **create** examples.
The members of the committee working for the AI-based organization **creates** examples.
The members of the committee working for the AI-based organization create examples.|

*Widely available tools like Google Docs' grammar checker uses transformer-based language models to propose alternative word choices in near-real time. While prior generations of tools could highlight non-words ("I gave **thier** dog a bone"), or even distinguish common word substitutions based on local context ("I gave **there** dog a bone"), the current generation can make recommendations based on much more distant or subtle cues. Here, the underlined word influences which word is flagged as a problem from 9 words away. Image credit: Michael Littman via https://docs.google.com/.*

A third subfield of AI that has generated substantial recent interest is probabilistic program induction,[102] in which learning a concept or skill is accomplished by using a model based on probabilities to generate a computer program that captures the concept or performs the skill. Like supervised learning and reinforcement learning, most probabilistic program induction methods to date have fit squarely into the "narrow AI" category, in that they require significant human engineering of specialized programming languages and produce task-specific programs that can't easily be generalized.

While these and other machine-learning methods are still far from producing fully general AI systems, in the last few years important progress has been made toward making AI systems more general. In particular, progress is underway on three types of related capabilities. First is the ability for a system to learn in a self-supervised or self-motivated way. Second is the ability for a single AI system to learn in a continual way to solve problems from many different domains without requiring extensive retraining for each. Third is the ability for an AI system to generalize between tasks—that is, to adapt the knowledge and skills the system has acquired for one task to new situations, with little or no additional training.

## Self-Supervised Learning With the Transformer Architecture

➤ SQ5.A

Significant progress has been made in the last five years on self-supervised learning, a step towards reducing the problem of reliance on large human-labeled training sets. In self-supervised learning, a learning system's input can be an incomplete example, and the system's job is to complete the example correctly. For instance, given the partial sentence "I really enjoyed reading your…," one might predict that the final word is "book" or "article," rather than "coffee" or "bicycle." Systems trained in this way, which output probabilities of possible missing words, are examples of neural network language models. No explicit human-created labels are needed for self-supervised learning because the input data itself plays the role of the training feedback.

Such self-supervised training methods have been particularly successful when used in conjunction with a new architecture for deep neural networks called the transformer.[103] At its most basic level, a transformer is a neural network optimized for processing sequences with long-range dependencies (for example, words far apart in a sentence that depend on one another), using the idea of "attention" weights to focus processing on the most relevant parts of the data.

Transformer-based language models have become the go-to approach for natural language processing, and have been used in diverse applications, including machine translation and Google web search. They can also generate ➤ SEE SQ2.A convincingly human-like text ➤ SEE SQ9.A.

Transformers trained with self-supervised learning are a promising tool for creating more general AI systems, because they are applicable to or easily integrated with a wide variety of data—text, images, even protein-folding structures[104]—and, once trained, they can either immediately or with just a small amount of retraining known as "fine-tuning" achieve state-of-the-art narrow AI performance on difficult tasks.

➤ SQ5.B
## Continual and Multitask Learning

Significant advances have been made over the last several years in AI systems that can learn across multiple tasks while avoiding the pervasive problem of catastrophic interference between the tasks, in which training the system on new tasks causes it to forget how to perform tasks it has already learned. Much of this progress has come about due to advances in *meta-learning* methods.

Meta-learning refers to machine-learning methods aimed at improving the machine-learning process itself. One influential approach is to train a deep neural network on a variety of tasks, where the objective is for the network to learn general-purpose, transferable representations, as opposed to representations tailored specifically to any particular task.[105] The learned representations are such that a neural network trained in this way could be fine-tuned for a variety of specific tasks with only a small number of training examples for a given task. Meta-learning has also led to progress in probabilistic program induction, by enabling abstraction strategies that learn general-purpose program modules that can be configured for many different tasks.[106]

In *continual learning*, a learning agent is trained on a sequence of tasks, each of which is seen only once. The challenges of continual learning are to constantly use what has been learned in the past to apply to new tasks, and, in learning new tasks, to avoid destroying what has already been learned. While continual learning, like meta-learning, has been researched for decades in the machine-learning community, the last several years have seen some significant advances in this area. Examples of new approaches include training systems that mimic processes in the brain, known as neuromodulatory processes, to learn gating functions that turn on and off network areas to enable continual learning without forgetting.[107]

## Making Deep Reinforcement Learning More General

For decades, the game of Go ➤ SEE SQ2.C has been one of AI's grand challenge problems, one much harder than chess due to Go's vastly larger space of possible board configurations, legal moves, and strategic depth. In 2016, DeepMind's program AlphaGo definitively conquered that challenge, defeating Lee Sedol, one of the world's best human Go players, in four out of five games. AlphaGo learned to play Go via a combination of several AI methods, including supervised deep learning, deep reinforcement learning, and an iterative procedure for exploring possible lines of play called Monte Carlo tree search.[108] While AlphaGo was a landmark in AI history, it remains a triumph of narrow AI, since the trained program was only able to perform a single task: playing Go. Later developments in the AlphaGo line of research have drastically reduced the reliance on example games played by humans, Go-specific representations, and even access to the rules of the game in advance. Nevertheless, the learned strategies are thoroughly game-specific. That is, the methodology for producing the Go player was general, but the Go player was not.

In the last several years, much research has gone into making deep-reinforcement-learning methods more general. A key part of reinforcement learning is the definition of reward signals in the environment. In AlphaGo, the only reward signal was winning or losing the game. However, in real-world domains, a richer set of reward signals may be necessary for reinforcement-learning algorithms to succeed. These reward signals are usually defined by a human programmer and are specific to a particular task domain. The notion of *intrinsic motivation* for a learning agent refers to reward signals that are intended to be general—that is, useful in any domain. Intrinsic motivation in AI is usually defined in terms of seeking novelty: an agent is rewarded for exploring new areas of the problem space, or for being wrong in a prediction (and thus learning something new). The use of intrinsic motivation has a long history in reinforcement learning,[109] but in the last few years it has been used as a strategy for more general reinforcement-learning systems designed to perform multitask learning or continual learning where the same

learning system is trained to solve multiple problems.[110]

Another set of advances in reinforcement learning is in synthesizing representations of generative world models—models of an agent's environment that can be used to simulate "imagined" scenarios, in which an agent can test policies and learn without being subject to rewards or punishments in its actual environment. Such models can be used to generate increasingly complex or challenging scenarios to allow learning to be scaffolded via a useful "curriculum." Using deep neural networks to learn and then generate such models has resulted in progress in reinforcement learning's generality and speed of learning.[111]

## Common Sense

These recent approaches attempt to make AI systems more general by enabling them to learn from a small number of examples, learn multiple tasks in a continual way without inter-task interference, and learn in a self-supervised or intrinsically motivated way. While these approaches have shown promise on several restricted domains, such as learning to play a variety of video games, they are still only early steps in the pursuit of general AI. Further research is needed to demonstrate that these methods can scale to the more diverse and complex problems the real world has to offer.

An important missing ingredient, long sought in the AI community, is common sense. The informal notion of common sense includes several key components of general intelligence that humans mostly take for granted, including a vast amount of mostly unconscious knowledge about the world, an understanding of causality (what factors cause events to happen or entities to have certain properties) ❯ SEE SQ3.D, and an ability to perceive abstract similarities between situations—that is, to make analogies.[112] Recent years have seen substantial new research, especially in the machine-learning community, in how to imbue machines with common sense abilities.[113] This effort includes work on enabling machines to learn causal models[114] and intuitive physics,[115] describing our everyday experience of how objects move and interact, as well as to give them abilities for abstraction and analogy.[116]

AI systems still remain very far from human abilities in all these areas, and perhaps will never gain common sense or general intelligence without being more tightly coupled to the physical world. But grappling with these issues helps us not only make progress in AI, but better understand our own often invisible human mechanisms of general intelligence.

> Since the first AI100 report, public understanding has broadened and become more nuanced, starting to move beyond Terminator and robot overlord fears.

# SQ6. HOW HAS PUBLIC SENTIMENT TOWARDS AI EVOLVED, AND HOW SHOULD WE INFORM/ EDUCATE THE PUBLIC?

Over the last few years, AI and related topics have gained traction in the zeitgeist. Research has tracked this trend quantitatively: In the 2017–18 session of the US Congress, for instance, mentions of AI-related words were more than ten times higher than in previous sessions. Similarly, web searches for "machine learning" have roughly doubled since 2016.[117]

Since the first AI100 report, public understanding has broadened and become more nuanced, starting to move beyond Terminator and robot overlord fears. Overtaking these concerns for many members of the public are the prospects of social and economic impacts from AI, especially negative impacts such as discriminatory effects, economic inequality, and labor replacement or exploitation, topics discussed extensively in the prior report. In addition, there is a great deal of discussion around the increasing risks of surveillance as well as how AI and social media are involved in manipulation and disinformation. These discussions contribute to growing concerns among researchers, policymakers, and governments about establishing and securing public trust in AI (and in science and technology more broadly). As a result, a wide range of initiatives have focused on the goal of promoting "trustworthy AI."[118]

Public awareness of the benefits of AI skews toward anticipated breakthroughs in fields such as health and transportation, with comparatively less awareness of the *existing* benefits of AI embedded in applications already in widespread use. While popular culture regularly references the AI capabilities of virtual assistants such as Alexa, there is less public awareness of AI's involvement in everyday technologies commonly accessed without the mediation of an artificial agent: benefits such as speech-to-text, language translation, interactive GPS navigation, web search, and spam filtering. Media coverage of AI often distorts and exaggerates its potential at both the positive and negative extremes, but it has helped to raise public awareness of legitimate concerns about AI bias, lack of transparency and accountability, and the potential of AI-driven automation to contribute to rising inequality.

There are notable regional and gender differences in public sentiment about AI, as revealed in a 2020 Pew survey:[119] opinions in Asian countries are largely positive, while those of countries in the West are heavily divided and more skeptical. Men overall expressed far more positive attitudes about AI than women did. Educational differences were also significant; age and political orientation less so. A 2019 survey by the Centre for the Governance of AI[120] at Oxford's Future of Humanity Institute noted that positive attitudes about AI are greater among those who are "wealthy, educated, male, or have experience with technology."

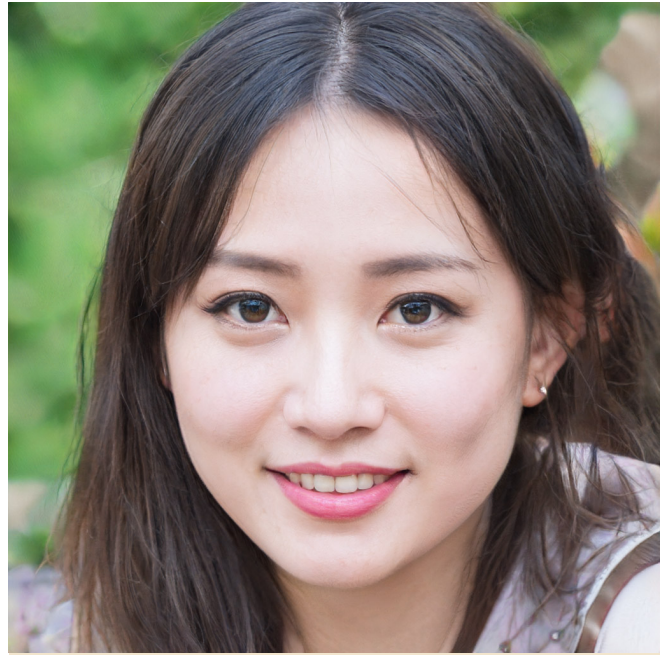

*Neural networks, trained on tens of thousands of portrait photographs of faces, can now generate novel high-resolution images that appear compellingly like pictures of real human faces. The technology behind this development, generative adversarial networks (GANs), has advanced rapidly since its introduction in 2014. Current versions still include telltale visual artifacts, like the strangely absent right shoulder in this image. Nonetheless, the previously unattainable level of realism raises concerns about the use of this technology to spread realistic disinformation. From: https://github.com/NVlabs/stylegan2.*

## Primary Drivers of Public Understanding and Sentiment

Recent media coverage has been heavily focused on the negative impacts of AI, including bias, disinformation, and deepfakes. Coverage in 2020 shifted somewhat to AI's potential for supporting the pandemic response through contact tracing, transmission forecasting, and elder care, and coverage of some notable AI developments such as GPT-3 ➤ **SEE SQ2.A** also spurred public interest. Since the public is not always able to discern which harms, risks, and benefits are assignable to artificial intelligence and machine learning and which emerge from broader technology platform and business-model-use cases ("surveillance capitalism,"[121] ➤ **SEE WQ2.A** for example, the assembly, maintenance, and trade in mass quantities of personal data) or from simpler algorithmic tools that don't involve AI, some concerns may be misplaced.

AI researchers have not been as engaged publicly as they need to be , although there have been attempts to reach a broader audience. An example was a 2017 staged debate between Gary Marcus, NYU psychology professor and author, and Yann LeCun, chief AI scientist at Facebook and Turing Award winner, about how much specialized information we need to build into AI systems ➤ SEE SQ12.A and how much they should learn for themselves.[122] Generally, though, accurate scientific communication has not engaged a sufficiently broad range of publics in gaining a realistic understanding of AI's limitations, strengths, social risks, and benefits, and tends to focus on new methods, applications, and progress toward artificial general intelligence. Given the historical boom/bust pattern in public support for AI,[123] it is important that the AI community not overhype specific approaches or products and create unrealistic expectations.

➤ SQ6.A    Governments, universities, and nonprofits are attempting to broaden the reach of AI education, including investing in new AI-related curricula. Groups such as AI4ALL[124] and AI4K12[125] are receiving increasing attention, supported by a sense that today's students need to be prepared to live in and contribute to an AI-fueled world, as well as by widespread concerns about an overall lack of diversity in the field. At the college level, curricula that include data science and AI/data ethics are becoming more widespread. At the level of the research community, several prominent AI conferences now require that research papers include explicit broader impact statements that discuss positive and negative societal consequences of the work.[126]

Rhetoric surrounding an AI "race" between the US and China has framed investment and understanding

> Technological determinism is the false claim that new technologies such as AI shape society independently of human choices and values, in a manner that humans are helpless to control, alter or steer.

about AI as an urgent national security issue.[127] This attention also contributes to a divergence between the EU's focus on AI governance and human rights protections and the US and UK's focus on economic growth and national security.[128] In addition to the "AI race" narrative are framings of AI as engaged in a zero-sum competition or battle for dominance with humans.[129] Yet these framings both obscure the powerful human agencies that today constitute what we call "AI," and feed a dangerous illusion of *technological determinism:* ➤ SEE SQ10.B the false claim that new technologies such as AI shape society independently of human choices and values, in a manner that humans are helpless to control, alter or steer. In addition to disguising human responsibility for the future shape of AI, these race or competition narratives also obscure meaningful possibilities for the future of AI to be developed in collaborative and participatory ways, or designed to support and enhance human agency rather than undermine it.

Industry and tech evangelists and policymakers[130] are another source of public information about AI. Most of the messaging surrounds promulgating "AI for good" and "responsible/ethical AI" narratives, although these messages raise some concerns about ethics-washing, or insincere corporate use of ethical framings to deflect regulatory and public scrutiny.[131] A minority of voices are still pushing a narrative of technological determinism or inevitability,[132] but more nuanced views of AI as a human responsibility are growing, including an increasing effort to engage with ethical considerations. For example, Google has teams that study ethics, fairness, and bias—although the company's public credibility in this regard took a hit with the controversial departure of Timnit Gebru and Margaret Mitchell, co-leaders of one of the ethical AI teams.[133] There has also been some industry advancement of progressive/reformist AI narratives that rethink the application of technology in social justice or critical theory terms,[134] but it remains limited.

There is a growing critical narrative around unscientific commercial and academic attempts to use AI, particularly tools for facial, gait, or sentiment analysis for behavioral prediction or classification amounting to a "new phrenology."[135] A powerful movement against law-enforcement use of facial-recognition technology peaked in influence during summer 2020, in the wake of the protests against police violence and systemic racism. IBM, Amazon, and Microsoft all announced some sort of pause or moratorium on the use of the technology.[136] Broad international movements in Europe, the US, China, and the UK have been pushing back against the indiscriminate use of facial-recognition systems on the general public.[137]

# Improving and Widening Public Understanding of AI: Where Do We Go From Here?

The AI community could take a lesson from the climate-science community in how to improve its public outreach. Like so many scientists, climate researchers were initially reluctant to engage with outside audiences such as policymakers and the media. But over time it became clear that such engagement was essential to moving forward on some of the most pressing issues of our time—and, over the past decade or so, those scientists have made huge strides in public engagement.

A similar transformation in AI would be beneficial as society grapples with the impacts of these technologies. Some existing programs are working to address these concerns; for instance, the American Association for the Advancement of Science focused its 2020–2021 Leshner Leadership Institute Public Engagement Fellowships on AI.[138] But the challenge remains to identify which forms of public engagement are working—and also who we aren't yet reaching.

To help focus public relations, the AI community should facilitate a clearer public understanding that reduces confusion between AI and other information technologies, without artificially separating AI from other tech and platform structures that heavily influence its development and deployment. We should help the public acquire a useful taxonomy of AI that will support them in making relevant distinctions between the very different types and uses of AI tools. We should also be very clear and consistent that, while we believe advances in AI technology are being made and can have profound benefits for society, we do not support misleading hype that makes it sound as if the latest breakthrough is the one that changes everything.

We should responsibly educate the public about AI, making clear that different publics and subgroups face very different risks from AI, have different social expectations and priorities, and stand to gain or lose much more from AI than other groups. Our public education efforts need to navigate the challenges of providing accurate, balanced information to the public without pretending that there is some single objective, disinterested, and neutral view of AI to present.

Most importantly, we need to move beyond the goal of educating or talking *to* the public and toward more participatory engagement and conversation *with* the public. Work has already begun in many organizations on developing more deliberative and participatory models of AI public engagement.[139] Such efforts will be vital to boosting public interest in and capability for democratic involvement with AI issues that concern us all.

## SQ7. WHAT ARE GOVERNMENTS DOING TO ENSURE THAT AI IS DEVELOPED AND USED RESPONSIBLY?

As AI has grown in sophistication and its use has become more widespread, governments and public agencies have paid increasing attention to its development and deployment. This investment has been especially true in the last five years, when AI has become more and more

> Since the publication of the last AI100 report just five years ago, over 60 countries have engaged in national AI initiatives.

commonly used in consumer products and as private and government applications such as facial recognition[140] ➤ SEE SQ6.C have captured increasing public attention.

Since the publication of the last AI100 report just five years ago, over 60 countries have engaged in national AI initiatives,[141] and several significant new multilateral efforts are aimed at spurring effective international cooperation on related topics.[142] Increases in international government attention to AI issues reflect an understanding that the topic is complex and intersects with other policy priorities, including privacy, equity, human rights, safety, economics, and national and international security.

## Law, Policy, and Regulation ➤ SQ7.A

In the past few years, several legislative and international groups have awoken to the challenge of regulating AI effectively.[143] Few countries have moved definitively to regulate AI specifically, outside of rules directly related to the use of data. Several international groups have developed efforts or initiatives aimed at generating policy frameworks for responsible AI development and use, resulting in recommendations such as the AI Principles of the 38-member-country Organisation for Economic

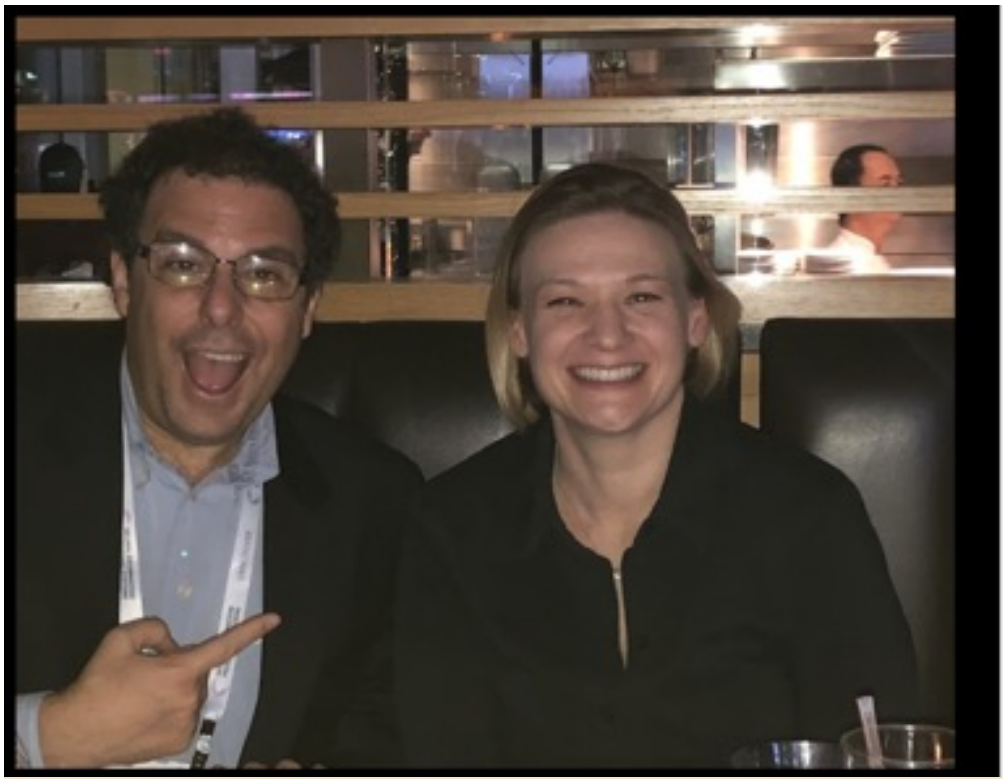

*Facial recognition technology, demonstrated here via Google Photos on a 2019 photo taken at an AI conference, can spot a wide range of individuals in photos and associate them with their names. Applying the same ideas to massive collections of imagery posted online makes it possible to spot and name strangers in public. The capability raises concerns about how AI can simplify mass intrusions into the privacy rights of citizens by governments and private companies all over the world. From: Michael Littman and https://photos.google.com/.*

Co-operation and Development[144]. The EU has been the most active government body to date in proposing concrete regulatory frameworks for AI. In April 2021, it published a new Coordinated Plan on AI, a step toward building a legal framework that "will help to make Europe a safe and innovation friendly environment for the development of AI."[145]

As of 2020, 24 countries—in Asia, Europe, and North America—had opted for permissive laws to allow autonomous vehicles to operate in limited settings ➤ SEE SQ2.E. Thirteen countries—in Africa, Europe, and Latin America—had discussed legislation on the use of autonomous lethal weapons, discussed in more detail below; only Belgium had enacted law.[146]

A range of governance approaches have started to emerge to ensure public safety, consumer trust, product reliability, accountability, and oversight. These efforts involve governments and public agencies, corporations, and civil society, as well as cooperation between the public and private sectors. For example, the US is working actively to develop frameworks for AI risk assessment and regulatory guidance for federal agencies, and is investigating both regulatory and nonregulatory approaches to oversight for AI technologies.[147] Such approaches might include sector-specific policy guidance, pilot studies, voluntary frameworks for compliance, formal standards, or other policy vehicles and related guidelines. This process necessarily involves the government identifying the statutory authorities of each agency.

The EU has been particularly active with concrete regulation, including the General Data Protection Regulation (GDPR), which includes some regulation of automated decision systems, and the Framework of Ethical Aspects of AI Robotics and Related Technologies, which proposes the creation of national supervisory bodies and the designation of high-risk technologies.[148] Canada's Bill C-11 proposes regulation of automated decision systems and has more robust support for people's right to explanations of automated decisions than the EU's approach.[149] Governmental consideration of antitrust action against big tech companies in the EU and US is driven in large part by the scale that has been achieved with the help of AI techniques. Interest in more concrete antitrust activity in the US, in particular, has increased with the Biden Administration, for example with the President's July 2021 Executive Order on competition, which prominently features the American information technology sector.[150]

Because AI is not just one technology but a constellation of capabilities being applied to diverse domains, the US regulates it as distinct products and applications—a reflection of the government's structure and resulting regulatory practices. For example, autonomous vehicle safety guidance and related policies fall under the purview of the Department of Transportation, while oversight and policies for healthcare applications fall to agencies such as the Food and Drug Administration. A cross-industry approach toward AI regulation related to more specific issues, such as data use, has the potential to provide more consistency, although it is still too early to formulate an informed policy along these lines.

For some technology areas, however, it is less clear-cut where government responsibility for regulation is situated. For example, the oversight of social media platforms has become a hotly debated issue worldwide.

As user bases for these companies have grown, so too have the companies' reach and power in issues ranging from medical misinformation to undermining elections. In 2020, approximately 48.3 percent of the global population used social media—a share projected to increase to over 56 percent by 2025.[151] International oversight of some kind is essential to minimize the risks to consumers worldwide.

Misinformation and disinformation are affected by a given platform's user bases, optimization algorithms, content-moderation policies and practices, and much more. In the US, free speech challenges are governed by constitutional law and related legal interpretations. Some companies have even gone so far as to appoint independent oversight boards, such as the Facebook Oversight Board created in 2020, to make determinations about enacting—and applying—corporate policy related to free speech issues to avoid stronger government oversight. Most content moderation decisions are made by individual companies on the basis of their own legal guidance, technical capacity, and policy interpretations, but debate rages on about whether active regulation would be appropriate. There are few easy answers about what kinds of policies should be enacted, and how, and who should regulate them.

## AI Research & Development as a Policy Priority

Globally, investment in AI research and development (R&D) in the past five years by both corporations and governments has grown significantly.[152] In 2020, the US government's investment in unclassified R&D in AI-related technologies was approximately $1.5 billion[153]—a number dwarfed significantly by estimates of the investments being made by top private sector companies in the same year. In 2015, an Obama Administration report made several

---

projections, since borne out, about the near future of AI R&D: AI technologies will grow in sophistication and ubiquity; the impact of AI on employment, education, public safety, national security, and economic growth will continue to increase; industry investment in AI will grow; some important areas concerning the public good will receive insufficient investment by industry; and the broad demand for AI expertise will grow, leading to job-market pressures.[154] The final report of the US National Security Commission on Artificial Intelligence,[155] published in 2021, echoes similar themes.

The Chinese Government's investment in AI research and development in 2018 was estimated to be the equivalent of $9.4 billion, supplemented by significant government support for private investment and strategy development.[156] In Europe, significant public investment increases have been made over the last five years, accompanied by a sweeping EU-led strategy with four prongs: enabling the development and uptake of AI in the EU; making the EU the place where AI thrives from the lab to the market; ensuring that AI works for people and is a force for good in society; and building strategic leadership in high-impact sectors.[157]

Overall, global governments need to invest more significantly in research, development, and regulation of issues surrounding AI, and in multidisciplinary and cross-disciplinary research in support of these objectives. Government investment should also include supporting K-12 education standards to help the next generation to live in a world infused with AI applications, and shaping market practices concerning the use of AI in public-facing applications such as healthcare delivery.

# Cooperation and Coordination on International Policy

Cooperative efforts among countries have also emerged in the last several years. In March 2018, the European Commission established a high-level expert group to support strategy and policy development for AI.[158] The same year, the Nordic-Baltic region released a joint strategy document,[159] and the UAE and India signed a partnership to spur AI innovation.[160] In 2020, the OECD launched an AI Policy Observatory, a resource repository for AI development and policy efforts.[161] In June 2020, the G7, led by the Canadian and French governments, established the Global Partnership on AI, a multilateral effort to promote more effective international collaboration on issues of AI governance.[162]

Though almost all countries expending resources on AI see it as a set of enabling technologies of strategic importance, important differences in country-by-country approaches have also emerged. Notably, China's record on human rights intersects meaningfully with its efforts to become a dominant AI power.[163] Authoritarian powers can put AI to powerful use in building upon and reinforcing existing citizen surveillance programs, which has widespread implications for global AI governance and use—in China, but also everywhere else in the world.[164] In addition, some international coordination could help ease tensions building up as nations strive to position themselves for dominance in AI.[165] Recently, a significant multilateral initiative between the US and the EU has emerged to support more effective coordination and collaboration, with an explicit focus on issues like technology standards and the misuse of technology threatening security and human rights.[166]

## Case Study: Lethal Autonomous Weapons

A case of a specific application of AI that has drawn international attention is lethal autonomous weapon systems (LAWS). LAWS are weapons that, after activation, can select and engage targets without further human intervention.[167] Dozens of countries around the world have operated limited versions of these systems for decades. Close-in weapon systems that protect naval ships and military bases from attacks often have automatic modes that, once activated, select and engage attacking targets without human intervention (though with human oversight).[168] But many AI and robotics researchers have expressed concerns about the way advances in AI, paired with autonomous systems, could generate new and dangerous weapon systems that threaten international stability.[169] Many express specific concerns around the use of autonomous drones for targeted killing—such as accountability, proliferation, and legality.

One challenge for governments in navigating this debate is determining what exactly constitutes a LAWS, especially as smarter munitions increasingly incorporate AI to make them harder for adversary defenses to detect and destroy. The United Nations Convention on Certain Conventional Weapons (CCW) has debated LAWS since 2013. It has convened a Group of Government Experts (GGE) that has met regularly to discuss LAWS.[170] While many smaller countries have endorsed a ban on LAWS, major militaries such as the United States and Russia, as well as NATO countries, have generally argued that LAWS are already effectively regulated under international humanitarian law, and that there are dangers in the unintended consequences of over-regulating technologies that have not yet been deployed.[171]

Regardless of how the LAWS debate in particular is resolved, the greater integration of AI by militaries around the world appears inevitable in areas such as training, logistics, and surveillance. Indeed, there are areas like mine clearing where this is to be welcomed. Governments will need to work hard to ensure they have the technical expertise and capacity to effectively implement safety and reliability standards surrounding these military uses of AI.

## From Principles to Practice

Beyond national policy strategies, dozens of governments, private companies, intergovernmental organizations, and research institutions have also published documents and guidelines designed to address concerns about safety, ethical design, and deployment of AI products and services. These documents often take the form of declarations of principles or high-level frameworks. Such efforts started becoming popular in 2017; in 2018, 45 of them were published globally. A total of at least 117 documents relating to AI principles were published between 2015 and 2020, the majority of which were published by companies.[172]

While these efforts are laudable, statements of responsible AI principles or frameworks in companies are of limited utility if they are incompatible with instruments of oversight, enforceability, or accountability applied by governments. Human rights scholars and advocates, for example, have long pushed for a rights-

---

based approach to AI-informed decision-making, rooted in international law and applicable to a wide array of technologies, and adaptable even as AI itself continues to develop.[173] Related debates have played out in government policy development, as in a 2019 discussion paper published by the Australian Human Rights Commission.[174] Such arguments also point to the need for due process and recourse in AI decision-making.

Several efforts have been made to move organizations working on setting principles toward the implementation, testing, and codification of more effective practice in responsible AI development and deployment. Many see this direction as a precursor, or essential ingredient, to effective policy-making. Efforts developed in the last five years include the Partnership on AI, a nonprofit multi-stakeholder organization created in 2016 by technology companies, foundations, and civil society organizations focused on best-practice development for AI.[175] Much of its work centers on best practice development for more responsible, safe, and user-centered AI, with the goal of ensuring more equitable, effective outcomes.

## Dynamic Regulation, Experimentation, and Testing

Appropriately addressing the risks of AI applications will inevitably involve adapting regulatory and policy systems to be more responsive to the rapidly advancing pace of technology development. Current regulatory systems are already struggling to keep up with the demands of technological evolution, and AI will continue to strain existing processes and structures.[176] There are two key problems: Policy-making often takes time, and once rules are codified they are inflexible and difficult to adapt. Or, put a different way, AI moves quickly and governments move slowly.

To deal with this mismatch of timescales, several innovative solutions have been proposed or deployed that merit further consideration. Some US agencies, such as the Food and Drug Administration, already invest heavily in *regulatory science*—the study of the act of effective regulation itself. This kind of investigation involves research and testing to address gaps in scientific understanding or to develop tools and methods needed to inform regulatory decisions and policy development.[177] (Should AI, for instance, be classified as a device, an aid, or a replacement for workers? The answer impacts how government oversight is applied.) Such approaches should be evangelized more widely, adopted by other agencies, and applied to new technology areas, including AI. Other proposals, drawing inspiration from industry approaches to developing goods and services, advocate for the creation of systems in which governments would hire private companies to act as regulators.[178]

Frameworks for "risk-based" rulemaking and impact assessments are also relevant to new AI-based technologies and capabilities. Risk-based regulatory approaches generally focus on activities that pose the highest risk to the public well-being, and in turn reduce burdens for a variety of lower-risk sectors and firms. In the AI realm specifically, researchers, professional organizations, and governments have begun development of AI or algorithm impact assessments (akin to the use of environmental impact assessments before beginning new engineering projects).[179]

⟩ SQ7.C

Experimentation and testing efforts are an important aspect of both regulatory and nonregulatory approaches to rulemaking for AI, and can take place in both real-world and simulated environments. Examples include the US Federal Aviation Administration's Unmanned Aircraft System (UAS) Test Sites program, which has now been running successfully for several years and feeds data, incident reports, and other crucial information directly back to the agency in support of rulemaking processes for safe UAS airspace integration.[180] In the virtual world, simulations have been built for testing AI-driven tax policy proposals, among other ideas, before deployment.[181] In some cases, third-party certification or testing efforts have emerged.[182] As with any emerging technology—and especially one so diverse in its applications as AI—effective experimentation and testing can meaningfully support more effective governance and policy design.

> As with any emerging technology—and especially one so diverse in its applications as AI—effective experimentation and testing can meaningfully support more effective governance and policy design.

# SQ8. WHAT SHOULD THE ROLES OF ACADEMIA AND INDUSTRY BE, RESPECTIVELY, IN THE DEVELOPMENT AND DEPLOYMENT OF AI TECHNOLOGIES AND THE STUDY OF THE IMPACTS OF AI?

In most research areas, and historically in AI, there has been a relatively clear differentiation between the roles of academia and industry. Academics focus more on basic research, education, and training, while industry focuses more on applied research and development in commercially viable application domains. In the field of AI in recent years, however, this distinction has eroded.

Although academia and industry have each played central roles in shaping AI technologies and their uses, their efforts have been loosely coordinated at best. Now, as AI takes on added importance across most of society, there is potential for conflict between the private and public sectors regarding the development, deployment, and oversight of AI technologies.

The last five years have seen considerable debate regarding the appropriate roles and relationship of academia and industry in the development and deployment of AI applications.[183] This debate arises from two facts. First, the commercial sector continues to lead in AI investment, research and applications, outpacing academia and government spending combined. In the

> SQ8.A

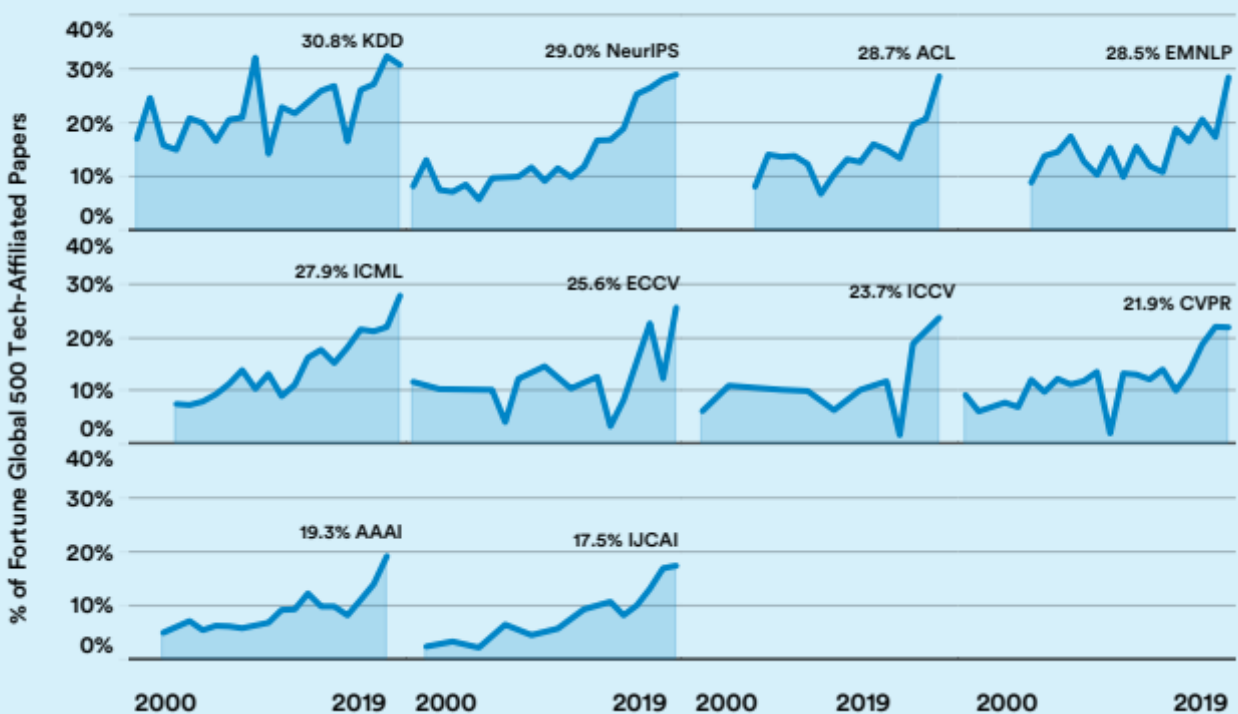

**SHARE of FORTUNE GLOBAL 500 TECH-AFFILIATED PAPERS**
Source: Ahmed & Wahed, 2020 | Chart: 2021 AI Index Report

*Corporate participation in academic conferences has been expanding. At flagship conferences like NeurIPS, nearly a third of all papers include a Fortune Global 500 affiliation. From: https://aiindex.stanford.edu/wp-content/uploads/2021/03/2021-AI-Index-Report_Master.pdf.*

US, private enterprises have spent over $80 billion on AI, while non-defense investment by the federal government in research and development is estimated at only $1.5 billion in 2020. Second, many researchers are opting out of academia for full-time roles in industry, and the long-term consequences of this shift are potentially worrying.[184] To understand the extent to which these concerns might affect how AI develops and shapes society, we must consider the range of ideal roles academia and industry might play.[185]

## Research and Innovation

It is now easier than ever to translate basic AI research into commercially viable products, thanks to the availability of relatively inexpensive, large-scale cloud computing, powerful open-source libraries, and pre-trained models for language, vision, and more. Access to such technology has created new incentives for university researchers, including faculty, postdocs, and graduate students, to create startups or seek other mechanisms to commercialize their intellectual property.

Meanwhile, the presence and influence of industry-led research at AI conferences has increased dramatically. For example, at the 2020 Neural Information Processing Systems Conference (NeurIPS), one of the premier, most widely attended and highly visible conferences in the area of machine learning, 21 percent of the papers were

---

contributed by industrial researchers.[186] This figure compares to 9.5 percent in 2005 (across all papers), the time at which this conference began to see a significant increase in submissions.[187] This shift raises concerns that published research is becoming more applied (and perhaps less free to confront issues at odds with corporate interests), at the risk of stifling long-term innovation and value. On the other hand, industry's increased presence might be helping catalyze the search for innovative solutions to real-world problems.

This increased mixing of academic and industrial research has raised concerns about the impact of "keeping up with the Joneses." A study of the amount of computing resources needed to train large natural-language processing models, such as the models known as transformers,[188] ➤ SEE SQ5.A noted that researchers trained nearly 4,800 models using the equivalent of 27 years of GPU compute time at an estimated cost of $103K-$350K (at cloud-compute market prices at the time). Such AI investments are impossible for most academic researchers.

Creating ways to share such models and evaluation environments would provide steps toward alleviating this imbalance. An interesting example comes from the decision by OpenAI to incrementally release the model parameters of their transformer-based GPT-2 network in 2019 and to provide access to its successor, GPT-3, ➤ SEE SQ2.A in 2020.[189] In 2021, the US National Security

> How to ideally allocate resources is an open problem that requires ongoing attention.

Commission on Artificial Intelligence recommended the federal creation of data repositories and access to large-scale computational resources.[190] How to ideally allocate resources is an open problem that requires ongoing attention.

## Research into Societal and Ethical Issues

As the line between academic and industry research in AI blurs, additional social and ethical issues come to the fore. Academic and industry researchers might have different perspectives on—and hence take different approaches to—many sociotechnical challenges that can be at least partially addressed by technical solutions, such as bias in machine-learned models, fairness in AI decision-making algorithms, privacy in data collection, and the emergence of polarization or filter-bubbles in social-media consumption.[191]

In addition, tighter coupling of academic and industrial research may reduce the focus on both longer-term problems and on those that might run counter to commercial interests. There is also a separate ethical consideration around IP ownership when students work directly with faculty whose intellectual products are partly owned by a company.

Companies that want to keep their customers satisfied have strong incentives to act when it comes to issues of privacy and fairness in the use of AI. One example of an action taken is corporate investment in The Partnership on AI, a nonprofit coalition of industry and university stakeholders committed to the responsible use of artificial intelligence.[192] But an incentive to act is not necessarily aligned with the desire to get it right.

## Development and Deployment

Application of advanced research and technology in real-world settings has traditionally occured outside of academia largely because of the high costs associated with development and deployment at scale. These include the costs of infrastructure, engineering, and testing; verification for robustness; and safety, logistics, and delivery—all of which are often most easily  absorbed by companies with a commercial interest in deployment and the specific skills needed to manage these activities. While this dynamic remains largely intact in AI, the last few years have seen academic researchers increasingly able to take their technological innovations out of the lab and deploy them in the wild. A notable example is Duolingo, a language learning system built by academics at Carnegie Mellon,

which went public in 2021 with a $5 billion valuation.[193]

Of course, not all real-world deployment is profit-oriented, and there's no reason that nonprofit applications that benefit the public can't be quickly created and put to use. For example, Oxford and Google collaborated on tracking COVID-19 variants,[194] and, in the US, several universities are cooperating with companies c3.ai and Microsoft to promote urgent applications of AI for future pandemics.[195] These developments have also played a major role in fostering non-commercial collaborations between industry and academia.

## Education and Training

Many people from across the academic research spectrum have decried a perceived brain drain as a large number of AI researchers have left university and research institute posts to join the industrial ranks. Research suggests that this trend has intensified in recent years.[196] According to one study, 131 AI faculty departed for industry (including startups) between 2004 and 2018, while another 90 took reduced academic roles for such pursuits.[197] The study concludes that these departures had a negative consequence on Ph.D. training within the discipline. While there has not yet been a sustained dip in computer science Ph.D. graduates—nor, presumably, AI students—due to faculty departures, there is fear that one may develop.[198]

As student interest in computer science and AI continues to grow, more universities are developing standalone AI/machine-learning programs, departments, and related degree programs.[199] Such programs include both traditional delivery and those that are partly, if not entirely, online.[200] The trends outlined above raise questions as to who will staff these programs, and how they will feed into the pipelines needed to produce AI talent, from software and application developers to Ph.D. students and the next generation of academic leaders.[201]

A partial answer is to encourage industry to play a broader role in training. Internships, for example, where students spend a few months working in a company, offer current and recent students the ability to obtain valuable hands-on experience while addressing applied research questions or strengthening their skills in AI development and deployment. Such opportunities amplify university-based education and can often jumpstart students' careers. Moreover, company-led courses are becoming increasingly common and can fill curricular gaps, especially if more students want access to basic AI education than universities can handle, or if students seek specialized skills that are best learned in the context of real-world applications.[202]

## Societal Impact: Monitoring and Oversight

A controversy involving AI ethics research and researchers at Google in early 2021[203] **▶ SEE SQ6.B** spurred community-wide concerns about reliance on companies to monitor and govern their own ethics practices. For instance, a company can easily withdraw support from any ethics group or initiative whose findings conflict with its near-term business interests.

When it comes to the societal impacts of AI, stakes are high in the academia-industry relationship. Beyond questions of privacy and fairness lie concerns about the potential for AI and machine-learning algorithms to create filter bubbles or shape social tendencies toward

> Serious research is needed to guide effective policy, and that's where academic/industry collaboration can have the greatest impact.

radicalization, polarization, and homogenization by influencing content consumption and user interactions. However, studying and assessing these issues is easiest when academic-industry collaborations facilitate access to data and platforms.

Reducing some of the negative consequences of this more enmeshed relationship may require government regulation and oversight, particularly to guide how societal impacts are monitored, promoted, and mitigated. Any changes in regulation, however, should be made in consultation with the researchers who have the clearest idea of what the key issues are and how they should be addressed. Serious research is needed to guide effective policy, and that's where academic/industry collaboration can have the greatest impact.

---

# SQ9. WHAT ARE THE MOST PROMISING OPPORTUNITIES FOR AI?

This section describes active areas of AI research and innovation poised to make beneficial impact in the near term. Elsewhere, ➤ SEE SQ10 we address potential pitfalls to avoid in the same time frame.

We focus on two kinds of opportunities. The first involves AI that augments human capabilities. Such systems can be very valuable in situations where humans and AI have complementary strengths. For example, an AI system may be able to synthesize large amounts of clinical data to identify a set of treatments for a particular patient along with likely side effects; a human clinician may be able to work with the patient to identify which option best fits their lifestyle and goals, and to explore creative ways of mitigating side effects that were not part of the AI's design space. The second category involves situations in which AI software can function autonomously. For example, an AI system may automatically convert entries from handwritten forms into structured fields and text in a database.

## AI for Augmentation

Whether it's finding patterns in chemical interactions that lead to a new drug discovery or helping public defenders identify the most appropriate strategies to pursue, there are many ways in which AI can augment the capabilities of people. Indeed, given that AI systems and humans have complementary strengths, one might hope that, combined, they can accomplish more than either alone. An AI system might be better at synthesizing available data and making decisions in well-characterized parts of a problem, while a human may be better at understanding the implications of the data (say if missing data fields are actually a signal for important, unmeasured information for some subgroup represented in the data), working with difficult-to-fully-quantify objectives, and identifying creative actions beyond what the AI may be programmed to consider.

Unfortunately, several recent studies have shown that human-AI teams often do not currently outperform AI-only teams.[204] Still, there is a growing body of work on methods to create more effective human-AI collaboration ➤ SEE SQ4.A in both the AI and human-computer-interaction communities. As this work matures, we see several near-term opportunities for AI to improve human capabilities and vica versa. We describe three major categories of such opportunities below.

### DRAWING INSIGHTS

There are many applications in which AI-assisted insights are beginning to break new ground and have large potential for the future. In chemical informatics and drug discovery,[205] AI assistance is helping identify molecules worth synthesizing in a wet lab. In the energy sector, patterns identified by AI algorithms are helping achieve greater efficiencies[206]. By first training a model to be very good at making predictions, and then working to understand why those predictions are so good, we have deepened our scientific understanding of everything from disease[207] to earthquake dynamics.[208] AI-based tools will

continue to help companies and governments identify bottlenecks in their operations.[209]

AI can assist with discovery. While human experts can always analyze an AI from the outside—for example, dissecting the innovative moves made by AlphaGo[210]—new developments in interpretable AI and visualization of AI are making it much easier for humans to inspect AI programs more deeply and use them to explicitly organize information in a way that facilitates a human expert putting the pieces together and drawing insights. For example, analysis of how an AI system internally organizes words (known as an *embedding* or a *semantic representation*) is helping us understand and visualize the way words like "awful" (formally "inspiring awe") undergo semantic shifts over time.[211]

## ASSISTING WITH DECISION-MAKING

The second major area of opportunity for augmentation is for AI-based methods to assist with decision-making. For example, while a clinician may be well-equipped to talk through the side effects of different drug choices, they may be less well-equipped to identify a potentially dangerous interaction based on information deeply embedded in the patient's past history. A human driver may be well-equipped for making major route decisions and watching for certain hazards, while an AI driver might be better at keeping the vehicle in lane and watching for sudden changes in traffic flow. Ongoing research seeks to determine how to divide up tasks between the human user and the AI system, as well as how to manage the interaction between the human and the AI software. In particular, it is becoming increasingly clear that all stakeholders need to be involved in the design of such AI assistants to produce a human-AI team that outperforms either alone. Human users must understand the AI system and its limitations to trust and use it appropriately, and AI system designers must understand the context in which the system will be used (for example, a busy clinician may not have time to check whether a recommendation is safe or fair at the bedside).

There are several ways in which AI approaches can assist with decision-making. One is by summarizing data too complex for a person to easily absorb. In oncology and other medical fields, recent research in AI-assisted summarization promises to one day help clinicians see the most important information and patterns about a patient.[212] Summarization is also now being used or actively considered in fields where large amounts of text must be read and analyzed—whether it is following news media, doing financial research, conducting search engine optimization, or analyzing contracts, patents, or legal documents. Summarization and interactive chat technologies have great potential to help ensure that people get a healthy breadth of information on a topic, and to help break filter bubbles rather than make them—by providing a range of information, or at least an awareness of the biases in one's social-media or news feeds. Nascent progress in highly realistic (but currently not reliable or accurate) text generation, such as GPT-3, may also make these interactions more natural.

In addition to summarization, another aid for managing complex information is assisting with making predictions about future outcomes (sometimes also called forecasting or risk scoring). An AI system may be able to reason about the long-term effects of a decision, and so be able to recommend that a doctor ask for a particular set of tests, give a particular treatment, and so on, to improve long-term outcomes. AI-based early warning systems are becoming much more commonly used in health settings,[213] agriculture,[214] and more broadly.

> SQ9.A  _Neural-network_ language models called "transformers" consisting of billions of parameters trained on billions of words of text, can be used for grammar correction, creative writing, and generating realistic text. In this example, the transformer-based GPT-3 produces a natural sounding product description for a non-existent, and likely physically impossible, toy. From: _https://www.gwern.net/docs/www/justpaste.it/b5a07c7305ca81b0de2d324f09445f9ef407c17e.html_

Conveying the likelihood of an unwanted outcome—be it a patient going into shock or an impending equipment failure—can help prevent a larger catastrophe. AI systems may also help predict the effects of different climate-change-mitigation or pandemic-management strategies and search among possible options to highlight those that are most promising.[215] These forecasting systems typically have limits and biases based on the data they were trained on, and there is also potential for misuse if people overtrust their predictions or if the decisions impact people directly ➤ **SEE WQ1**.

AI systems increasingly have the capacity to help people work more efficiently. In the public sector, relatively small staffs must often process large numbers of public comments, complaints, potential cases for a public defender, requests for corruption investigations, and more, and AI methods can assist in triaging the incoming information. On education platforms, AI systems can provide initial hints to students and flag struggling students to educators. In medicine, smartphone-based pathology processing can allow for common diagnoses to be made without trained pathologists, which is especially crucial in low-resource settings.[216] Language processing tools can help identify mental health concerns at both a population and individual scale and enable, for example, forum moderators to identify individuals in need of rapid intervention.[217] AI systems can help assist both clinicians and patients in deciding when a clinic visit is needed and provide personalized prevention and wellness assistance in the meantime.[218] More broadly, chatbots and other AI programs can help streamline business operations, from financial to legal. As always, while these efficiencies have the potential to expand the positive impact of low-resourced, beneficial organizations, such systems can also result in harm ➤ **SEE SQ10** when designed or integrated in ways that do not fully and ethically consider their sociotechnical context.[219]

Finally, AI systems can help human decision-making by leveling the playing field of information and resources. Especially as AI becomes more applicable in lower-data regimes, predictions can increase economic efficiency of everyday users by helping people and businesses find relevant opportunities, goods, and services, matching producers and consumers. These uses go beyond major platforms and electronic marketplaces; kidney exchanges,[220] for example, save many lives, combinatorial markets allow goods to be allocated fairly, and AI-based algorithms help select representative populations for citizen-based policy-making meetings.[221]

 ## AI AS ASSISTANT

A final major area of opportunity for augmentation is for AI to provide basic assistance during a task. For example, we are already starting to see AI programs that can process and translate text from a photograph,[222] allowing travelers to read signage and menus. Improved translation tools will facilitate human interactions across cultures. Projects that once required a person to have highly specialized knowledge or copious amounts of time—from fixing your sink to creating a diabetes-friendly meal—may become accessible to more people by allowing them to search for task- and context-specific expertise (such as adapting a tutorial video to apply to unique sink configuration).

Basic AI assistance has the potential to allow individuals to make more and better decisions for themselves. In the area of health, the combination of sensor data and AI analysis is poised to help promote a range of behavior changes, including exercise, weight loss, stress management, and dental hygiene.[223] Automated systems are already in use for blood-glucose control[224] and providing ways to monitor and coordinate care at home. AI-based tools can allow people with various disabilities—such as limitations in vision, hearing, fine and gross mobility, and memory—to live more independently and participate in more activities. Many of these programs can run on smartphones, further improving accessibility.[225]

Simple AI assistance can also help with safety and security. We are starting to see lane-keeping assistance and other reaction-support features in cars.[226] It is interesting that self-driving cars have been slow in development and adoption, but the level of automation and assistance in "normal" cars is increasing—perhaps because drivers value their (shared) autonomy with the car, and also because AI-based assistance takes certain loads off the driver while letting them do more nuanced tasks (such as waving or making eye contact with a pedestrian to signal they can cross). AI-assisted surgery tools are helping make movements in surgical operations more precise.[227] AI-assisted systems flag potential email-based phishing attacks to be checked by the user, and others monitor transactions to identify everything from fraud to cyberattacks.

# AI Agents on Their Own

Finally, there is a range of opportunities for AI agents acting largely autonomously or not in close connection with humans. Alphafold[228] recently made significant progress toward solving the protein-folding problem, and we can expect to see significantly more AI-based automation in chemistry and biology. AI systems now help convert handwritten forms into structured fields, are starting to automate medical billing, and have been used recently to scale efforts to monitor habitat biodiversity.[229] They may also help monitor and adjust operations in fields like clean energy, logistics, and communications; track and communicate health information to the public; and create smart cities that make more efficient use of public services, better manage traffic, and reduce climate impacts. The pandemic saw a rise in fully AI-based education tools that attempt to teach without a human educator in the loop, and there is a great deal of potential for AI to assist with virtual reality scenarios for training, such as practicing how to perform a surgery or carry out disaster relief. We expect many mundane and potentially dangerous tasks to be taken over by AI systems in the near future.

In most cases, the main factors holding back these applications are not in the algorithms themselves, but in the collection and organization of appropriate data and the effective integration of these algorithms into their broader sociotechnical systems **> SEE WQ1.A.** For example, without significant human-engineered knowledge, existing machine-learning algorithms struggle to generalize to "out of sample" examples that differ significantly from the data on which they were trained. Thus, if Alphafold, trained on natural proteins, fails on synthetic proteins, or if a handwriting-recognition system trained on printed letters fails on cursive letters, these failures are due to the way the algorithms were trained, not the algorithms per se. (Consider the willingness of big tech companies like Facebook, Google, and Microsoft

> The pandemic saw a rise in fully AI-based education tools and there is a great deal of potential for AI to assist with virtual reality scenarios for training, such as practicing how to perform a surgery or carry out disaster relief.

to share their deep learning algorithms and their reluctance to share the data they use in-house.)

Similarly, most AI-based decision-making systems require a formal specification of a reward or cost function, and eliciting and translating such preferences from multiple stakeholders remains a challenging task. For example, an AI controller managing a wind farm has to manage "standard" objectives such as maximizing energy produced and minimizing maintenance costs, but also harder-to-quantify preferences such as reducing ecological impact and noise to neighbors. As with the issue of insufficient relevant data, a failure of the AI in these cases is due to the way it was trained—on incorrect goals—rather than the algorithm itself.

In some cases, further challenges to the integration of AI systems come in the form of legal or economic incentives; for example, malpractice and compliance concerns have limited the penetration of AI in the health sector. Regulatory frameworks for safe, responsible innovation will be needed to achieve these possible near-term beneficial impacts.

---

# SQ10. WHAT ARE THE MOST PRESSING DANGERS OF AI?

 As AI systems prove to be increasingly beneficial in real-world applications, they have broadened their reach, causing risks of misuse, overuse, and explicit abuse to proliferate. As AI systems increase in capability and as they are integrated more fully into societal infrastructure, the implications of losing meaningful control over them become more concerning.[230] New research efforts are aimed at re-conceptualizing the foundations of the field to make AI systems less reliant on explicit, and easily misspecified, objectives.[231] A particularly visible danger is that AI can make it easier to build machines that can spy and even kill at scale ❯ SEE SQ7.B. But there are many other important and subtler dangers at present.

 ## Techno-Solutionism

One of the most pressing dangers of AI is techno-solutionism, the view that AI can be seen as a panacea when it is merely a tool.[232] As we see more AI advances, the temptation to apply AI decision-making to all societal problems increases. But technology often creates larger problems in the process of solving smaller ones. For example, systems that streamline and automate the application of social services can quickly become rigid and deny access to migrants or others who fall between the cracks.[233]

When given the choice between algorithms and humans, some believe algorithms will always be the less-biased choice. Yet, in 2018, Amazon found it necessary to discard a proprietary recruiting tool because the historical data it was trained on resulted in a system that was systematically biased against women.[234] Automated decision-making can often serve to replicate, exacerbate, and even magnify the same bias we wish it would remedy.

Indeed, far from being a cure-all, technology can actually create feedback loops that worsen discrimination. Recommendation algorithms, like Google's page rank, are trained to identify and prioritize the most "relevant" items based on how other users engage with them. As biased users feed the algorithm biased information, it responds with more bias, which informs users' understandings and deepens their bias, and so on.[235] Because all technology is the product of a biased system,[236] techno-solutionism's flaws run deep:[237] a creation is limited by the limitations of its creator.

## Dangers of Adopting a Statistical Perspective on Justice 

Automated decision-making may produce skewed results that replicate and amplify existing biases. A potential danger, then, is when the public accepts AI-derived conclusions as certainties. This determinist approach to AI decision-making can have dire implications in both criminal and healthcare settings. AI-driven approaches like PredPol, software originally developed by the Los Angeles Police Department and UCLA that purports to help protect one in 33 US citizens,[238] predict when, where, and how crime will occur. A 2016 case study of a US city noted that the approach disproportionately projected crimes in areas with higher populations of non-white and low-income residents.[239] When datasets disproportionately represents the lower power members of society, flagrant discrimination is a likely result.

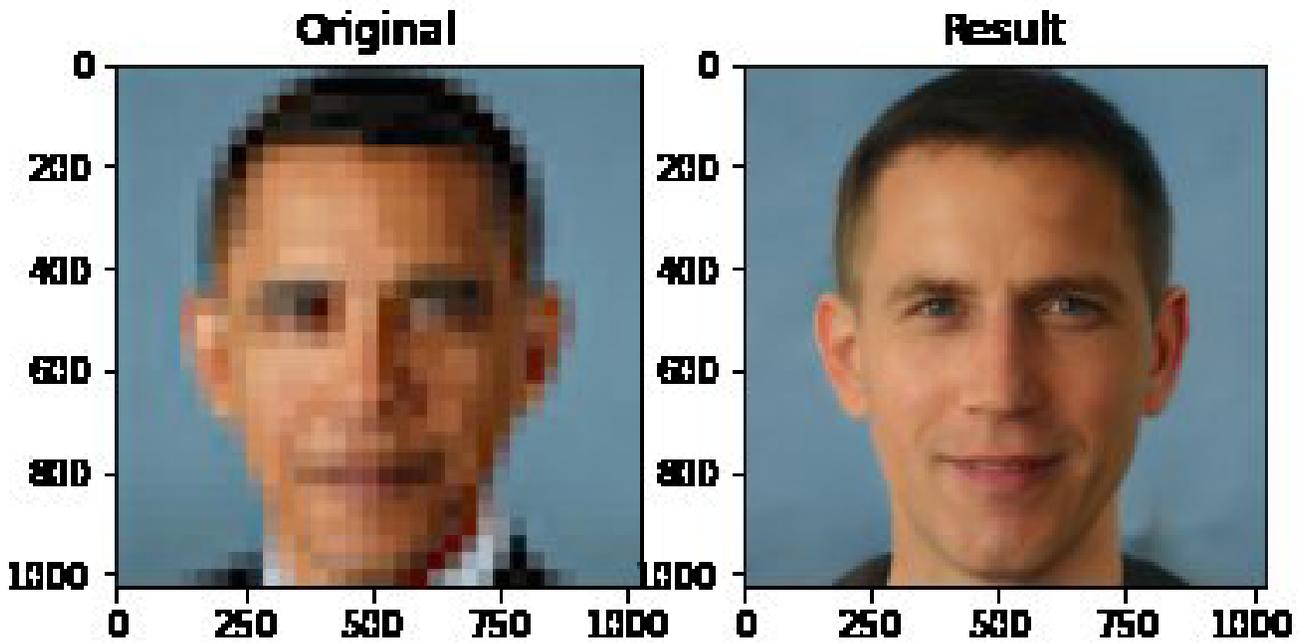

*Image-generation GANs can be used to perform other tasks like translating low-resolution images of faces into high resolution images of faces. Of course, such a transformation is not recovering missing information so much as it is confabulating details that are consistent with its input. As an example, the PULSE system tends to generate images with features that appear ethnically white, as seen in this input image of former US President Barack Obama. From: https://www.theverge.com/21298762/face-depixelizer-ai-machine-learning-tool-pulse-stylegan-obama-bias*

Sentencing decisions are increasingly decided by proprietary algorithms that attempt to assess whether a defendant will commit future crimes, leading to concerns that justice is being outsourced to software.[240] As AI becomes increasingly capable of analyzing more and more factors that may correlate with a defendant's perceived risk, courts and society at large may mistake an algorithmic probability for fact. This dangerous reality means that an algorithmic estimate of an individual's risk to society may be interpreted by others as a near certainty—a misleading outcome even the original tool designers warned against. Even though a statistically driven AI system could be built to report a degree of credence along with every prediction,[241] there's no guarantee that the people using these predictions will make intelligent use of them. Taking probability for certainty means that the past will always dictate the future.

There is an aura of neutrality and impartiality associated with AI decision-making in some corners of the public consciousness, resulting in systems being accepted as objective even though they may be the result of biased historical decisions or even blatant discrimination. All data insights rely on some measure of interpretation. As a concrete example, an audit of a resume-screening tool found that the two main factors it associated most strongly with positive future job performance were whether the applicant was named Jared, and whether he played high school lacrosse.[242] Undesirable biases can be hidden behind both the opaque nature of the technology used and the use of proxies, nominally innocent attributes that enable a decision that is fundamentally biased. An algorithm fueled by data in which gender, racial, class, and ableist biases are pervasive can effectively reinforce these biases without ever explicitly identifying them in the code.

Without transparency concerning either the data or the AI algorithms that interpret it, the public may be left in the dark as to how decisions that materially impact their lives are being made. Lacking adequate information to bring a legal claim, people can lose access to both due process and redress when they feel they have been improperly or erroneously judged by AI systems. Large gaps in case law make applying Title VII—the primary existing legal framework in the US for employment discrimination—to cases of algorithmic discrimination incredibly difficult. These concerns are exacerbated by algorithms that go beyond traditional considerations such as a person's credit score to instead consider any and all variables correlated to the likelihood that they are a safe investment. A statistically significant correlation has been shown among Europeans between loan risk and whether a person uses a Mac or PC and whether they include their name in their email address—which turn out to be proxies for affluence.[243] Companies that use such attributes, even if they do indeed provide improvements in model accuracy, may be breaking the law when these attributes also clearly correlate with a protected class like race. Loss of autonomy can also result from AI-created "information bubbles" that narrowly constrict each individual's online experience to the point that they are unaware that valid alternative perspectives even exist.

## Disinformation and Threat to Democracy

AI systems are being used in the service of disinformation on the internet, giving them the potential to become a threat to democracy and a tool for fascism. From deepfake videos to online bots manipulating public discourse by feigning consensus and spreading fake news,[244] there is the danger of AI systems undermining social trust. The technology can be co-opted by criminals, rogue states, ideological extremists, or simply special interest groups, to manipulate people for economic gain or political advantage. Disinformation poses serious threats to society, as it effectively changes and manipulates evidence to create social feedback loops that undermine any sense of objective truth. The debates about what is real quickly evolve into debates about who gets to decide what is real, resulting in renegotiations of power structures that often serve entrenched interests.[245]

## Discrimination and Risk in the Medical Setting

While personalized medicine is a good potential application of AI, there are dangers. Current business models for AI-based health applications tend to focus on building a single system—for example, a deterioration predictor—that can be sold to many buyers. However, these systems often do not generalize beyond their training data. Even differences in how clinical tests are ordered can throw off predictors, and, over time, a system's accuracy will often degrade as practices change. Clinicians and administrators are not well-equipped to monitor and manage these issues, and insufficient thought given to the human factors of AI integration has led to oscillation between mistrust of the system (ignoring it) and over-reliance on the system (trusting it even when it is wrong), a central concern of the 2016 AI100 report.

These concerns are troubling in general in the high-risk setting that is healthcare, and even more so because marginalized populations—those that already face discrimination from the health system from both structural factors (like lack of access) and scientific factors (like guidelines that were developed from trials on other populations)—may lose even more. Today and in the near future, AI systems built on machine learning are used to determine post-operative personalized pain management plans for some patients and in others to predict the likelihood that an individual will develop breast cancer. AI algorithms are playing a role in decisions concerning distributing organs, vaccines, and other elements of healthcare. Biases in these approaches can have literal life-and-death stakes.

In 2019, the story broke that Optum, a health-services algorithm used to determine which patients may benefit from extra medical care, exhibited fundamental racial biases. The system designers ensured that race was precluded from consideration, but they also asked the algorithm to consider the future cost of a patient to the healthcare system.[246] While intended to capture a sense of medical severity, this feature in fact served as a proxy for race: controlling for medical needs, care for Black patients averages $1,800 less per year.

New technologies are being developed every day to treat serious medical issues. A new algorithm trained to identify melanomas was shown to be more accurate than doctors in a recent study, but the potential for the algorithm to be biased against Black patients is significant as the algorithm was trained using majority light-skinned groups.[247] The stakes are especially high for melanoma diagnoses, where the five-year survival rate is 17 percentage points less for Black Americans than white. While technology has the potential to generate quicker diagnoses and thus close this survival gap, a machine-learning algorithm is only as good as its data set. An improperly trained algorithm could do more harm than good for patients at risk, missing cancers altogether or generating false positives. As new algorithms saturate the market with promises of medical miracles, losing sight of the biases ingrained in their outcomes could contribute to a loss of human biodiversity, as individuals who are left out of initial data sets are denied adequate care. While the exact long-term effects of algorithms in healthcare are unknown, their potential for bias replication means any advancement they produce for the population in aggregate—from diagnosis to resource distribution—may come at the expense of the most vulnerable.

# SQ11. HOW HAS AI AFFECTED SOCIOECONOMIC RELATIONSHIPS?

For millennia, waves of technological change have been perceived as a double-edged sword for the economy and labor market, increasing output and wealth but potentially reducing pay and job opportunities for typical workers. The Roman emperor Vespasian refused to adopt a productivity-enhancing construction technology due to its potential labor market impact;[248] the Luddites destroyed textile machinery in early 1800s England;[249] and, in the 1960s, arguably a golden age for the US labor market, experts warned that labor-saving technology could devastate US employment.[250]

And so it has been with the latest wave of innovation in the field of artificial intelligence. Though characterized by some as a key to increasing material prosperity for human society, AI's potential to replicate human labor at a lower cost has also raised concerns about its impact on the welfare of workers. Are these concerns warranted? The answer is surprisingly murky—complex, yes, but also difficult to characterize precisely. AI has not been responsible for large aggregate economic effects. But that may be because its impact is still relatively localized to narrow parts of the economy.

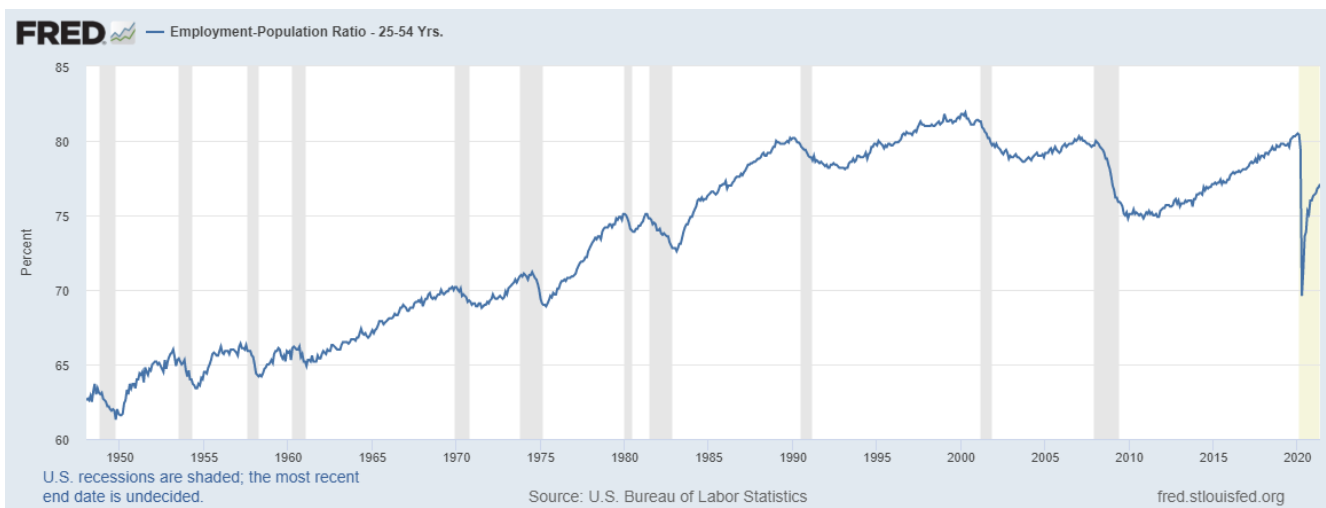

*Data from the US Bureau of Labor Statistics shows that employment as a fraction of the population reached a 20-year high right before the pandemic, suggesting that the growth of AI is not yet producing large-scale unemployment. From: https://fred.stlouisfed.org/series/LNS12300060#0*

## SQ11.A The Story So Far

The recovery from the 2008–2009 recession was sluggish both in North America and Western Europe.[251] Unemployment, which had spiked to multi-decade highs, came down only slowly. This weak recovery was happening at the same time as major innovations in the field of AI and a new wave of startup activity in high tech. To pick one example, it seemed like self-driving cars were just around the corner and would unleash a mass displacement of folks who drive vehicles for a living. So AI (sometimes confusingly referred to as "robots") became a scapegoat for weak labor markets.[252]

To some extent, this narrative of "new technology is taking away jobs" has recent precedents—the two prior labor market recoveries, beginning in 1991 and 2001, also started out weak, and that weakness was subsequently associated with technological innovation.[253] But the possibility of applying the same narrative to the 2008–2009 recession was quickly dispelled by the post-2009 data. Productivity—the amount of economic output that can be produced by a given amount of economic inputs—grew at an exceptionally slow rate during the 2010s, both in the US and in many other countries,[254] suggesting job growth was weak because economic growth was weak, not because technology was eliminating jobs. Employment grew slowly, but so did GDP in western countries.[255] And, after a decade of sluggish recovery, in early 2020 (on the eve of the COVID-19 pandemic), the share of prime-working-age Americans with a job reached its highest level since 2001.[256] In Western Europe, that share hit its highest level since at least 2005. The layperson narrative of the interplay between artificial intelligence technology and the aggregate economy had run ahead, and to some extent become disconnected, from the reality that economists were measuring "on the ground".

In other words, worries by citizens, journalists and policymakers about widespread disruption of the global labor market by AI have been premature. Other forces have been much more disruptive to workers' livelihoods. And, at least in the 2010s, the labor market has been far more capable of healing than many commentators expected.

## AI and Inequality

AI has frequently been blamed for both rising inequality or stagnant wage growth, both in the United States and beyond. Given the history of skill-biased technological change may have played a role in generating inequality,[257] this worry is reasonable to consider. Looking back, the evidence here is mixed, but it's mostly clear that, in the grand scheme of rising inequality, AI has thus far played a very small role.

The first reason, most importantly, is that the bulk of the increase in economic inequality across many countries predates significant commercial use of AI. Arguably, it began in the 1980s.[258] The causes for its increase over that period are hard to disentangle and are much debated—globalization, macroeconomic austerity, deregulation, technological innovation, and even changing social norms could all have played a role. Unless we are willing to call all of these disparate societal trends "AI," there's no way to pin current economic inequality on AI.

The second reason is that, even in the most recent decade, the most significant factors negatively impacting the labor market have not been AI related. Aggregate demand was weak in the US and many western countries during the early years of the decade, keeping wage growth weak (particularly for less educated workers). And, to some degree, impacts that are directly attributable to *technology* are not necessarily attributable to AI specifically; for example, consider the relatively large impact of technologies like camera phones, which wiped out the large photography firm Kodak.[259]

## Localized Impact

In sectors where AI is more prevalent—software and financial services, for example—its labor-market impact is likely to be more meaningful. Yet even in those industries and US states where AI job postings (an imperfect proxy) are more prevalent, they only account for a one to three percent share of total postings. Global corporate investment in AI was $68 billion in 2020, which is a non-trivial sum but small in relative terms: gross private investment over all categories in the US alone was almost $4 trillion in 2020.[260] It's not always easy to differentiate AI's impact from other, older forms of technological automation, but it likely reduces the amount of human labor going into repetitive tasks.[261]

## How the Pie Is Sliced

Economists have historically viewed technology as increasing total economic value (making the pie bigger), while acknowledging that such growth can create winners and losers (some people may end up with smaller slices than when the entire pie was smaller). But it's also conceivable that some new technologies, including AI, might end up simply reslicing a pie of unchanged size. Stated differently, these technologies might be adopted by firms simply to redistribute surplus/gains to their owners.[262] That situation would parallel developments over recent decades like tax cuts and deregulation, which have had a small positive effect on economic growth at best[263] but have asymmetrically benefited the higher end of the income and wealth distributions. In such a case, AI could have a big impact on the labor market and economy without registering any impact on productivity growth. No evidence of such a trend is yet apparent, but it may become so in the future and is worth watching closely.

## Market Power

AI's reliance on big data has led to concerns that monopolistic access to data disproportionately increases market power. If that's correct, then, over time, firms that acquire particularly large amounts of data will capture monopoly profits at the expense of consumers, workers, and other firms.[264] This explanation is often offered for the dominance of big tech by a small number of very large, very profitable firms. (And it might present an even bigger risk if "data monopolies" are allowed by regulators to reduce competition across a wider range of industries.) Yet over the past few decades, consolidation and market power have increased across a range of industries as diverse as airlines and cable providers—so, at the present moment, access to and ownership of data are at most just one factor driving growing concentration of wealth and power.[265] Still, as data and AI propagate across more of the economy, data as a driver of economic concentration could become more significant.

## The Future

To date, the economic significance of AI has been comparatively small—particularly relative to expectations, among both optimists and pessimists, of massive transformation of the economy. Other forces—globalization, the business cycle, and a pandemic—have had a much, much bigger and more intense impact in recent decades.

But the situation may very well change in the future, as the new technology permeates more and more of the economy and expands in flexibility and power.

Economists have offered several explanations for this lag[266]; other technologies that ultimately had a massive impact experienced a J-curve, where initial investment took decades to bear fruit.[267] What should we expect in the context of AI?

First, there is a possibility that the pandemic will accelerate AI adoption; according to the World Economic Forum, business executives are currently expressing an intent to increase automation.[268] Yet parallel worries during the prior economic expansion failed to materialize,[269] and hard evidence of accelerating automation on an aggregate scale is hard to find.[270]

Second, AI will contend with another extremely powerful force: demographics. Populations are aging across the world. In some western countries, workforces are already shrinking. It may be that instead of "killing jobs," AI will help alleviate the crunch of retiring workforces.[271]

Third, technological change takes place over a long time, oftentimes longer than expected.[272] It took decades for electricity[273] and the first wave of information technology[274] to have a noticeable impact on economic data; any future wave of technological innovation is also unlikely to hit all corners of the economy at once. (This insight also helps to contextualize relative disappointment in areas like self-driving vehicles.[275] **▶ SEE SQ2.E.** Change can be slow, even when it's real.) A "hot" labor market in which some sectors of the economy expand labor demand even as others shrink is a useful insurance policy against persistent technology-driven unemployment.

Fourth, AI and other cutting-edge technologies may end up driving inequality. We may eventually see technologically-driven mass unemployment. Even if jobs remain plentiful, the automation-resistant jobs might end up being primarily relatively low-paying service-sector jobs. In the middle of the 20th century, western governments encountered and mitigated such challenges via effective social policy and regulation; since the 1970s, they have been more reluctant to do so. To borrow a phrase from John Maynard Keynes, if AI really does end up increasing "economic possibilities for our grandchildren,"[276] society and government will have it within their means to ensure those possibilities are shared equitably. For example, unconditional transfers such as universal basic income—which can be costly in a world dependent on human labor but could be quite affordable in a world of technology-fueled prosperity and are less of a disorganized patchwork than our current safety net—could play a significant role.[277] But if policymakers underreact, as they have to other economic and labor pressures buffeting workers over the past few decades, innovations may simply result in a pie that is sliced ever more unequally.

> If policymakers underreact, as they have to other economic and labor pressures buffeting workers over the past few decades, innovations may simply result in a pie that is sliced ever more unequally.

# SQ12. DOES IT APPEAR "BUILDING IN HOW WE THINK" WORKS AS AN ENGINEERING STRATEGY IN THE LONG RUN?

Every scientific discipline has foundational questions. In human psychology, for example, there is the nature-versus-nurture question. How much of our behavior is due to our genes, and how much to our environment and upbringing?

AI also has its own fundamental nature-versus-nurture-like question. Should we attack new challenges by applying general-purpose problem-solving methods, or is it better to write specialized algorithms, designed by experts, for each particular problem? Roughly, are specific AI solutions better engineered in advance by people (nature) or learned by the machine from data (nurture)?

➤ SQ12.A

In a March 2019 blog post,[278] Richard Sutton—one of the leading figures in reinforcement learning—articulated the "nurture" perspective. "The biggest lesson that can be read from 70 years of AI research," he wrote, "is that general methods that leverage computation are ultimately the most effective, and by a large margin." He backed up this claim with some compelling examples drawn from subfields of AI such as computer games (where no chess- or Go-specific strategies are present in championship-level chess or Go programs), speech recognition (where statistics has steadily replaced linguistics as the engine of success) and computer vision (where human-designed strategies for breaking down the problem have been replaced by data-driven machine-learning methods).

Another leading figure, Rodney Brooks, replied with his own blog post,[279] countering that each of AI's notable successes "have all required substantial amounts of human ingenuity"; general methods alone were not enough. This framing is something of a turnaround for Brooks, one of the founders of behavior-based robotics, as he is well known for trying to build intelligent behaviors from simple methods of interacting with the complex world.

This fundamental debate has dogged the field from the very start. In the 1960s and 1970s, founders of the field—greats like Nobel prize winner Herbert Simon and Turing Award winner Alan Newell—tried to build general-purpose methods. But such methods were easily surpassed with the specialized hand-coded knowledge poured into expert systems in the 1980s. The pendulum swung back in the 2010s, when the addition of big data and faster processors allowed general-purpose methods like deep learning to outperform specialized hand-tuned methods. But now, in the 2020s, these general methods are running into limits, and many in the field are questioning how we best make continued progress.

One limitation is the end of Moore's Law. We can no longer expect processing power to double every two years or so, as it has since the beginning of the computer age.[280] After all, every exponential trend in the real world must eventually wind down. In this case, we are starting to run into quantum limits and development costs. One reaction to this constraint is to build specialized hardware, optimized to support AI software. Google's Tensor Processing Units (TPUs) are an example of this specialized approach.[281]

Another limit is model size. A record was set in May 2020 by GPT-3, a neural network language model with 175 billion parameters. GPT-3 is more than ten times the size of the previous largest language model, Turing NLG, introduced just three months earlier. A team at

## 5. WELL, WHADDYA SAY?
by Kevin G. Der / Will Shortz

© 2021, American Crossword Puzzle Tournament

37 Across: Fans of nature?

OpenAI calculated[282] that, since 2012, the amount of computation used in the largest AI training runs has been increasing exponentially, with a doubling time of roughly three-and-half months. Even if Moore's Law were to continue, such an accelerated rate of growth in model size is unsupportable.

Sustainability constitutes an additional limit. Many within the field are becoming aware of the carbon footprint of building such large models. There are significant environmental costs. A 2015 report[283] cautioned that computing and communications technologies could consume half of the electricity produced globally by 2030 if data centers cannot be made more efficient. Fortunately, companies have been making data centers more efficient faster than they have been increasing in size and are mostly switching to green energy, keeping their carbon footprint stable over the past five years. GPT-3 cost millions of dollars to build, but offsetting the $CO_2$ produced to train it would cost only a few thousand dollars. Microsoft, which provided the compute for GPT-3, has been carbon neutral since 2012 and has made commitments to further environmental improvements in the years to come.[284]

Availability of data holds things back. Deep learning methods often need data sets with tens of thousands, hundreds of thousands, or even millions of examples. There are plenty of problems where we don't have such large data sets. We might want to build models to predict the success of heart-lung transplants, but there is limited data available to train them—the number of such operations that have been performed worldwide is just a few hundred. In addition, machine-learning methods like deep learning struggle to work on data that falls outside their training distribution.

Existing systems are also quite brittle. Human intelligence often degrades gracefully. But recent adversarial attacks demonstrate that current AI methods are often prone to error when used in new contexts.[285] We can change a single pixel in the input to an object-recognition system and it suddenly classifies a bus as a banana. Human vision can, of course, be easily tricked, but it is in very different ways to computer vision systems. Clearly, AI is "seeing" the world idiosyncratically compared to human beings. Despite significant research on making systems more robust, adversarial methods continue to succeed and systems remain brittle and unpredictable.

And a final limit is semantic. AI methods tend to be very statistical and "understand" the world in quite different ways from humans. Google Translate will happily use deep learning to translate "the keyboard is dead" and "the keyboard is alive" word by word without pausing for thought, as you might, about why the metaphor works in the former but not the latter.

The limitations above are starting to drive researchers back into designing specialized components of their systems to try to work around them. The recent dominance of deep learning may be coming to an end.

What, then, do we make of this pendulum that has swung backwards and forwards, from nature to nurture and back to nature multiple times? As is very often the case, the answer is perhaps likely to be found somewhere in between. Either extreme position is a straw man. Indeed, even at "peak nurture," we find that learning systems benefit from using the right architecture for the right job—transformers for language ❯ SEE SQ5.A and convolutional nets for vision, say. Researchers are constantly using their insight to identify the most effective learning methods for any given problem. So, just as psychologists recognize the role of both nature and nurture in human behavior, AI researchers will likely need to embrace both general- and special-purpose hand-coded methods, as well as ever

❯ SQ12.B

> AI researchers will likely need to embrace both general- and special-purpose hand-coded methods, as well as ever faster processors and bigger data.

# WQ1. HOW ARE AI-DRIVEN PREDICTIONS MADE IN HIGH-STAKES PUBLIC CONTEXTS, AND WHAT SOCIAL, ORGANIZATIONAL, AND PRACTICAL CONSIDERATIONS MUST POLICYMAKERS CONSIDER IN THEIR IMPLEMENTATION AND GOVERNANCE?[287]

faster processors and bigger data.

Indeed, the best progress on the long-term goals of replicating intelligent behaviors in machines may be achieved with methods that combine the best of both these worlds.[286] The burgeoning area of neurosymbolic AI, which unites classical symbolic approaches to AI with the more data-driven neural approaches, may be where the most progress towards the AI dream is seen over the next decade.

Researchers are developing predictive systems to respond to contentious and complex public problems. These AI systems emerge across all types of domains, including criminal justice, healthcare, education and social services—high-stakes contexts that can impact quality of life in material ways. Which students do we think will succeed in college? Which defendants do we predict will turn up for a future court date? Who do we believe will benefit the most from a housing subsidy?

We know that the development of a predictive system in the real world is more than a technical project; it is a political one, and its success is greatly influenced by how a system is or is not integrated into existing decision-making processes, policies, and institutions. This integration depends on the specific sociological, economic and political context. To ensure that these systems are used responsibly when making high-impact decisions, it is essential to build a shared understanding of the types of sociotechnical challenges that recur across different real-world experiences. Understanding these processes should also help us build tools that effectively capture both human and machine expertise.

Below are some core socio-technical considerations that scholars and practitioners should pay attention to over the next decade.

## Problem Formalization

What problem is AI being used to solve? What is being predicted or optimized for, and by whom? Is AI the only or best way of addressing the problem before us? Are there other problems we might instead turn our attention to addressing? Which aspects of a problem can we address using AI, and which can't we? The ways we define and formalize prediction problems shape how an algorithmic system looks and functions. Although the act of problem definition can easily be taken for granted as outside the purview of inquiry for AI practitioners, it often takes place incrementally as a system is built.[288] In the best cases, it can be an opening for public participation, as the process of refining vague policy goals and assumptions brings competing values and priorities to the fore.[289]

Even subtle differences in problem definition can significantly change resulting policies. Tools used to apportion scarce resources like access to permanent housing can have quite different impacts depending on whether "need" is understood as the likelihood of future public

> The next generation of AI researchers and practitioners should be trained to give problem formalization critical attention.

assistance, the probability of re-entry into homeless services, or something else.[290] In the context of financial subsidies for families to prevent income shocks, slightly different formalizations of the same policy goal can reverse the order in which families are prioritized to receive a subsidy.[291]

Problem definition processes must also stop short of assuming that a technical intervention is warranted in the first place.[292] Technical solutions may themselves be part of the problem, ▶ SEE SQ10.B particularly if they mask the root causes of social inequities and potential nontechnical solutions.[293] The next generation of AI researchers and practitioners should be trained to give problem formalization critical attention. Meanwhile, practitioners might use their tools to study higher levels of the power hierarchy, using AI to predict the behaviors of powerful public-sector institutions and actors, not solely the least privileged among us.[294]

# Integration, Not Deployment

We often use the term "deployment" to refer to the implementation of an AI system in the real world. However, deployment carries the connotation of implementing a more or less ready-made technical system, without regard for specific local needs or conditions. Researchers have described this approach as "context-less dropping in."[295] The most successful predictive systems are not dropped in but are thoughtfully integrated into existing social and organizational environments and practices. From the outset, AI practitioners and decision-makers must consider the existing organizational dynamics, occupational incentives, behavioral norms, economic motivations, and institutional processes that will determine how a system is used and responded to. These considerations become even more important when we attempt to make predictive models function equally well across different jurisdictions and contexts that may have different policy objectives and implementation challenges.

As in other algorithmic systems, the kinds of visible and invisible labor the system depends on are key concerns in public decision-making.[296] Frontline workers—like judges, caseworkers, and law enforcement officers who interact directly with an algorithmic system—ultimately shape its effects, and developers must prioritize frontline workers' knowledge and interests for integration to be successful. Resource constraints also matter: How will systems be maintained and updated over time?[297] How can systems be made to correct course if they don't work as expected? The answers to these questions depend on contextual social knowledge as much as technical know-how.

Collaborative design with stakeholders like frontline workers and affected communities can be a promising way to address these concerns, though it's crucial to ensure that such participation is not tokenistic.[298] Systems may also benefit when developers document both the social contexts in which a model is likely to perform successfully and the organizational and institutional processes that led to its development and integration. This practice borrows the logic of similar recent efforts to better document the data used to train machine-learning models as well as documenting the resulting models themselves.[299] Formalizing these considerations might make it easier to determine whether a system can be easily adapted from one setting to another.

A heartbreaking example of how the integration process can go wrong is found in the use of AI to help treat patients with COVID-19. AI systems were among the first to detect the outbreak,[300] and many research teams sprang into action to find ways of using AI technology to identify patterns and recommend treatments. Ultimately, these efforts were deemed unsuccessful as a combination of difficulty in sharing high quality data, a lack of expertise at the intersection of medicine and data science, and over optimism in the technology resulted in systems "not fit for clinical use."[301]

## Quantitative Analyses

### Prediction Quality

The following tables show the evaluation result for different attributes/categories. Both models perform fairly (< 5% performance differences between categories) on our targeted Active Single Person Image Set.

COCO Val2017 Single Person Image Set

| Gender | % dataset | Keypoint mAP (Lightning) | Keypoint mAP (Thunder) |
|--------|-----------|--------------------------|------------------------|
| Male | 63.1 | 67.4 | 78.7 |
| Female | 36.9 | 65.4 | 76.6 |

| Age | % dataset | Keypoint mAP (Lightning) | Keypoint mAP (Thunder) |
|-----|-----------|--------------------------|------------------------|
| Young | 72.2 | 65.6 | 76.6 |
| Middle-age | 17.1 | 68.0 | 78.0 |
| Old | 10.7 | 72.1 | 81.5 |

| Skin Tone | % dataset | Keypoint mAP (Lightning) | Keypoint mAP (Thunder) |
|-----------|-----------|--------------------------|------------------------|
| Darker | 26.8 | 60.5 | 74.4 |
| Medium | 4.0 | 61.2 | 73.7 |
| Lighter | 69.2 | 74.4 | 82.9 |

Active Single Person Image Set

| Gender | % dataset | Keypoint mAP (Lightning) | Keypoint mAP (Thunder) |
|--------|-----------|--------------------------|------------------------|
| Male | 46.0 | 90.2 | 93.7 |
| Female | 54.0 | 87.8 | 92.3 |

| Age | % dataset | Keypoint mAP (Lightning) | Keypoint mAP (Thunder) |
|-----|-----------|--------------------------|------------------------|
| Young | 87.6 | 89.1 | 93.3 |
| Middle-age | 10.5 | 89.3 | 91.5 |
| Old | 1.9 | 85.7 | 90.0 |

| Skin Tone | % dataset | Keypoint mAP (Lightning) | Keypoint mAP (Thunder) |
|-----------|-----------|--------------------------|------------------------|
| Darker | 15.4 | 89.1 | 93.1 |
| Medium | 2.5 | 92.2 | 93.3 |
| Lighter | 82.1 | 92.9 | 95.4 |

*The use of dataset datasheets and model cards are two recent proposals for documenting the inputs and outputs of machine-learning systems so that they can be used responsibly in applications. This example model card comes from the MoveNet.SinglePose model that predicts body position from images.* From: *https://storage.googleapis.com/movenet/MoveNet.SinglePose%20Model%20Card.pdf.*

## Diverse Governance Practices

Finally, new predictive technologies may demand new public-governance practices. Alongside the production of new technical systems, we need to consider what organizational and policy measures ➤ SEE SQ7 should be put in place to govern the use of such systems in the public sector. New proposals in both the US and the European Union exemplify some potential approaches to AI regulation.[302]

Appropriate measures may include establishing policies that govern data use—determining how data is shared or retained, whether it can be publicly accessed, and the uses to which it may be put, for instance—as well as standards around system adoption and procurement. Some researchers have proposed implementing algorithmic impact assessments ➤ SEE SQ7.C akin to environmental impact assessments.[303] Matters are further complicated by questions about jurisdiction and the imposition of algorithmic objectives at a state or regional level that are inconsistent with the goals held by local decision-makers.[304]

A related governance concern is how change will be managed: How, when, and by whom should systems be audited to assess their impacts?[305] Should models be given expiration dates to ensure that they are not creating predictions that are hopelessly outdated? The COVID-19 pandemic is a highly visible example of how changing conditions invalidate models—patterns of product demands, highway traffic[306], stock market trends[307], emergency-room usage, and even air quality changed rapidly, potentially invalidating models trained on prior data about these dynamics.

The growth of facial-recognition technologies ➤ SEE SQ6.C illustrates the diversity of governance strategies that states and municipalities are beginning to develop around AI systems. Current governance approaches range from accuracy- or agency-based restrictions on use, to process-oriented rules about training and procurement processes, to moratoria and outright bans.[308] The diversity of approaches to governance around facial recognition may foreshadow how governments seek to address other types of AI systems in the coming decade.

Successfully integrating AI into high-stakes public decision-making contexts requires difficult work, deep and multidisciplinary understanding of the problem and context, cultivation of meaningful relationships with practitioners and affected communities, and a nuanced understanding of the limitations of technical approaches. It also requires sensitivity to the politics surrounding these high-stakes applications, as AI increasingly mediates competing political interests and moral commitments.

> Successfully integrating AI into high-stakes public decision-making contexts requires difficult work, deep and multidisciplinary understanding of the problem and context, cultivation of meaningful relationships with practitioners and affected communities, and a nuanced understanding of the limitations of technical approaches.

# WQ2. WHAT ARE THE MOST PRESSING CHALLENGES AND SIGNIFICANT OPPORTUNITIES IN THE USE OF ARTIFICIAL INTELLIGENCE TO PROVIDE PHYSICAL AND EMOTIONAL CARE TO PEOPLE IN NEED?[309]

AI devices are now moving into some of our most intimate settings, augmenting, and in some cases replacing, human-given care. Smart home devices can give Alzheimer's patients medication reminders, pet avatars and humanoid robots can offer companionship, and chatbots can help veterans living with PTSD treat their mental health.

These intimate forms of AI caregiving challenge how we think of core human values, like privacy, compassion, trust, and the very idea of care itself. In doing so, they raise questions about what conceptions of care and well-being should be encoded within these technologies and whether technology can be purpose-built with certain capabilities of care—such as compassion, responsiveness, and trustworthiness. We can even ask whether there are forms of care that intimate AI is better positioned to give than a human caregiver.

From smart home sensors to care robots, new markets in intimate AI also urge us to examine complex challenges surrounding the automation of care work, such as the continual data collection required for attentive care and the labor concerns of replacing human caregivers with autonomous systems. We must ask if these devices perpetuate racial or gender stereotypes, as when social companion robots for the elderly speak only in a female voice, and whether AI technologies in health and welfare serve mainly to discharge us of our responsibilities towards people in need. The demand for physical separation and sterility during the COVID-19 pandemic has only brought new questions about the role of technological mediation, whether by a robot or a digital assistant whose help was called for in a patient's dying moments.[310]

---

308 https://ainowinstitute.org/regulatingbiometrics-spivack-garvie.pdf
309 This topic was the subject of a two-day workshop entitled "Coding Caring," convened at Stanford University in May 2019. The workshop was organized with support from AI100, the Presence Artificial Intelligence in Medicine: Inclusion & Equity (AiMIE) 2018 Seed Grants, and the McCoy Family Center for Ethics in Society at Stanford University. The discussion involved practitioners from the health and AI industries along with designers and researchers bringing a diversity of analytical frameworks, including feminist ethics of care, bioethics, political theory, postcolonial theory, and labor theory. The workshop was co-organized by Thomas Arnold, Morgan Currie, Andrew Elder, Jessica Feldman, Johannes Himmelreich, and Fay Niker. More information on the event can be found at https://ai100.stanford.edu/sites/g/files/sbiybj9861/f/coding_caring_workshop_report_1000w_0.pdf.

## Autonomous Systems Are Enhancing Human-to-Human Care

AI offers extraordinary tools to support caregiving and increase the autonomy and well-being of those in need. AI-analyzed x-rays now bring a higher degree of confidence to medical diagnoses. AI is starting to help clinicians diagnose and treat wounds faster and with more accuracy, while phone apps with AI capabilities allow patients to monitor chronic wounds from home— an especially useful function for patients in rural settings. Researchers are developing AI-powered wheelchairs to help children navigate with more independence and avoid obstacles, while trials have found that robot interventions improve language and social functioning in children with autism, who may feel comfortable around the robots' consistent, repetitive behavior.[311]

Support for aging in place and institutional care will benefit from these technological interventions, which offer physical assistance and companionship as well as health and safety monitoring. Some patients may even express a preference for robotic care in contexts where privacy is an acute concern, as with intimate bodily functions or other activities where a non-judgmental helper may preserve privacy or dignity. These technologies can greatly improve patients' lives, but they can also reshape traditional caring relationships. By mediating or replacing human-to-human care, their use raises questions of how care receivers form relationships with AI providers. More broadly, the introduction of AI also requires us to ask when AI care can augment human caring in ways that meaningfully address the inadequacies of current care systems. Might there also be certain situations when AI could offer short-term solutions, but reduce important care infrastructures—whether family or institutional—in the long-term?

## Autonomous Systems Should Not Replace Human-Care Relationships

While some occupational knowledge can be standardized and codified, most care relations require improvisation and an understanding of specific contexts that will be difficult, if not impossible, to generalize in code. Radiologists will remain in charge of cancer diagnoses, and in elder care, particularly for dementia patients, companion robots will not replace the human decision-makers who increase a patient's comfort through intimate knowledge of their conditions and needs. The use of AI technologies in caregiving should aim to supplement or augment existing caring relationships, not replace them, and should be integrated in ways that respect and sustain those relationships.

Take the example of care robots or personal assistants replacing human caregiving. A major concern is that these technologies offer an illusory form of care and reciprocity. According to ethicist Nell Noddings, care is not simply the fulfilling of an instrumental need or outcome; caring is a relational act between caregiver and care receiver that requires time and commitment, presence, and attention, and should foster the care receiver's independence and self-determination.[312] Good care demands respect and dignity, things that we simply do not know how to code into procedural algorithms.

While AI can now be trained to be more responsive and dynamic than in the past, care robots still have limited ability to interpret complex situations. They have less capacity for open-ended improvisation and no agency beyond their designed technical attributes. These systems are not moral agents, and they do not make sacrifices through their care work. Although a person might feel they were being cared for by a robotic caregiver, the emotions associated with the relationship would not meet many criteria of human flourishing. There is also concern that the effects of artificial sentiment could be deceptive or manipulative. And while human caregivers may at times be disinclined to deliver good care (considering that the caring process also places great demands on the caregivers), and while AI could at times offer a more dignified caring process that well-informed patients prefer, in many situations these technologies will not replace the standard of genuine human-to-human care.

## Autonomous Care Technologies Produce New Challenges

AI is likely to change norms around care in ways that could introduce new harms. If technology leads us to believe that challenges in caregiving can be solved through technical ➤ SEE SQ10.B rather than social or economic solutions, for instance, we could increasingly absolve care practitioners, family members and state service providers from their responsibilities towards care receivers. Replacing human judgment with AI systems may also lead to the loss of occupational knowledge in the caregiving field.

➤ WQ2.A

Another important ethical concern, particularly around smart homes or robot companions, is their invasive surveillance of patients, particularly in the intimate sphere of the home. Intuition Robotics' social companion robot ElliQ,[313] for instance, allows relatives to monitor a senior family member living alone. Using an app, relatives can check on their loved one, access a networked camera, and check sensors that track activity. Although it is intended for safety and companionship, this technology can give control of data to family members, rather than the elderly themselves, which raises questions around privacy and consent. Similarly, Amazon Echo's Care Hub and Google's Nest Hub support the ability to monitor elderly family members' activity feeds.

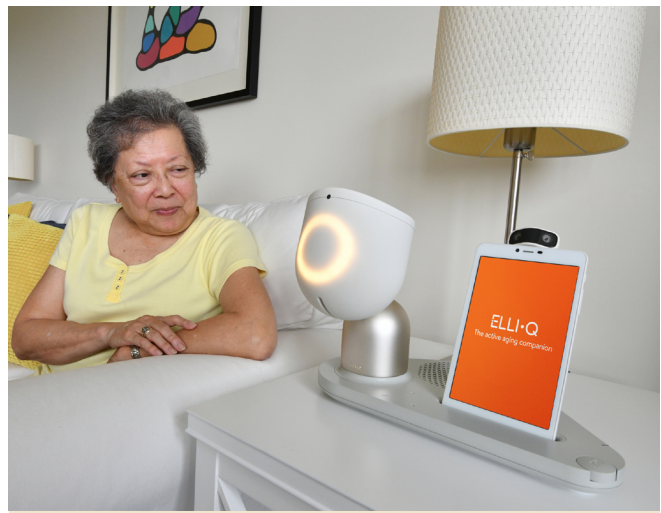

*Caption: In-home sensors and robots are on the rise, offering new ways to provide support and care, but also raising concerns about the negative impacts of pervasive surveillance. The ElliQ robot is shown here. From:* [https://blog.elliq.com/hubfs/Beta%20user%20ElliQ.png](https://blog.elliq.com/hubfs/Beta%20user%20ElliQ.png)

# Caring AI Should Be Led by Social Values, Not the Market

The number of people around the world aged 80 or over will grow from 143 million in 2019 to 426 million in 2050, according to the UN.[314] By that date, demographic projections in Europe and North America also expect one in four people to be aged 65 or over. Meanwhile, researchers predict a global shortage of caregivers. In Japan, the shortfall is predicted to be 370,000 care workers by 2025,[315] while the EU's anticipated shortfall is 4 million care workers by 2030.[316] This deficiency is in part due to the occupation's relatively low social status, as caregivers typically receive low pay and are devalued compared to other healthcare professionals.[317] In this landscape, robots become a tempting option to address the widening gulf between care needs and services.

However, AI caring technologies should not be market-led technical fixes for these problems, which are largely economic, political, and cultural, and innovation and convenience through automation should not come at the expense of authentic care. Instead, regulators, developers, and funders should put resources into supporting better human care. To this end, there is an urgent need to slow down and subject care technologies to regulating bodies and pre-market approval that can intervene in the technical designs and policies around AI care. Too often, oversight has been neglected during implementation; such was the case when medical facilities adopted electronic medical record systems with little input from doctors and nurses,[318] and it is likely to be the case with AI-based caregiving. While AI applications may seem inevitable, oversight can put ethical practices into place prior to real-world use.

> There is an urgent need to slow down and subject care technologies to regulating bodies and pre-market approval that can intervene in the technical designs and policies around AI care.

Further, while many societies prize technological innovation, caregiving is too often stigmatized and left to the private sphere of women. Today, caregiving is also a racialized and class-based practice that remains invisible and underfunded. AI should address, rather than reinforce, these inequities. An ethics-of-care approach, in particular, directs us to consider how AI technologies can be part of economic structures that honor and support care work, rather than to create new forms of invisible labor through their maintenance.[319]

Finally, should we place certain demands on the role of the designer in the caring ecosystem? Is the engineer of a caring technology taking part in care work? Is the engineer properly placed to understand the context of use, and does the engineering process incorporate diverse voices? Does the design process involve the input of caregivers and care receivers? The development of any caregiving technology should incorporate the perspectives of care receivers and caregivers, while advocating for designs that are sensitive to cross-cultural differences and potential biases.

---

# CONCLUSIONS

The field of artificial intelligence has made remarkable progress in the past five years and is having real-world impact on people, institutions and culture. The ability of computer programs to perform sophisticated language- and image-processing tasks, core problems that have driven the field since its birth in the 1950s, has advanced significantly. Although the current state of AI technology is still far short of the field's founding aspiration of recreating full human-like intelligence in machines, research and development teams are leveraging these advances and incorporating them into society-facing applications. For example, the use of AI techniques in healthcare is becoming a reality, and the brain sciences are both a beneficiary of and a contributor to AI advances. Old and new companies are investing money and attention to varying degrees to find ways to build on this progress and provide services that scale in unprecedented ways.

The field's successes have led to an inflection point: It is now urgent to think seriously about the downsides and risks that the broad application of AI is revealing. The increasing capacity to automate decisions at scale is a double-edged sword; intentional deepfakes or simply unaccountable algorithms making mission-critical recommendations can result in people being misled, discriminated against, and even physically harmed. Algorithms trained on historical data are disposed to reinforce and even exacerbate existing biases and inequalities. Whereas AI research has traditionally been the purview of computer scientists and researchers studying cognitive processes, it has become clear that *all* areas of human inquiry, especially the social sciences, need to be included in a broader conversation about the future of the field. Minimizing the negative impacts on society and enhancing the positive requires more than one-shot technological solutions; keeping AI on track for positive outcomes relevant to society requires ongoing engagement and continual attention.

Looking ahead, a number of important steps need to be taken. Governments play a critical role in shaping the development and application of AI, and they have been rapidly adjusting to acknowledge the importance of the technology to science, economics, and the process of governing itself. But government institutions are still behind the curve, and sustained investment of time and resources will be needed to meet the challenges posed by rapidly evolving technology. In addition to regulating the most influential aspects of AI applications on society, governments need to look ahead to ensure the creation of informed communities. Incorporating understanding of AI concepts and implications into K-12 education is an example of a needed step to help prepare the next generation to live in and contribute to an equitable AI-infused world.

The AI research community itself has a critical role to play in this regard, learning how to share important trends and findings with the public in informative and actionable ways, free of hype and clear about the dangers and unintended consequences along with the opportunities and benefits. AI researchers should also recognize that complete autonomy is not the eventual goal for AI systems. Our strength as a species comes from our ability to work together and accomplish more than any of us could alone. AI needs to be incorporated into that community-wide system, with clear lines of communication between human and automated decision-makers. At the end of the day, the success of the field will be measured by how it has empowered all people, not by how efficiently machines devalue the very people we are trying to help.

# PANEL MEMBER BIOS

**MICHAEL L. LITTMAN (STUDY PANEL CHAIR)**
Michael holds a Ph.D. in computer science. He is the Royce Family Professor of Teaching Excellence in Computer Science at Brown University (US), where he studies reinforcement learning and its interaction with people. A past chair of the Association for the Advancement of Artificial Intelligence (AAAI) conference and the International Conference on Machine Learning (ICML), he is a fellow of the ACM and AAAI, and is a 2020-2021 AAAS Leshner Fellow on public engagement in AI. He is currently a member of AI Hub and was communications co-chair at the Neural Information Processing Systems conference.

**IFEOMA AJUNWA**
Ifeoma holds a Ph.D. in sociology, as well as a J.D. She is an associate professor in the University of North Carolina School of Law (US) and founding director of their artificial intelligence and decision-making research program. Her research interests are at the intersection of law and technology, with a particular focus on the ethical governance of workplace technologies, as well as on diversity and inclusion in the labor market and the workplace. She is a 2019 recipient of the NSF CAREER Award and a 2018 recipient of the Derrick A. Bell Award.

**GUY BERGER**
Guy holds a Ph.D. in economics and a degree in math. He is principal economist at LinkedIn (US). His research includes analysis of the data at LinkedIn, such as identifying the top countries where people are moving to for work, the industries most likely to hire career-switchers, and the most in-demand skills. He warns that the need for soft skills will skyrocket in the future of robots and automation. He works with companies and governments to help people find jobs.

**CRAIG BOUTILIER**
Craig holds a Ph.D. in computer science. He is principal scientist at Google (US). He previously worked in academia for 24 years. He works on various aspects of decision-making under uncertainty, social choice and mechanism design with a current focus on sequential decision models and their application to recommender systems. He is a Fellow of the Royal Society of Canada (RSC), the Association for Computing Machinery (ACM) and the Association for the Advancement of Artificial Intelligence (AAAI). He received the 2018 ACM/SIGAI Autonomous Agents Research Award and served as Editor-in-Chief of the Journal of Artificial Intelligence Research.

**MORGAN CURRIE**
Morgan is a lecturer in data and society in science, technology and innovation studies at the University of Edinburgh (UK). Her research looks at open and administrative data, automation in the welfare state, activists' data practices, social justice and the city, web maps and cultural mapping. She was awarded a Ph.D. in information studies at the University of California, Los Angeles, and a master's degree in new media at the University of Amsterdam, before holding a postdoc in the Digital Civil Society Lab at Stanford University. Morgan was one of the organizers of the AI100 Coding Caring workshop.

**FINALE DOSHI-VELEZ**
Finale holds a Ph.D. in computer science. She is a John L. Loeb associate professor of computer science at the Harvard Paulson School of Engineering and Applied Sciences (US). Her research spans probabilistic models, reinforcement learning, and interpretable machine learning, largely focused toward healthcare applications. The undergraduate machine-learning course she teaches includes both technical material and significant exposure to machine learning with a sociotechnical system perspective.

GILLIAN HADFIELD

Gillian holds a Ph.D. in economics and a J.D. She is a professor of both law and strategic management at the University of Toronto (Canada), where she holds the Schwartz Reisman Chair in Technology and Society and is inaugural director of the Schwartz Reisman Institute for Technology and Society. She served on the Steering Committee for the AAAI AI Ethics and Society Conference and was a co-organizer of the first NeurIPS Workshop on Cooperative AI. Her research is focused on what she calls the science of normativity—modeling societal norms—as well as the design of legal and regulatory systems, and contract design and law.

MICHAEL C. HOROWITZ

Michael holds a Ph.D. in government and a degree in political science. He is professor of political science and the interim director of Perry World House at the University of Pennsylvania (US) and is co-author of the book *Why Leaders Fight*. Michael previously worked for the Office of the Undersecretary of Defense for Policy in the Department of Defense. His research interests include technology and global politics, military innovation, the role of leaders in international politics, and forecasting.

CHARLES ISBELL

Charles holds a Ph.D. in computer science. His research focuses on applying statistical machine learning to building autonomous agents that must live and interact with large numbers of other intelligent agents, some of whom may be human. Charles received the Black Engineer of the Year Modern Day Technology Leader Award (2009). He is the John P. Imlay, Jr. Dean of the College of Computing at Georgia Tech (US) and is a fellow of Association for the Advancement of Artificial Intelligence (AAAI) and the Association for Computing Machinery (ACM). He has presented congressional testimony on his work in both online education and machine learning.

HIROAKI KITANO

Hiroaki holds a Ph.D. in computer science. He is president and CEO of Sony Computer Science Laboratories and Sony AI, as well as a professor at Okinawa Institute of Science and Technology Graduate University (Japan). He carried out research on AI and natural language processing at Carnegie Mellon University and on systems biology at the California Institute of Technology. He is a recipient of the prestigious Computers and Thought Award (1993) honoring early career scientists in AI. Hiroaki is developing AI systems for making scientific discoveries and is closely involved in policy-setting in Japan.

KAREN LEVY

Karen holds a Ph.D. in sociology and a J.D. She is an assistant professor of information science at Cornell University (US) and an associated faculty member at Cornell Law School. Her research considers the legal, organizational, social, and ethical aspects of data-intensive technologies, focused on inequality, care/intimacy, privacy, and labor. Karen is a New America National Fellow and she previously served as a law clerk in the United States Federal Courts. Karen was one of the organizers of the AI100 Prediction in Practice workshop.

TERAH LYONS

Terah holds a degree in social studies with a focus on network theory and complex systems, and her professional work has focused primarily on technology policy. She was the founding executive director of the Partnership on AI and is a former policy advisor to the US chief technology officer in the White House Office of Science and Technology Policy (OSTP), where she helped establish and direct the White House Future of Artificial Intelligence Initiative. Terah currently sits on the steering committee of the AI Index.

MELANIE MITCHELL

Melanie holds a Ph.D. in computer science. She is the Davis Professor of Complexity at the Santa Fe Institute and professor of computer science (currently on leave) at Portland State University (US). Her current research focuses on conceptual abstraction, analogy-making, and visual recognition in AI systems. Melanie is the author or editor of six books on artificial intelligence, cognitive science, and complex systems, the latest being *Artificial Intelligence: A Guide for Thinking Humans*.

JULIE SHAH

Julie holds a Ph.D. in autonomous systems. She is an associate professor in the department of aeronautics and astronautics at MIT (US) and heads MIT's Interactive Robotics Group. Julie is also an associate dean of social and ethical responsibilities of computing. Her work aims to imagine the future of work by designing collaborative robot teammates that enhance human capability. She has translated her work to manufacturing assembly lines, healthcare applications, transportation and defense. She was also a member of the 2016 AI100 Study Panel and previously worked at Boeing Research and Technology.

STEVEN SLOMAN

Steve holds a Ph.D. in psychology. He is a professor in the cognitive, linguistic, and psychological sciences department at Brown University (US), specializing in higher-level cognition, especially causal reasoning and collective knowledge. He has studied how the systems that constitute thought interact to produce conclusions, conflict, and conversation, and how our interpretation of how the world works influences how we evaluate events and decide what to do. He co-authored a book on collective cognition, *The Knowledge Illusion: Why We Never Think Alone*.

SHANNON VALLOR

Shannon holds a Ph.D. in philosophy. She is the Baillie Gifford Chair in the ethics of data and AI at the University of Edinburgh (UK). Her research focuses on the ethics of AI, robotics, data science, digital media and other emerging technologies. She was a visiting researcher and AI ethicist at Google. She served on the steering committee of ACM/AAAI's AI Ethics and Society conference. Shannon also chairs the Scottish government's Data Delivery Group and the University of Edinburgh's AI and Data Ethics Advisory Board.

TOBY WALSH

Toby holds a Ph.D. in artificial intelligence. He is Scientia Professor of Artificial Intelligence at the University of New South Wales (Australia) and served as scientific director of the information technology research center NICTA (now Data61). His research addresses aspects of AI in automated reasoning, constraint programming, social choice, and game theory. He was editor-in-chief of the *Journal of Artificial Intelligence Research* and helped release an open letter calling for a ban on offensive autonomous weapons. He is the author of two trade books on artificial intelligence.

# ANNOTATIONS ON THE 2016 REPORT

The Study Panel added annotations to the 2016 report to highlight places where comparisons between the two reports were illuminating. The online version of the report includes these annotations as hover text and links. This section summarizes briefly some of the high-level comparisons.

The 2016 panel focused their report on the North American context and considered discussion of defense and military applications of AI to be out of scope. The 2021 report includes several comments about how AI is recognized as influencing and being influenced by geopolitical and international security considerations. These include the observation that emerging regulatory approaches vary across regions, and that AI and the "AI race" are viewed as issues of national security. In addition, sentiments regarding military applications of AI influence the research directions of some scientists. In the US, the defense department's investments in technology have helped spur some of the most important advances in AI in the past five years. The report includes recommendations for further investments in the creation of federal data and computational resources. Finally, numerous countries have begun to develop national AI policy strategies and to legislate the use of AI technologies.

The 2016 report listed a set of challenges associated with the future of AI, including: developing safe and reliable hardware for transportation and service robotics; challenges in "smoothly" interfacing AI with human experts; gaining public trust; overcoming fears of marginalization of humans with respect to employment and the workplace; and diminishing interpersonal interactions (for example, through new entertainment technologies). In contrast, the 2021 report details social and ethical concerns and harms related to the conception, implementation, and deployment of AI technologies. Many of the descriptions of potential harms foreshadowed in 2016 were counterbalanced with abstract descriptions of a different, more positive possible future that could be achieved "through careful deployment." The 2021 report makes clear that many concerns and harms are no longer hypothetical, and are not merely technological problems to be solved. The shift in views regarding social and ethical concerns can be seen in the use of terms like "bias," "privacy, "security," and "safety" in the 2016 and 2021 reports, as highlighted in the annotations.

The page numbers below refer to the <u>pdf version of the 2016 report</u>.

## Page 1

"in the years following the immediately prior report": One of the biggest differences between the 2016 and 2021 reports is that the 2021 report is the first to have an immediately prior report. As such, we're seeing the changes in perspectives in a way that may have been harder to see in 2016.

"compatible ➤ SEE SQ10.A with human cognition ➤ SEE SQ4.A": Stuart Russell, a highly visible member of the AI community and co-author of the most commonly used textbook in the field of AI, wrote a book called *Human Compatible* in 2019. However, his use of the phrase is focused on the alignment problem—building powerful AI systems that have objectives that are consistent with human values—whereas the report authors were probably referring to the more nuts-and-bolts topic of building AI systems that collaborate directly with people. At the end of the day, however, the key to solving the problem of keeping powerful AI systems under humanity's control may indeed begin by framing AI systems more from the perspective of how they augment instead of replace human intelligence.

"connected set of reflections": These annotations are our attempt to make these connections.

"<u>AAAI Asilomar study</u>": Peter Stone, who led the 2016

Study Panel and now serves as the chair of the AI100 Standing Committee was also a participant in the Asilomar Study, as was Standing Committee member Sheila McIlraith. Craig Boutilier is the only member of this year's panel to have also attended the Asilomar meeting.

## Page 2

"Professor Peter Stone": Another connection between the reports is that Prof. Stone now serves as the chair of the AI100 Standing Committee and oversaw the panel's work on the 2021 report. In addition, Prof. Julie Shah served on both Study Panels and led the development of these annotations.

"seventeen-member": The 2021 panel also included 17 people.

"AI ➤ SEE SQ2 and Life in 2030 ➤ SEE SQ10": Whereas the 2016 study was framed around the idea of looking ahead to life in 2030, much of the 2021 study is focused on life in 2021. It talks about advances in AI that have become quite visible to people today and focuses on challenges that are now evident. In a sense, this difference is superficial—both reports are grounded in the technology of their time and attempt to extrapolate on future impacts. However, the difference also captures something important about the change in how AI is viewed in 2021 compared to even just five years earlier. More of society is able to see impacts of AI and thus more of the conversation is grounded in society looking *toward* AI instead of *from* the AI field toward the future. The community of people with expertise in and motivation to talk about AI has broadened considerably and it is natural that this new, more heterogeneous community would first reflect on itself.

## Page 3

"military applications" ➤ SEE SQ7.B: The standing questions for 2021 didn't explicitly call out military applications, though they are covered as one of the topics connecting AI and governance.

## Page 4

"Deep learning" ➤ SEE SQ12.B: Deep learning remains an important driver of visible progress in the field. The limitations of depending on a primarily indirect and empirical basis for defining AI systems are becoming evident, and the pendulum may be starting to swing back toward more explicitly designed systems.

"self-driving cars" ➤ SEE SQ2.E: Efforts are still underway to bring self-driving cars to market with some progress and some realization that the problem is harder and more nuanced than was originally perceived. The 2016 report also acknowledges that these challenges are not to be underestimated.

"healthcare ➤ SEE SQ2.F diagnostics and targeted ➤ SEE SQ10.E treatments" ➤ SEE WQ2: AI technologies for healthcare diagnostics and targeted treatments are successfully employed and offer promising opportunities for further development and positive impact, particularly through augmentation and support (rather than replacement) of healthcare providers. The report also acknowledges the risks of AI solutions deployed in health.

"They will facilitate delivery" ➤ SEE SQ2.D: The 2021 report also views embodied AI as offering significant opportunity and notes the past five years have seen great advances in logistics systems, delivery robots, and emerging applications for "intimate embodied AI" in the home.

"potentially profound positive impacts" ➤ SEE SQ10.C: We remain similarly optimistic. However, society's broader worries about the future in terms of inequity, discrimination, and our ability to work together to address our most significant global challenges are reflected in concerns about the future of AI.

## Page 5

"human labor" ➤ SEE SQ11: Although these concerns persist, it is worth keeping in mind that large-scale economic disruption due to AI is not yet evident and will likely take many years, if it happens at all.

"society approaches AI with a more open mind" ➤ SEE SQ6.A: The 2021 report is less sanguine about open-mindedness. The attitude of "build now, see if it helps later" can be seen as responsible for deploying technology that results in asymmetric harms to vulnerable members of society and technology that quickly becomes embedded into broader systems that make it difficult to roll back. Considering possible negative consequences beforehand has become something the field is trying to embrace more broadly in education and even as part of the process of disseminating basic research results.

# Page 6

"human health ➤ SEE SQ7.A, safety" ➤ SEE SQ9.C: The 2016 report discusses ways in which AI technologies are anticipated to improve public safety and highlights anticipated challenges in the development and fielding of safety technologies—due to hardware, software, or integration—and anticipated challenges in gaining public trust. The 2021 report also highlights potential benefits of AI to public safety, albeit fewer than the 2016 report, but emphasizes governance challenges and highlights the scale and scope of national and international efforts to address concerns about safety, ethical design, and deployment of AI products and services.

"spend heavily" ➤ SEE SQ8.A: These trends have continued, even accelerated, in 2021.

"parking challenges become obsolete" ➤ SEE SQ10.B: The 2021 report highlights the danger of seeing AI as a panacea instead of a tool. The 2016 report includes several passages that could be interpreted as examples of this kind of techno-solutionism.

"As a society, ➤ SEE WQ2.A we are now at a crucial juncture" ➤ SEE SQ6.C: The 2016 report raised the potential for ethical and social issues, including privacy concerns, with emerging technologies. The 2021 report validates those concerns with examples of mass intrusions into the privacy rights of citizens by governments and private companies all over the world, and it references a range of research and corporate investments and emerging governance approaches.

# Page 7

"'general purpose' AI" ➤ SEE SQ5: Both the 2016 and the 2021 reports contend that artificial general intelligence does not yet exist. The 2021 report outlines progress that is underway on three types of capabilities that advance towards, although do not achieve, artificial general intelligence.

"Advances in healthcare" ➤ SEE SQ2.F: Broader adoption and pathways to deployment have indeed accelerated healthcare applications of AI.

# Page 8

"build trust with these communities" ➤ SEE WQ1.A: Framing the issue as one of "trust building" suggests that the main impediment is convincing the communities to accept the help of AI systems. As AI is deployed and analyzed more broadly, it is becoming clear that the process is much more complicated. The process of bringing AI technology to a community requires a great deal of push and pull from all involved parties.

"improved cameras and drones for surveillance" ➤ SEE SQ6.C: Arguably, the recent progress limiting the use of facial-recognition systems in law enforcement stems from a lack of trust in law enforcement, not in AI. (Although inconsistent recognition performance does undermine trust.)

"care must be taken to avoid systematizing human bias" ➤ SEE SQ3.E: The 2016 report identified "bias" in AI tools as a concern to be addressed by developers and mitigated through "well-deployed AI tools," with the caveat that concerns will grow and will resist quick resolution since views on bias are colored by personal experience and value judgements. In contrast, the 2021 report recognizes bias as a sociotechnical challenge that can only be addressed partially by technical solutions.

"AI is poised to replace people" ➤ SEE SQ11.A: Concerns highlighted in the 2016 report regarding the future potential for widespread disruption of the global labor market by AI are viewed in the 2021 report as having been premature, although still worthy of attention. In the 2021 report, the view is that AI has not been responsible for large aggregate economic effects. Both reports share concerns regarding, and advocate a proactive approach to address, inequities in the distribution of benefits to be realized by AI technologies. The 2021 report highlights concerns regarding new forms of "invisible human labor" that AI technologies increasingly depend on.

## Page 9

"collaborate effectively ➤ SEE SQ3.A with people" ➤ SEE SQ4.A: Collaboration remains a grand challenge, even in our understanding of the human cognitive processes that support it.

## Page 10

"A vigorous and informed debate": These debates are well underway.

"removing impediments" ➤ SEE SQ7.A: The 2021 panel allocated considerably less focus to removing impediments and more support for intelligent application of oversight and restrictions to help limit societal harm. (The 2016 report acknowledges "best practices need to be spread, and regulatory regimes adapted," which is in line with what the 2021 report advocates.)

"to ensure that the data": In 2021, there is an appreciation that bias in AI systems stems from issues broader than the data, including all the various ways human choices influence how any machine-learning system is put together. We recommend, for example, Charles Isbell's 2020 keynote talk at the major machine-learning conference Neural Information Processing Systems (NeurIPS).

## Page 11

"the hands of a fortunate few": These concerns remain pressing and are even more widely appreciated than in 2016.

## Page 12

"currently 'hot' areas of AI research" ➤ SEE SQ2: These topics remain popular in the AI research community, along with an explosion of work in algorithmic fairness and associated topics at the boundary between AI and society.

"Artificial intelligence is": An alternative definition is that artificial intelligence is about getting a machine to carry out behaviors that we think of as requiring intelligence. This view is useful in that it doesn't put a great deal of emphasis on the specifics of the machine or the technique used to create the behavior. It also captures an important yet frustrating aspect of artificial intelligence—once a machine can carry out a behavior, we tend to stop thinking of it as something that requires intelligence. Real-time navigation aids that decide when and how to describe upcoming turns to guide you to your destination are not thought of as AI, even within the field. But there's no question that it would have been considered an AI problem just a few decades ago. This phenomenon is known as the "AI Effect," as mentioned in the 2016 report.

## Page 13

"beat human players at the game of chess" ➤ SEE SQ2.C: Today, chess programs that can handily outplay the best human masters of all time are referred to as "chess engines" and are used less as an opponent and more as an analysis tool for people improving their play. The field now concentrates more on games that are considerably more challenging for machines than chess turned out to be.

## Page 15

"such as audio, speech, and natural language processing" ➤ SEE SQ2.A: Deep learning remains a significant driver of recent successes in AI. The 2016 report notes that "inroads" were being made in applying deep networks to natural language processing—programs that solve problems related to the meaning of text. These inroads became some of the most dramatic advances in AI, with large-scale neural language models like GPT-3 pushing the boundaries on a wide range of problems.

"Computer vision is currently the most prominent" ➤ SEE SQ2: In addition to incremental advances in the areas listed here, the last five years have seen application of deep learning to simultaneous *combinations* of these areas—like text and images, and reinforcement learning and robotics.

## Page 16

"Research on collaborative systems" ➤ SEE SQ3.A: Collaborative systems have not experienced the same increases in attention and flashy successes as the areas listed above. In part, that is because they include people, and people are complicated. At this point, it seems likely that human-AI interaction research will receive increased funding and development energy as the most consequential and difficult problems in AI lay at this interface.

"agents to augment human abilities" ➤ SEE SQ4.A: AI research into cooperative games is now being emphasized, looking for similar "successes" as those realized in competitive game performance.

## Page 17

"developing systems that are human-aware" ➤ SEE SQ4.A: The 2016 report emphasizes the potential value of human-AI interaction across a number of applications, with primary focus on developing AI that fits to the human and eases frictions in interacting with AI technologies. The 2021 report reinforces this growing interest with its focus on applications and research challenges associated with cooperation and collaboration between humans and AI systems.

"reemergence" ➤ SEE SQ12.B: The 2021 panel also believes the future will bring more attention to the integration of classical model-oriented approaches with the more recent data-driven learning approaches.

## Page 18

"eight of them: transportation ➤ SEE SQ2.E; home/ service robotics ➤ SEE SQ2.D; healthcare ➤ SEE SQ2.F; education ➤ SEE SQ6.A; low-resource communities; public safety and security; employment and workplace; and entertainment" ➤ SEE WQ1: Of these eight, the 2021 report covers advances in transportation, home/service robots, and healthcare. Education is mentioned primarily in terms of the need for education about AI. The use of AI in low-resource communities, entertainment (in the form of social networks), and public safety and security appear primarily as areas of concern; and the topic of employment and workplace is not covered beyond its relevance to economics. There is additional discussion of financial applications and recommendation systems.

"Autonomous transportation will soon be commonplace": Rodney Brooks, the robotics pioneer and 2016 AI100 panelist who helped bring robotic vacuum cleaning to the home, predicted that AI-driven taxi services will appear in 50 of the 100 largest cities no earlier than 2028—not in time for the next AI100 report but potentially before the 2030 purview of the 2016 report.

"suddenly": From the vantage point of 2021, a transition to self-driving cars no longer seems like it will be sudden. Gradual roll outs of monitored, self-driving taxis and commercial delivery vehicles will take place in different cities and at different scales, providing valuable experience about the practical strengths and limitations of the technology before they become ubiquitous. For a sense of scale, expect something less like the rapid switch to smartphones and more like the gradual switch to flat-screen TVs.

# Page 19

"gradually": Advanced Driver Assistance Systems (ADAS) are much more commonplace today than they were in 2016. For one example, AI systems that analyze sensor information to determine if another car is in your blindspot were available in 30% of new cars in 2016 but are in 90% today.

# Page 20

"semi-autonomous approach is sustainable": Indeed, the three Tesla drivers who have died while using Autopilot since 2016 all involved collisions that are easy for an alert driver to see but hard for the car's AI system to detect.

"by 2020": That prediction is overly optimistic, in retrospect.

"how much better": A self-driving shuttle pilot program in one US city, for example, found that human attendants frequently took over the task of driving, especially during left turns in traffic. In addition, drizzle and hardware failures would often leave the shuttle unable to sense its surroundings, preventing its autonomous use. For a variety of reasons, the technology hasn't become robust enough to be used in typical settings. The threshold for acceptance may be much further away than was believed in 2016.

# Page 21

"eliminate the need": Amara's Law says we tend to overestimate the effect of a technology in the short run and underestimate the effect in the long run. These long-term effects are still well in the future but could indeed be quite significant.

# Page 23

"Segway": For technology enthusiasts, the end of the Segway in summer 2020 was a sad occasion. Given the life-and-city-changing aspirations of the technology—similar to those predicted for AI—it also serves as a valuable lesson that we don't always understand the broader influences of technology adoption.

"Uber and Lyft": These services are rarely thought of as AI, but they make a tremendous number of subtle tradeoffs and extrapolations that are driven by machine learning, combinatorial optimization, and other technologies central to the field.

# Page 24

"Special purpose ➤ SEE WQ2 robots will ➤ SEE SQ7.A deliver packages" ➤ SEE SQ2.D: The 2021 report sees important progress in AI systems integration challenges and highlights regulatory preparations as well as social and ethical concerns regarding widespread deployment of the systems.

"difficulty of creating reliable, market-ready hardware": Research, development, and deployment of self-driving cars or autonomous vehicles are still recognized as a rapidly developing application area. Although progress towards fully autonomous vehicles has not lived up to the expectations of some, the challenges were foreshadowed in the 2016 report.

"not materialized": This state of affairs has not changed as of 2021.

# Page 25

"social interaction" ➤ SEE WQ2: Partly, the focus on social interaction has been because it is easier to "fake" social interation than physical manipulation. That is, while understanding people and interacting with them at a deep level is almost certainly more challenging than picking up objects, there is value in providing simple conversational interfaces, which can be built using current technology. There are significant concerns about how to ethically deploy social care robots, however, in part because people are relatively easily taken in by the trappings of sociality.

# Page 26

"current healthcare delivery" ➤ SEE SQ9: The 2021 report identifies a broad range of carefully targeted AI healthcare applications that are becoming prevalent or are poised to become prevalent.

## Page 27

"But several barriers have limited progress to date": An additional issue is that current deep learning systems, even as they deliver high accuracy on the training images they are given, can produce wildly varying diagnoses for individual patients. When the same patient could be diagnosed as having pneumonia or cardiomegaly, depending on the random initialization of the learning algorithm before training, it arguably violates established medical ethics principles..

## Page 30

"technology acceptance among the elderly" ➤ SEE WQ2: Progress on physical assistance for elder care has been slow. It is still viewed as a promising future use case of AI, with caveats regarding social and ethical considerations brought forward by empirical research studying the development and implementation of such technologies.

"Sharing of information" ➤ SEE WQ2.A: As these technologies have become more readily available, it is becoming clear that the issues of sharing private information with family members or even healthcare professionals are complex.

## Page 34

"online ➤ SEE SQ2.H resources" ➤ SEE SQ9.B: One model that may prove helpful is for students to learn from pre-packaged video (no AI needed) but to get help from AI systems (improved versions of search and recommendation) to identify the most useful video for their specific needs.

## Page 36

"inherently more easily audited" ➤ SEE SQ3.F: While it is true that "opening up" an AI system and looking at how it makes its decisions is much easier than carrying out the same operation on a human being, human beings have an advantage that we can ask them what they are doing. In the past five years, explainable AI systems that can justify their decisions have received a great deal of attention, but progress has been slow. In many cases, it appears that the learned systems simply have no meaningful justification for their actions. Alternative methods for building AI approaches that are explainable from the ground up may be needed to achieve the goal of making AI systems accountable.

"low quality of event identification": This issue remains challenging as the events of interest are extremely rare and current machine-learning-based programs for classifying events need many, many realistic examples for accurate training.

## Page 37

"first tool pointing toward predictive policing" ➤ SEE SQ10.C: These technologies have become more visible in the last five years, giving more people opportunities to raise concerns.

"impact security": The 2016 report noted that cities had already begun to deploy AI technologies for public safety and security, and that applications aimed at improving the security of individuals and communities bring concerns that can be addressed if the systems are "well deployed." The report took a balanced view on the promise and potential pitfalls with a tone that the path can be negotiated to largely realize the benefits. The 2021 report highlights the widespread use of AI surveillance technologies worldwide, more urgently emphasizes the harms of invasive surveillance as a means to improving security of individuals and communities, and highlights that AI is increasingly viewed as important to dimensions of international security.

"Law enforcement agencies ➤ SEE WQ2.A are increasingly interested": The 2016 report noted that cities had already begun to deploy AI technologies for public safety and security, and it expressed that applications aimed at improving the security of individuals and communities bring concerns that can be addressed if the systems are "well deployed." The report took a balanced view on the promise and potential pitfalls, with a tone that the path can be negotiated to largely realize the benefits. The 2021 report highlights the widespread use of AI surveillance technologies worldwide, more urgently emphasizes the harms of invasive surveillance as a means to improving security of individuals and communities, and highlights that AI is increasingly viewed as important to dimensions of international security.

# Page 38

"not too distant future" **➤ SEE SQ11.A:** This concern remains, although employment was actually quite high right before the COVID-19 pandemic.

# Page 40

"sometimes to the detriment of interpersonal interaction" **➤ SEE SQ10.D:** It is interesting to note that the highlighted concern about social networks was the way they were interfering with face-to-face social interaction. Today, the danger of social networks generating the most attention is the creation of filter bubbles and the turbo-charged spread of misinformation.

# Page 41

"progressively more human-like" **➤ SEE SQ3.C:** At present, the most human-like text is produced by large-scale language models. Keeping the chatbots driven by these models from exhibiting antisocial or unwanted behavior remains a significant challenge.

"prevent their emergence": Arguably, they emerged. Discussion today is more focused on how to mitigate the resulting harms.

# Page 42

"help build trust": The concept of trust is central to the 2016 report but is much less prominent in the 2021 report. One reason for this difference is the perception that trust should not be viewed as an "add on" after an AI system is built and deployed as a way of getting everyone on board. Instead, those impacted by the implementation of an AI system should be engaged early on in the design of the system. Doing so builds trust by making the systems trustworthy instead of by building trust in systems after the fact.

"value they create for human lives": Both panels define the success of AI not as a technical exercise but as a means of enhancing human flourishing.

# Page 43

"three general policy recommendations": These remain great suggestions.

# Page 46

"as through it were human" **➤ SEE SQ3.B:** We propose reconceptualizing some of the challenge of the Turing Test as building an AI system that can communicate or work intelligently with a person without them thinking they are working with a person.

# Page 48

"regulation is inevitable" **➤ SEE SQ7.A:** Although, superficially, the 2021 report is more supportive of regulation, the two reports agree in broad strokes. Smart regulation is good and appropriate. Uninformed regulation is a bad idea.

"A recent multi-year study": The concern that a strict approach to regulation may actually exacerbate privacy concerns has, unfortunately, not been a part of recent public conversations on privacy in technology.